\documentclass{elsarticle}
\pdfoutput=1
\usepackage{amsmath}
\usepackage{amssymb}
\usepackage{xcolor}
\usepackage{url}
\usepackage[margin=1.0in]{geometry}
\usepackage{algorithm}
\usepackage[noend]{algpseudocode}
\usepackage{comment}
\usepackage{placeins}
\usepackage{afterpage}
\usepackage{amsthm}
\usepackage{soul}
\usepackage{siunitx}
\usepackage{mathtools}

\usepackage{tabularx}
\usepackage{subfig}
\usepackage{afterpage}
\usepackage{rotating}  
\usepackage{fancyhdr}
\usepackage{tabularx}
\usepackage{pgfplots}
\usepackage{caption}
\usepackage{babel,blindtext}
\usepackage{adjustbox}
\newtheorem{remark}{Remark}

\usepackage{natbib}
\usepackage{graphicx}

\usepackage{multirow}
\usepackage{slashbox}

\usepackage{float}
\newfloat{algorithm}{t}{lop}

\newtheorem{definition}{Definition}

\usepackage{lineno}
\begin{document}

\begin{frontmatter}

\title{A deep learning approach to solve forward differential problems on graphs}
\author[rvt]{Yuanyuan Zhao} 
\author[rvt2]{Massimiliano Lupo Pasini \corref{cor1}}\ead{lupopasinim@ornl.gov}

\address[rvt]{University of Alaska Fairbanks, 505 S Chandalar, Fairbanks, AK, USA, 99775}

\address[rvt2]{Oak Ridge National Laboratory, Computational Sciences and Engineering Division, 1 Bethel Valley Road, Oak Ridge, TN, USA, 37831}

\cortext[cor1]{Corresponding author}

\date{}

%\newpageafter{author}

\begin{abstract}
    We propose a novel deep learning (DL) approach to solve one-dimensional non-linear elliptic, parabolic, and hyperbolic problems on graphs. A system of physics-informed neural network (PINN) models is used to solve the differential equations, by assigning each PINN model to a specific edge of the graph. Kirkhoff-Neumann (KN) nodal conditions are imposed in a weak form by adding a penalization term to the training loss function. Through the penalization term that imposes the KN conditions, PINN models associated with edges that share a node coordinate with each other to ensure continuity of the solution and of its directional derivatives computed along the respective edges. Using individual PINN models for each edge of the graph allows our approach to fulfill necessary requirements for parallelization by enabling different PINN models to be trained on distributed compute resources. Numerical results show that the system of PINN models accurately approximate the solutions of the differential problems across the entire graph for a broad set of graph topologies. 
\end{abstract}


\end{frontmatter}

{\footnotesize \noindent This manuscript has been authored in part by UT-Battelle, LLC, under contract DE-AC05-00OR22725 with the US Department of Energy (DOE). The US government retains and the publisher, by accepting the article for publication, acknowledges that the US government retains a nonexclusive, paid-up, irrevocable, worldwide license to publish or reproduce the published form of this manuscript, or allow others to do so, for US government purposes. DOE will provide public access to these results of federally sponsored research in accordance with the DOE Public Access Plan (\url{http://energy.gov/downloads/doe-public-access-plan}).}

\section{Introduction}
Several physical, social, computer vision, and engineering problems can be modeled by solving ordinary differential equations (ODEs) or partial differential equations (PDEs) on graphs. 
%Examples of computer vision applications where differential equations are solved on a graph are the simplification of a point-cloud or mesh while preserving its geometry by solving an isotropic diffusion equation and filtering a textured image without loosing its texture instead of solving an anisotropic diffusion equation \cite{manifold_pdes}. 
For instance, solving first-order hyperbolic equations on graphs can model gas propagation in networks \cite{ STEINBACH2007345,  DOMSCHKE20151003,  Egger_2020, M2AN_2014__48_1_231_0, 1556-1801_2017_3_381,  Herty2010} and traffic flow \cite{piccoli_traffic_network_2006}, solving diffusion problems can model the behavior of carbon nano-structure \cite{carbon_nanostructures}, solving convection-diffusion problems  can model groundwater flow \cite{OPPENHEIMER2000223, 1937-5093_2011_4_1081, GARCIA2015120}, and solving the Schr\"{o}dinger equation defined on a graph can model quantum mechanical behavior of materials at atomic scale \cite{DUCA2021109324}. 
Examples of social applications are modeling international trades and space-temporal patterns of information spread and exchange over social networks \cite{social_network}.
In energy engineering, solving second order ODEs on graphs is needed to compute voltage phase angles and voltage frequencies in power networks \cite{Cheng2018}. In structural engineering, the Euler-Bernoulli equation defined on graphs describes the relationship between the deflection of a system of joint beams and the applied load \cite{Lagnese1993, KIIK20151871, Berkolaiko_2022}. In chemical engineering, solving reaction-diffusion equations on molecular graphs models the reactivity of atoms in molecular compounds \cite{WALLACE20172035}.

Differential equations defined on graphs are also used for control problems \cite{GUGAT202259, Gugat2011FlowCI, 7402932, family_exponentials, avdonin2022exact, AVDONIN201952, alam2021control}, where input parameters of a graph system are tuned to obtain a desired behavior at specific locations and/or time, as well as inverse problems \cite{Kurasov_2005, https://doi.org/10.48550/arxiv.2103.16727, vibration3040028, https://doi.org/10.48550/arxiv.2202.00944, GuanYangWu+2021+577+585}, where experimental measurements on the current stage of a graph systems are used to infer the underlying conditions that led to the current stage. 

Although several theoretical works studied the spectral properties of differential operators and the well-posedness of differential problems defined on graphs \cite{https://doi.org/10.48550/arxiv.1505.00185, MERCIER2008174, KOTTOS199976, VONBELOW1985309, https://doi.org/10.1007/s00023-017-0601-2, doi:10.1137/14097865X}, numerical methods to solve ODEs and PDEs on graphs have been mostly developed to address very specific applications \cite{mesh_refinement_gas_pdes, 10.1007/978-3-030-43651-3_45, GRUNDEL202260, 10.1007/978-3-319-27517-8_1,Herty2010, Gugat2011FlowCI, PESENSON2005203, Wybo2015}. Only in the last decade the numerical studies have been broadened to address a more general framework, with recent computational studies being dedicated to develop a general finite element framework for elliptic and parabolic differential operators defined on graphs \cite{10.1093/imanum/drx029, Leugering2017, Stoll2021}. These studies provide theoretical results for the well-posedness of the variational formulation of the differential problem for different classes of initial and boundary conditions, as well as convergence results along with numerical examples for computational validation of the finite element method (FEM). When the data is affected by noise, FEM interpolates both signal and noise. Stabilization techniques (e.g., Tikhonov regularization) \cite{ghattas_willcox_2021} reduce the effect of noise interpolation in FEM but they inevitably perturb the nature of the underlying problem.

An alternative to finite elements that also avoids interpolation arises from physics informed machine learning (ML) \cite{weather_piml, piml, https://doi.org/10.48550/arxiv.2204.00538}. In particular, physics-informed deep learning (DL) has shown effective performance to capture complex input-output relations between the data and the solution to a differential problem \cite{raissi2019physics, Chen:20, https://doi.org/10.48550/arxiv.2202.11821}, and graph neural network (GNN) models have been used in this context to solve differential equations on entire graphs \cite{graphNN_pde, https://doi.org/10.48550/arxiv.2108.01938}. Although the use of a GNN model can help solving a differential problem defined on a graph at a moderate scale using message passing policies to transfer information across nodes, such an approach does not scale to complex systems described by large graphs.

In this work, we propose a novel DL technique to solve forward ODE and PDE problems on graphs. Each edge of the graph is assigned a neural network (NN) to solve the differential equation on that specific edge. NNs associated with edges that share a node in common coordinate with each other through a penalty term in the training loss function that imposes the Kirkhoff-Neumann (KN) conditions in a weak-sense. Relying on fitting models rather than interpolating, our DL methodology sets a promising path to solve inverse and control differential problems on graphs when the data is affected by noise. Moreover, assignment of different edges to different NN models enables the algorithm to potentially leverage distributed high-performance computing (HPC) resources to concurrently train different NNs, thus allowing for the DL training to scale with the size of the graph.

The paper is organized as follows. In Section \ref{background} we introduce some definitions and notations to set the mathematical background used in the remainder of the manuscript, and describe three nonlinear differential problems defined on graphs. In Section \ref{pinn_section} we describe our DL approach to solve differential equations on metric graphs. 
In Section \ref{num} we present the numerical results for the three differential problems on graphs with varying topology. In Section \ref{conclusion} we summarize our conclusions and discuss future directions.

\section{Mathematical Background}\label{background}

\subsection{Differential equations defined on metric graphs}
\label{section_pdes}

A metric graph is a graph in which each edge is endowed with a metric structure. Often (but not always), its edges can be identified with intervals on the real line. The differential operator acts on functions defined on the edges and nodes of the underlying metric graph. 

Let $G(E,V)$ be a finite metric graph with a set of nodes $V$ and a set of edges $E$. Let $e_j \in E$ be the open segment (nodes excluded) that represents an edge, with $j$ denoting any integer in the range $\{1,\ldots, J\}$, where $J$ is the cardinality of $E$, and let $\overline{e}_j$ represent the closure of $e_j$ (nodes included). Each edge $e_j$ between two nodes is identified with an interval $(0,l_j)$, where one of its two nodes is identified with the coordinate $0$ and the other node is identified with the coordinate $l_j$. Let $\Gamma$ be the set of boundary nodes, that is, the nodes with only one incident edge. For cyclic graphs, $\Gamma = \emptyset$. 
%The choice of edge directions is arbitrary and only for the convenience of expressing equations and models. 

\begin{definition}
A function $y$ defined on $(G,E)$ is said to respect the KN conditions if the following equalities hold \cite{alexander2020}:
\begin{equation}\label{s-d}
\begin{cases}
y_{j_1}=y_{j_2} \quad e_{j_1}, e_{j_2} \sim v, \quad \text{for every} \quad  v \in V \setminus \Gamma, \\
\sum_{e_j \sim v } \partial y_j(v)=0, \quad \text{for every} \quad v \in V \setminus \Gamma
\end{cases}
\end{equation}
where $y_j$ denotes the restriction of the function $y$ on the edge $e_j \in E$, $\partial y_j(v)$ denotes the
derivative of $y_j$ at the node $v$ taken along the  edge  $e_{j}$
in the direction outwards the node, and $e_j \sim v$ means the edge $e_j$ is incident to the node $v$.
\end{definition}

We now characterize the three types of one-dimensional differential equations that we solve on a metric graph: non-linear elliptic, non-linear parabolic, and non-linear hyperbolic. In the following, we limit the discussion to differential problems where only the reaction term is allowed to be non-linear and we impose homogeneous Dirichlet boundary conditions on the boundary nodes. For each differential problem, we pre-define the exact solution by ensuring that its analytical expression has sufficient regularity to apply the differential operator on it, and the forcing term is calculated by applying the differential operator on the exact solution. 
When we set up the training of the PINN models, we pretend that we only know the forcing term, the boundary conditions, and the initial conditions (only for time-dependent problems), and we expect the PINN models to reconstruct the exact solution. 
The fact that we pre-define the exact solution gives us complete knowledge about the differential problem and provides a reliable validation of our DL method. 

\subsubsection*{Non-linear elliptic problem}
Let $\mu_j \in C^1(e_j)$ and $b_j \in C^1(e_j)$ for every $e_j \in E$ denote the diffusion and transport coefficient, respectively. Let $\sigma_j \in C^0(e_j)$ for every $e_j \in E$ denote the reaction term. Let $u_j \in C^2(e_j) \cap C^1(\overline{e_j})$ for every $e_j \in E$ be the exact solution to the ODE. 
With these notations, the non-linear elliptic problem defined on the metric graph $(G,E)$ is 

\begin{equation}  
\label{elliptic}
\begin{cases}
   \displaystyle \mathcal{L}_1(u)= -\frac{d }{dx} \bigg( \mu_j(x) \frac{d u_j(x)}{dx} \bigg ) + \frac{d}{dx}\bigg ( b_j(x)u_j(x) \bigg) + \sigma_j(x, u_j(x))u_j(x) =f_j(x), \\
   \hspace{5cm}  j=0, \dots, J-1, \quad 0 < x < l_j, \\
    
        u_j(v)=0 \quad e_j \sim v, \quad v \in \Gamma, \\
        u_{j_1}=u_{j_2} \quad e_{j_1}, e_{j_2} \sim v,\quad  v \in V \setminus \Gamma, \\
    
    \sum_{e_j \sim v} \partial u_j(v)=0, \quad v \in V \setminus \Gamma.
\end{cases}
\end{equation}
The regularity chosen on the exact solution $u_j$ ensures that the forcing term $f_j$ on the right-hand side of the equation is $f_j \in C^0(e_j)$ for every $e_j \in E$.

\subsubsection*{Non-linear parabolic problem}

Let $\mu_j \in C^0((0,T]; C^1(e_j))$ and $b_j \in C^0((0,T]; C^1(e_j))$ for every $e_j \in E$ denote the diffusion and transport coefficient, respectively. Let $\sigma_j \in C^0((0,T]; C^0(e_j))$ for every  $e_j \in E$ denote the reaction coefficient. 
Let $u_j \in C^0((0,T]; C^2(e_j) \cap C^1(\overline{e_j}))\cap C^1([0,T]; C^0(e_j))$ for every $e_j \in E$ be the exact solution of the parabolic PDE such that $u_j(v,t)=0$ for every $e_j \in E$ and $v \in \Gamma$ at any time $t\in [0,T]$. The regularity of the exact solution ensures that the initial condition $g_j \in C^2(e_j) \cap C^1(\overline{e_j})$ is compatible with the homogeneous Dirichlet boundary conditions since $g_j(v)=0$ for every $j\in J$ and $v \in \Gamma$. 
With these notations, the non-linear parabolic problem defined on the metric graph $(G,E)$ is 

\begin{equation} 
\label{parabolic}
\begin{cases}
   \displaystyle \mathcal{L}_2(u)=\frac{\partial}{\partial t} u_j(x,t) -\frac{\partial }{\partial x} \bigg( \mu_j(x,t) \frac{\partial u_j(x, t)}{\partial x} \bigg ) + \frac{\partial}{\partial x}\bigg ( b_j(x,t)u_j(x,t) \bigg) + \sigma_j(x, t, u_j(x,t))u_j(x,t) =f_j(x,t), \\ \hspace{5cm} j=0, \dots, J-1, \quad 0 < x < l_j, \quad 0 < t \le T,\\
        u_j(v,t)=0, \quad e_j \sim v, \quad v \in \Gamma, \quad 0 \le t \le T, \\
        u_{j_1}(v,t)=u_{j_2}(v,t), \quad e_{j_1}, e_{j_2} \sim v,\quad  v \in V \setminus \Gamma, \quad 0 \le t \le T, \\
    \sum_{e_j \sim v} \partial u_j(v,t)=0, \quad v \in V \setminus \Gamma, \quad 0 \le t \le T, \\
    u_j(x, 0) = g_j(x).
\end{cases}
\end{equation}
The regularity chosen on the exact solution $u_j$ ensures that the forcing term $f_j$ on the right-hand side of the equation is $f_j \in C^0((0,T]; C^0(e_j))$ for every $e_j \in E$.

\subsubsection*{Non-linear hyperbolic problem}
Let $b_j \in C^0((0,T]; C^1(e_j))$ for every $e_j \in E$ denote the transport coefficient. Let $\sigma_j \in C^0((0,T];C^0(e_j))$ for every  $e_j \in E$ denote the reaction coefficient. Let $u_j \in C^0((0,T]; C^1(\overline{e_j}))\cap C^1([0,T]; C^0(e_j))$ for every  $e_j \in E$ be the exact solution of the hyperbolic PDE such that $u_j(v,t)=0$ for every  $e_j \in E$ and $v \in \Gamma$ at any time $t\in [0,T]$. 
The regularity of the exact solution ensures that the initial condition $g_j \in C^0(\overline{e_j})$. 
With these notations, the non-linear hyperbolic problem defined on the metric graph $(G,E)$ is 
\begin{equation} 
\label{hyperbolic}
\begin{cases}
   \displaystyle \mathcal{L}_3(u)=\frac{\partial}{\partial t} u_j(x,t) + \frac{\partial }{\partial x}\bigg ( b_j(x,t)u_j(x,t) \bigg) + \sigma_j(x, t, u_j(x,t))u_j(x,t) =f_j(x,t), \\ \hspace{4cm} j=0, \dots, J-1,\quad 0 < x < l_j, \quad 0<t\le T, \\
           u_{j_1}(v,t)=u_{j_2}(v,t), \quad e_{j_1}, e_{j_2} \sim v,\quad  v \in V \setminus \Gamma, \quad 0 \le t \le T, \\
    \sum_{e_j \sim v} \partial u_j(v,t)=0, \quad v \in V \setminus \Gamma, \quad 0 \le t \le T, \\
    u_j(x, 0) = g_j(x).
  \end{cases}
\end{equation}
The regularity chosen on the exact solution $u_j$ ensures that the forcing term $f_j$ on the right-hand side of the equation is $f_j \in C^0((0,T];C^0(e_j))$ for every $e_j \in E$.

\begin{remark}
The regularity imposed on the analytical formula of the exact solution for each differential problem ensures the exact solution $u_j$ can be learnt by a NN \cite{hornik1989multilayer, funahashi1989approximate, leshno1993multilayer}.
\end{remark}

\section{System of physics-informed neural networks for differential problems defined on metric graphs}
\label{pinn_section}
PINN models are deep feedforward networks, also called feedforward NNs or multilayer perceptrons (MLPs) \cite{mlp, goodfellow} used to approximate the solution to differential problems \cite{pinns2019}. We adopt an all-at-once approach by defining an augmented domain where time and space are treated jointly. For time-dependent problems, both coordinates in space and time as passed as input features to the NN models.  
Given an edge $e_j$ to which the MLP has been assigned, the MLP has to compute an approximation of the solution
\begin{equation}
u(x), \quad 0< x < l_j \quad \text{for the elliptic problem}
\end{equation}
or
\begin{equation}
u(x, t), \quad 0< x < l_j, \quad 0\le t \le T \quad \text{for the hyperbolic/parabolic problem}
\end{equation}
An MLP provides an approximation $\tilde{u}$ of $u$ as a composite function: 
\begin{equation}
\tilde{u}(x) = F(x, \mathbf{w}) = h_\ell(h_{\ell-1}(\ldots h_0(x))) \quad \text{for the elliptic problem}
\label{composition}
\end{equation}
\begin{equation}
\tilde{u}(x,t)=F(x, t, \mathbf{w}) = h_\ell(h_{\ell-1}(\ldots h_0(x, t)))\quad \text{for the hyperbolic/parabolic problem}
\label{composition}
\end{equation}
where $\tilde{u}:\mathbb{R}^a \rightarrow \mathbb{R}$, $F:\mathbb{R}^a \times \mathbb{R}^n\rightarrow \mathbb{R}$ (with $a=1$ for elliptic problems and $a=2$ for hyperbolic or parabolic problems), $h_0:\mathbb{R}^a\rightarrow \mathbb{R}^{k_1}$, $h_\ell:\mathbb{R}^{k_\ell}\rightarrow \mathbb{R}$ and $h_p:\mathbb{R}^{k_{p}}\rightarrow \mathbb{R}^{k_{p+1}}$ for $p=1,\ldots,\ell-1$, and $\mathbf{w}\in \mathbb{R}^n$ denotes the regression coefficients of the NN across all the layers, with 
\begin{equation}
    n = (a \cdot k_1) \cdot (k_1 \cdot k_2) \cdot \ldots \cdot (k_{\ell-1} \cdot k_\ell).
\end{equation}
The number $\ell$ quantifies the complexity of the composition and is equal to the number of hidden layers in the MLP. Therefore, $h_1$ corresponds to the first hidden layer of the MLP, $h_2$ is the second hidden layer, and so on. {Each one of the functions $h_i$'s combines the regression coefficients between consecutive hidden layers through nonlinear activation functions that allow MLPs to learn nonlinear relations between the input $x$ and the output $u(x)$ for steady problems and between the input $(x,t)$ and the output $u(x,t)$ for time-dependent problems.}

\subsection{Dataset description}
In this section we describe how we formatted the data to train and validate the PINN models. 
From now on, we will use the following notation:
\begin{itemize}
\item $i$: index of a sample point on an edge
\item $N^{\text{space}}_{\text{train}}$: number of training data samples per unit length on each edge 
\item $N^{\text{space}}_{\text{validate}}$: number of validation data samples per unit length on each edge 
\end{itemize}

\subsubsection*{Dataset for elliptic problems}
The training data for points inside each edge of the graph is formatted as 
\begin{equation}\{(x_{j,i}, f_{j,i}) : \quad j =0, \dots, J-1, \quad i =0, \dots, N^{\text{space}}_{\text{train}}\cdot(l_j-1)\}
\end{equation}
and the boundary conditions are imposed with additional training data samples formatted as
\begin{equation}
\{(0, 0) \quad \text{and} \quad (l_j, 0) : \quad j =0, \dots, J-1\}.
\end{equation}
The validation data set is formatted as 
\begin{equation}
\{(x_{j,i}, u_{j,i}) : \quad j =0, \dots, J-1, \quad i =0, \dots, N^{\text{space}}_{\text{validate}}\cdot (l_j-1)\}.
\end{equation}

\subsubsection*{Dataset for hyperbolic and parabolic problems}
Denote with:
\begin{itemize}
\item $T$: time horizon for time-dependent differential problems
\item $\tau$: index of a sample point in the time interval $[0,T]$ 
\item $N^{\text{time}}_{\text{train}}$: number of training data samples in the time interval $[0,T]$ 
\item $N^{\text{time}}_{\text{validate}}$: number of validation data samples in the time interval $[0,T]$ 
\end{itemize}

\noindent The training data for points inside each edge of the graph at a time $t=t_\tau > 0$ is formatted as \begin{equation}\{(x_{j,i}, t_\tau, f_{j,i,\tau}) : \quad j =0, \dots, J-1, \quad i =0, \dots, N^{\text{space}}_{\text{train}}\cdot(l_j-1), \quad \tau=0, \dots, N^{\text{time}}_{\text{train}}-1\}
\end{equation}
and the boundary conditions are imposed with additional training data samples formatted as
\begin{equation}\{(0, t_\tau, 0) \quad \text{and} \quad (l_j, t_\tau, 0) : \quad j =0, \dots, J-1, \quad \tau=1, \dots, N^{\text{time}}_{\text{train}}-1\}.
\end{equation}
The initial condition is imposed by feeding the PINN model with training data for points over the entire edge (including the nodes at the extreme) at the initial time $t=0$
\begin{equation}\{(x_{j,i}, 0, u_{j,i,0}) : \quad j =0, \dots, J-1, \quad i =0, \dots, N^{\text{space}}_{\text{train}}\cdot (l_j-1)\}.
\end{equation}

\noindent The validation data to assess the performance of the PINN models is formatted as \begin{equation}\{(x_{j,i}, t_\tau, u_{j,i,\tau}) : \quad j =0, \dots, J-1, \quad i =0, \dots, N^{\text{space}}_{\text{validate}}\cdot(l_j-1), \quad \tau=0, \dots, N^{\text{time}}_{\text{validate}}-1\}.
\end{equation}

\subsection{Loss function}
\label{loss_section}
In this section we describe how we constructed the training loss function for the system of PINNs defined on edges of a metric graph. 
The loss function is made of multiple terms that are additively combined to define the global loss function which is minimized during the training of the PINN models. One term of the loss function computes the mismatch between target data and PINN predictions at the nodes of the metric graph, whereas the other term computes the mismatch between the target data and the PINN predictions along the edges. 
Denote with $j$ a generic edge in the metric graph and denote with $\mathbf{w}_j\in \mathbb{R}^{n}$ the parameters of the PINN model used to solve the differential problem on the edge with index $j$. 

\subsubsection*{Node and edge loss functions for elliptic problems}
Let us define
\begin{equation}
f_{j,i} = \mathcal{L}_1(u(x_{j,i})),\quad
\tilde{f}_{j,i} = \mathcal{L}_1(F(x_{j,i}, \mathbf{w}_j)),\quad
\tilde{u}_{j}(v) = F(x_j(v),\mathbf{w}_j) 
\end{equation}
where $v$ is one of the two nodes at the extremes of the edge the edge $e_j$. Based on the convention adopted in this manuscript, either $x_j(v) = 0$ or $x_j(v) = l_j$.

\noindent The loss function $node\_loss_v:\mathbb{R}^n\rightarrow \mathbb{R}$ at the node $v$ is defined as 
\begin{equation} \label{nloss1}
    \quad node\_loss_v(\mathbf{w}_j) =\begin{cases} \frac{1}{M_f} \left( \sum_{e_{j_1}, e_{j_2} \sim v} (\tilde{u}_{j_1}(v)-\tilde{u}_{j_2}(v))^2 +\left(\sum_{e_j \sim v} \partial \tilde{u}_{j}(v)\right)^2\right), \quad v \in V\setminus \Gamma \\ \frac{1}{M_f} \tilde{u}_{j}^2(v), \quad v \in \Gamma, \quad e_j \sim v,  \end{cases}
\end{equation}
where $\tilde{u}_{j}$ denotes the model outputs of $u_{j}$ in Equation \eqref{elliptic}, $e_j\sim v$ denotes an edge that has the node $v$ as one of the two extremes, and $e_{j_1}, e_{j_2} \sim v$ refers to a couple of edges that share the node $v$ in common, and 

\begin{equation}\label{Mfe}
M_f = \max_{j=0}^{J-1}\; \max_{i = 0}^{N^{\text{train}}_{\text{space}}*l_j-1}\;  |f_{i,j}|^2.
\end{equation}

\noindent The edge loss function $edge\_loss_j:\mathbb{R}^n\rightarrow \mathbb{R}$ is defined as 
\begin{equation} \label{eloss1}
    edge\_loss_j(\mathbf{w}_j) = \frac{1}{M_f} \frac{1}{N^{\text{train}}_{\text{space}}} \left( \sum_{i = 0}^{N^{\text{train}}_{\text{space}}\cdot (l_j-1)} (f_{j,i,\tau} - \tilde{f}_{j,i,\tau})^2\right)
\end{equation}
where $M_f$ is defined as in Equation \eqref{Mfe}.  We normalize both $node\_loss_v(\mathbf{w}_j)$ and $ edge\_loss_j(\mathbf{w}_j) $ by the factor $M_f$ so that we can use the same parameters to model different solutions make the loss functions more comparable.  

\subsubsection*{Node and edge loss functions for hyperbolic and parabolic problems}
Let us define
\begin{equation}
\tilde{u}_{j,\tau}(v, t_\tau) = F(v,t_\tau ,\mathbf{w}_j)
\end{equation}
and 
\begin{equation}
f_{j,i,\tau} = \mathcal{L}_2(u(x_{j,i},t_\tau)),\quad
\tilde{f}_{j,i,\tau} = \mathcal{L}_2(F(x_{j,i},t_\tau, \mathbf{w}_j)) \quad \text{for the parabolic problem}
\end{equation}
\begin{equation}
f_{j,i,\tau} = \mathcal{L}_3(u(x_{j,i},t_\tau)),\quad
\tilde{f}_{j,i,\tau} = \mathcal{L}_3(F(x_{j,i},t_\tau, \mathbf{w}_j)) \quad \text{for the hyperbolic problem}
\end{equation}

\noindent The loss function $node\_loss_v:\mathbb{R}^n\rightarrow \mathbb{R}$ at the node $v$ is defined as 
\begin{equation} \label{nloss2}
    node\_loss_v(\mathbf{w}_j) =\begin{cases} \frac{1}{M_f} \bigg(\sum_{e_{j_1}, e_{j_2} \sim v}\sum_{\tau =0}^{N^{\text{train}}_{\text{time}}-1} (\tilde{u}_{j_1,\tau}(v)-\tilde{u}_{j_2,\tau}(v))^2 \\ \hspace{5cm} +\sum_{e_j \sim v}\sum_{\tau=0}^{N^{\text{train}}_{\text{time}}-1} (\partial \tilde{u}_{j,\tau}(v))^2\bigg), \quad v \in V\setminus \Gamma, \\ \frac{1}{M_f} \sum_{\tau=0, \dots, N^{\text{train}}_{\text{time}}-1} \tilde{u}_{j,\tau}^2(v), \quad v \in \Gamma, \quad e_j \sim v, \end{cases}
\end{equation}
where $\tilde{u}_j$ denotes the model outputs of $u_j$ in Equations \eqref{parabolic} and \eqref{hyperbolic}. 

\noindent The edge loss function $edge\_loss_j:\mathbb{R}^n\rightarrow \mathbb{R}$ is defined as 
\begin{multline} \label{eloss2}
    edge\_loss_j(\mathbf{w}_j) = \frac{1}{(N^{\text{train}}_{\text{space}}\cdot l_j)N^{\text{train}}_{\text{time}}} \left(\frac{1}{M_f} \sum_{i = 0}^{N^{\text{train}}_{\text{space}}\cdot(l_j-1)}\;\sum_{\tau=0}^{N^{\text{train}}_{\text{time}}-1} (f_{j,i,\tau} - \tilde{f}_{j,i,\tau})^2 \right.
    \\ \left.+ \frac{1}{M_u} \sum_{i=0}^{N^{\text{train}}_{\text{space}}\cdot (l_j-1)} (u_{j,i,0} - \tilde{u}_{j,i,0})^2\right) 
\end{multline}
where 

\begin{equation}M_u= \max_{j=0}^{J-1} \;\max_{i = 0}^{N^{\text{train}}_{\text{space}}\cdot(l_j-1)} \;  \;|u_{i,j,0}|^2.
\end{equation}
\subsubsection*{Cumulative loss function for the system of PINNs over the entire metric graph}

 Let us assume for simplicity that the PINN architecture is the same on each edge. Denote with $\mathbf{w}_j$ the regression coefficients for the edge $e_j$ and denote with
\begin{equation}
\mathbf{w}_{\text{tot}} = \begin{bmatrix} \mathbf{w}_1 \\ \vdots \\ \mathbf{w}_J
\end{bmatrix}  \in \mathbb{R}^{nJ}
\end{equation}
the aggregation of regression coefficients for each PINN model assigned to one edge in the metric graph. 
The global loss function $L:\mathbb{R}^{nJ}\rightarrow \mathbb{R}$ is defined as follows
\begin{equation}
L(\mathbf{w}_{\text{tot}})= \sum_{j \in J} edge\_loss_j(\mathbf{w}_{j}) + \lambda \bigg( \sum_{i \in I}\sum_{j \in J(v_i)} node\_loss_i(\mathbf{w}_{j}) \bigg).
\label{global_loss}
\end{equation}
The continuity of the solution and its directional derivatives along the edges is maintained by imposing the KN conditions in a weak form through the second term on the right side of Equation \eqref{global_loss}. 
We use the basic differential multiplier method (BDMM) \cite{platt1987constrained} to iteratively update the values of the regression coefficients $\mathbf{w}_{\text{tot}}$ and the values of the penalization term $\lambda$ from iteration $q$ to iteration $(q+1)$ during the training of the system of PINN models according to the following formula
\begin{equation}
\begin{cases} 
\mathbf{w}_{\text{tot}, q+1} = \mathbf{w}_{\text{tot}, q} -lr_{\mathbf{w}_{\text{tot}}} \cdot \nabla_{\mathbf{w}_{\text{tot}}} L(\mathbf{w}_{\text{tot}, q})\\
\lambda_{q+1} = \lambda_q + lr_\lambda \cdot \left(\sum_{v \in V}\sum_{e_j \sim v} node\_loss_v(\mathbf{w}_{j, q})\right) 
\end{cases}
\label{BDMM_update}
\end{equation}
where $lr_{\mathbf{w}_{\text{tot}}}$ and $lr_\lambda$ are the learning rates for $\mathbf{w}_{\text{tot}}$ and $\lambda$, respectively. 
We highlight that the BDMM step in Equation \eqref{BDMM_update} performs a step of gradient descent to update the aggregated regression coefficient $\mathbf{w}_{\text{tot}}$ and performs a step of gradient ascent to update the value of the penalization term $\lambda$.

\section{Numerical results} \label{num}
We present numerical results obtained with the predictive performance of PINN models to solve the non-linear elliptic, non-linear parabolic, and non-linear hyperbolic problems defined in \eqref{elliptic}, \eqref{parabolic}, and \eqref{hyperbolic} on metric graphs. 
In Section \eqref{setup_problems}, we specify the values of the coefficients of the equations, boundary conditions, and initial conditions (when needed) for the differential problems we solve. 
In Section \ref{pinn_setup}, we characterize the neural network architectures to construct the PINN model for each problem. 
In Section \ref{one_edge_results}, we compare the accuracy of PINN models using different activation functions to solve differential problems on a single edge. The conclusions drawn from this first set of studies will be used to select the activation function to perform the numerical tests described in the following sections. In Section \ref{star_graph_results}, we assess the robustness of our DL approach for varying values of the degree of a node (i.e., number of incidental edges). In Section \ref{general_graph_results}, we solve the differential problems on two graphs with complex topologies.  

\subsection{Setup of the differential problems}
\label{setup_problems}
For all the differential problems considered, we set the coefficient of the differential equation as
$\mu_j(x) = 1$, $b_j(x)=0$, and $\sigma_j(x,u_j(x))=[u_j(x)]^2$ for $j=0,\ldots,J-1$. 

\noindent For the non-linear elliptic problem, we set the exact solution $u_j$ on each edge as 
\begin{equation}
u_j(x) = \cos\left(\frac{2  \pi x}{l_j} \right)-1), \quad j=0, \dots, J-1, \quad 0 \le x \le l_{j-1}.
\end{equation}

\noindent For the non-linear parabolic problem, we set the exact solution $u_j$ on each edge as 
\begin{equation}
u_j(x,t)= 30 \exp(-t)  \left(\cos\left(\frac{2  \pi x}{l_j} \right)-1\right),
\quad j=0, \dots, J-1, \quad 0 \le x \le l_{j-1}, \quad 0, \le t \le T.
\end{equation}

\noindent For the non-linear hyperbolic problem, we set the exact solution $u_j$ on each edge as 
\begin{equation}
u_j(x,t)= 10 \cos(t) \left( \cos\left(\frac{2  \pi x}{l_j} \right)-1\right),
\quad j=0, \dots, J-1, \quad 0 \le x \le l_{j-1}, \quad 0, \le t \le T.
\end{equation}

\noindent For each problem, we compute the forcing terms $f_j$ by applying the differential operator on $u_j$, which for elliptic problems consist in evaluating $\mathcal{L}_1(u_j(x))$ and for hyperbolic and parabolic problems consists in evaluating $\mathcal{L}_2(u_j(x,t))$ and $\mathcal{L}_3(u_j(x,t))$, respectively. For the parabolic and hyperbolic systems, we set the initial conditions $g_j(x) = u_j(x,0)$. We use the forcing term data $f_j$, the boundary value data, and the initial value data $g_j$ to train the neural networks and calculate an approximate solution $\tilde{u}_j$ by minimizing the training loss functions defined in Section \ref{loss_section}. We validate the accuracy of the neural network by computing the validation error as the mean absolute difference between the exact solution $u_j$ and the DL approximation $\tilde{u}_j$ across all validation sample points. For the non-linear elliptic problem, the validation error is defined as 
\begin{equation}
    \operatorname{err}= \frac{1}{J (N_{\text{validate}}^{\text{space}}\cdot l_j)} \sum_{j=0}^{J-1}\;\; \sum_{i=0}^{N_{\text{validate}}^{\text{space}}\cdot (l_j-1)} \left|u_{j,i} - \tilde{u}_{j,i}\right|. 
\end{equation}
For the parabolic and hyperbolic problems, the validation error is defined as 
\begin{equation}
    \operatorname{err}=  \frac{1}{J (N_{\text{validate}}^{\text{space}}\cdot l_j) N_{\text{validate}}^{\text{time}}} \sum_{j=0}^{J-1}\;\;\sum_{i=0}^{N_{\text{validate}}^{\text{space}}\cdot (l_j-1)}\;\;\sum_{k=0}^{N_{\text{validate}}^{\text{time}}-1} \left|u_{j,i,k} - \tilde{u}_{j,i,k}\right|.
\end{equation}

\subsection{PINN architecture}
\label{pinn_setup}
For all three differential problems, we performed hyperparameter optimization over the number of layers and number of neurons in each layer of the PINN model. As a result, we selected the PINN architectures described in Table \ref{pinn_hyperparameters} due to their relatively small output errors. For the non-linear elliptic problem, the input layer of the PINN architecture has one neuron because the spatial domain is one-dimensional. Each network contains two hidden layers with 25 neurons each, and one output layer has one neuron because the solution attains scalar values. For the non-linear parabolic problem and the non-linear hyperbolic problem, the input layer has two neurons (one for the space coordinate and one for the time coordinate). Each network contains two hidden layers with 50 neurons each, and one output layer has one neuron because the solution attains scalar values. The hyperparameters used to construct the PINN model for each differential problems solved are provided in the Table \ref{pinn_hyperparameters}.

\begin{table}[!htbp]
\centering
\begin{tabular}{|l|*{3}{|p{25mm}|}}
\hline
&{Elliptic}&{Parabolic}&{Hyperbolic} \\\hline\hline
\# Neurons per layer & [1 25 25 1] & [2 50 50 1] & [2 50 50 1] \\ \hline
\# network parameters & 1350 & 5300 & 5300 \\ \hline
\end{tabular}
\caption{Hyperparameters for PINN training}
\label{pinn_hyperparameters}
\end{table}

\subsection{Training setup}
The PINN models are implemented with the PyTorch library and the graph domains are defined using the PyTorch Geometric \cite{fey_2019, torch_geometric} library. 
The training is performed with the Adam method \cite{adam} with an initial learnign rate set to $1\mathrm{e}-3$. 
The number of training data samples and validation data samples used for each iterative update of $\mathbf{w}_{\text{tot}}$ and $\lambda$ is provided in Table \ref{pinn_dataset}. 
The training of each DL model was performed on an NVIDIA K40 GPU provided by the Compute and Data Environment for Science (CADES) at Oak Ridge National Laboratory (ORNL). 

\begin{table}[!htbp]
\centering
\begin{tabular}{|l|*{3}{|p{25mm}|}}
\hline
&{Elliptic}&{Parabolic}&{Hyperbolic} \\\hline\hline
$N^{\text{train}}_{\text{space}}$  & 2000 & 100
& 100 
\\\hline
$N^{\text{train}}_{\text{time}}$ & -
& 20 & 20 \\\hline
$N^{\text{validate}}_{\text{space}}$  & 500 & 500
& 500
\\\hline
$N^{\text{validate}}_{\text{time}}$ & -
& 500 & 500 \\\hline
\end{tabular}
\caption{Training and validation datasets}
\label{pinn_dataset}
\end{table}

\subsection{Comparison of activation functions for PINNs on one edge}
\label{one_edge_results}

The choice of the activation function can drastically affect the predictive performance of the NNs \cite{hornik1989multilayer, leshno1993multilayer, funahashi1989approximate, ziyin2020neural}. This motivates us to assess the predictive performance of the PINN model to solve the differential problems in Equations \eqref{elliptic}, \eqref{parabolic}, and \eqref{hyperbolic} on a single edge using different activation functions. The PINN architecture is fixed and the activation function is the only hyperparameter that is allowed to change. The activation functions we compare are \texttt{ReLU}, sigmoid, $\sin^2(x)$, and $\sin^2(x)+x$. For different activation functions used, we assess the robustness of the PINN model to handle edges of variable length by measuring the final accuracy of the PINN model to solve the differential problems on segments of length equal to 1, 5, and 10. 

The validation errors of the PINN model in solving the non-linear elliptic, non-linear parabolic, and non-linear hyperbolic problems are shown in Tables \ref{eledge}, \ref{paredge}, and \ref{adedge}, respectively. The plots using different activation functions are provided in the Appendix.
For all three differential problems, when the edge length is $1$, the performances of the PINN models using the sigmoid, $\sin^2(x)$, and $\sin^2(x)+x$ are similar, and they all significantly outperform the PINN model using the \texttt{ReLU} activation function. When the edge length is increased, we notice that the PINN models using the $\sin^2(x)$ and $\sin^2(x)+x$ activation functions produce significantly smaller errors than the PINN models using the \texttt{ReLU} and sigmoid activation functions.  
Our numerical results in Tables \ref{eledge}--\ref{adedge} show that $\sin^2(x)$ still outperforms the monotonic activation function $\sin^2(x)+x$ proposed in \cite{ziyin2020neural} when the function to learn is indeed periodic. In particular, we notice that the advantage of the activation function $\sin^2(x)$ over $\sin^2(x)+x$ becomes more pronounced when the number of wavelengths included in the edge increases due to an increase of the edge length. Based on this assessment, since the numerical results presented in this work are limited to approximating periodic solutions to differential problems, henceforth we choose $\sin^2(x)$ as activation function for the PINN models. 

%%%%%%%%%%%%%%%%%%%%%%%%%%%%%%%%%%%%%%%%%%%%%%%%%

\begin{table}[!htbp]
\centering
\begin{tabular}{|l|*{4}{|p{20mm}|}}
\hline
\backslashbox{edge length}{activation func.}
&\makebox[3em]{relu}&\makebox[3em]{sigmoid}&\makebox[3em]{$\sin^2(x)$}
&\makebox[3em]{$\quad\quad \sin^2(x)+x$}\\\hline\hline
1& $1.4655e+00$ & $4.2809e-04$ & $1.3542e-04$ & $1.8515e-05$
 \\ \hline
5 &  $1.1142e+00$ &  $3.2258e-01$ & $2.4161e-04$ & $3.1159e-03$
 \\\hline
10 & $1.0979e+00$ & $5.6701e-01$ & $1.3437e-03$ & $5.0190e-03$ \\\hline
\end{tabular}
\caption{Validation error of the PINN model using \texttt{ReLU}, sigmoid, $\sin^2(x)$ and $\sin^2(x)+x$ activation functions to solve the non-linear elliptic problem on edge with length equal to 1, 5, and 10.} \label{eledge}
\end{table}

\begin{table}[!htbp]
\centering
\begin{tabular}{|l|*{4}{|p{20mm}|}}
\hline
\backslashbox{edge length}{activation func.}
&\makebox[3em]{relu}&\makebox[3em]{sigmoid}&\makebox[3em]{$\sin^2(x)$}
&\makebox[3em]{$\quad\quad \sin^2(x)+x$}\\\hline\hline
1& $3.9598e-01$ & $1.4537e-01$  & $2.0664e-01$  & $2.0652e-01$
 \\ \hline
5 & $5.5414e+00$  &  $4.8648e+00$   & $2.1506e-01$ &  $4.0540e-01$
 \\\hline
10 & $1.0596e+01$ &  $9.7715e+00$  &  $1.9799e-01$ & $6.9083e-01$ \\\hline
\end{tabular}
\caption{Validation error of the PINN model using \texttt{ReLU}, sigmoid, $\sin^2(x)$ and $\sin^2(x)+x$ activation functions to solve the non-linear parabolic problem on edge with length equal to 1, 5, and 10.} \label{paredge}
\end{table}

\begin{table}[!htbp]
\centering
\begin{tabular}{|l|*{4}{|p{20mm}|}}
\hline
\backslashbox{edge length}{activation func.}
&\makebox[3em]{relu}&\makebox[3em]{sigmoid}&\makebox[3em]{$\sin^2(x)$}
&\makebox[3em]{$\quad\quad \sin^2(x)+x$}\\\hline\hline
1& $1.9027e-01$ & $5.6781e-02$ & $1.4505e-02$ &   $1.9684e-02$
 \\ \hline
5 &  $2.4712e+00$ &  $3.0064e-01$ &  $3.1154e-02$ & $1.4285e-01$
 \\\hline
10 & $4.4436e+00$ & $2.4927e+00$ & $4.7374e-02$ &  $4.9312e-01$\\\hline
\end{tabular}
\caption{Validation error of the PINN model using \texttt{ReLU}, sigmoid, $\sin^2(x)$ and $\sin^2(x)+x$ activation functions to solve the non-linear hyperbolic problem on one edge with length equal to 1, 5, and 10.} \label{adedge}
\end{table}

\subsection{Star graphs with increasing number of edges}
\label{star_graph_results}
\begin{figure}[ht]
\subfloat[Star graph with three edges]{
	\begin{minipage}[c][0.4\textwidth]{
	   0.5\textwidth}
	   \includegraphics[width=0.9\textwidth]{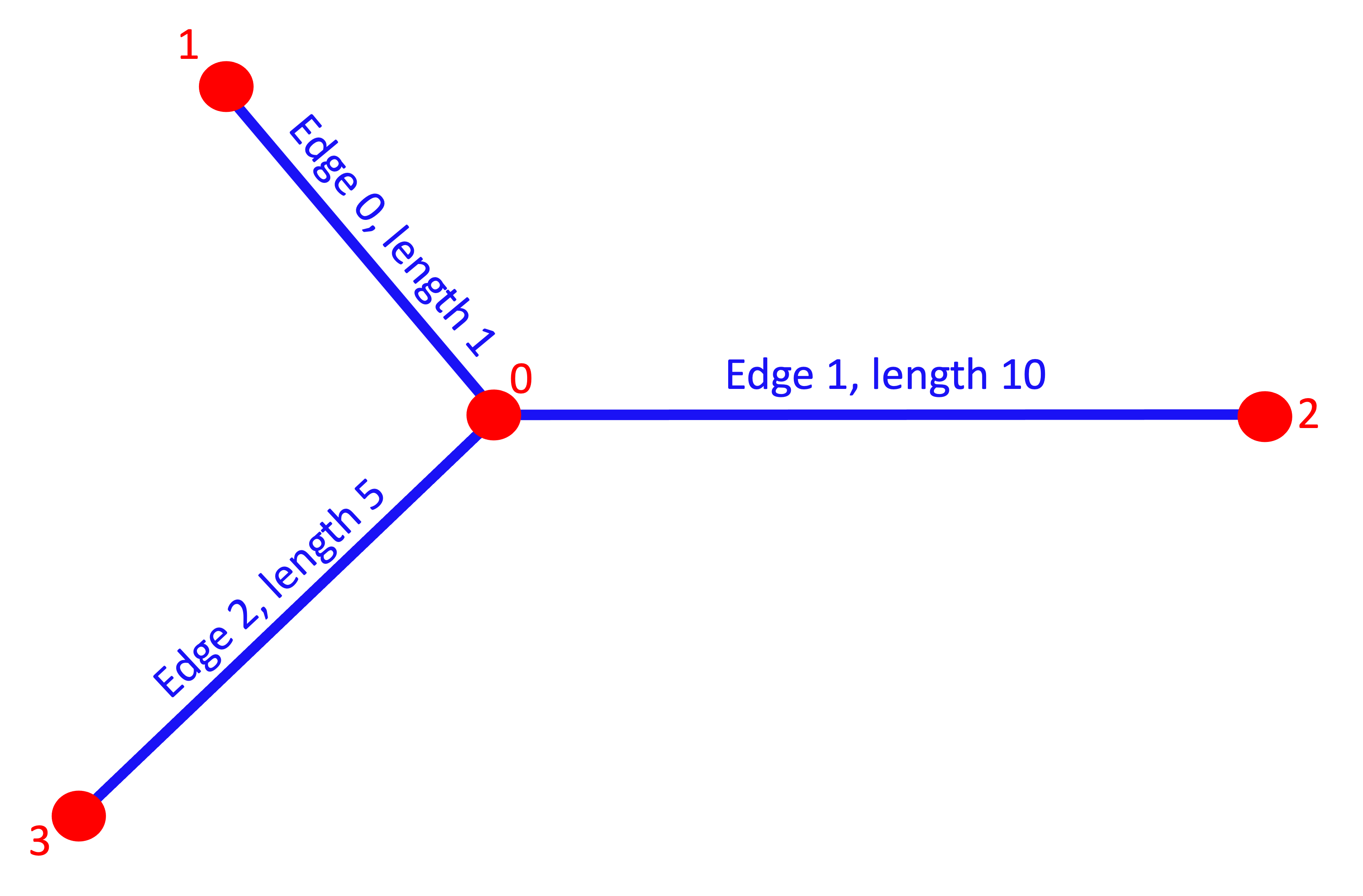}
	\end{minipage}}
\subfloat[Star graph with five edges]{
	\begin{minipage}[c][0.4\textwidth]{
	   0.5\textwidth}
	   \includegraphics[width=0.9\textwidth]{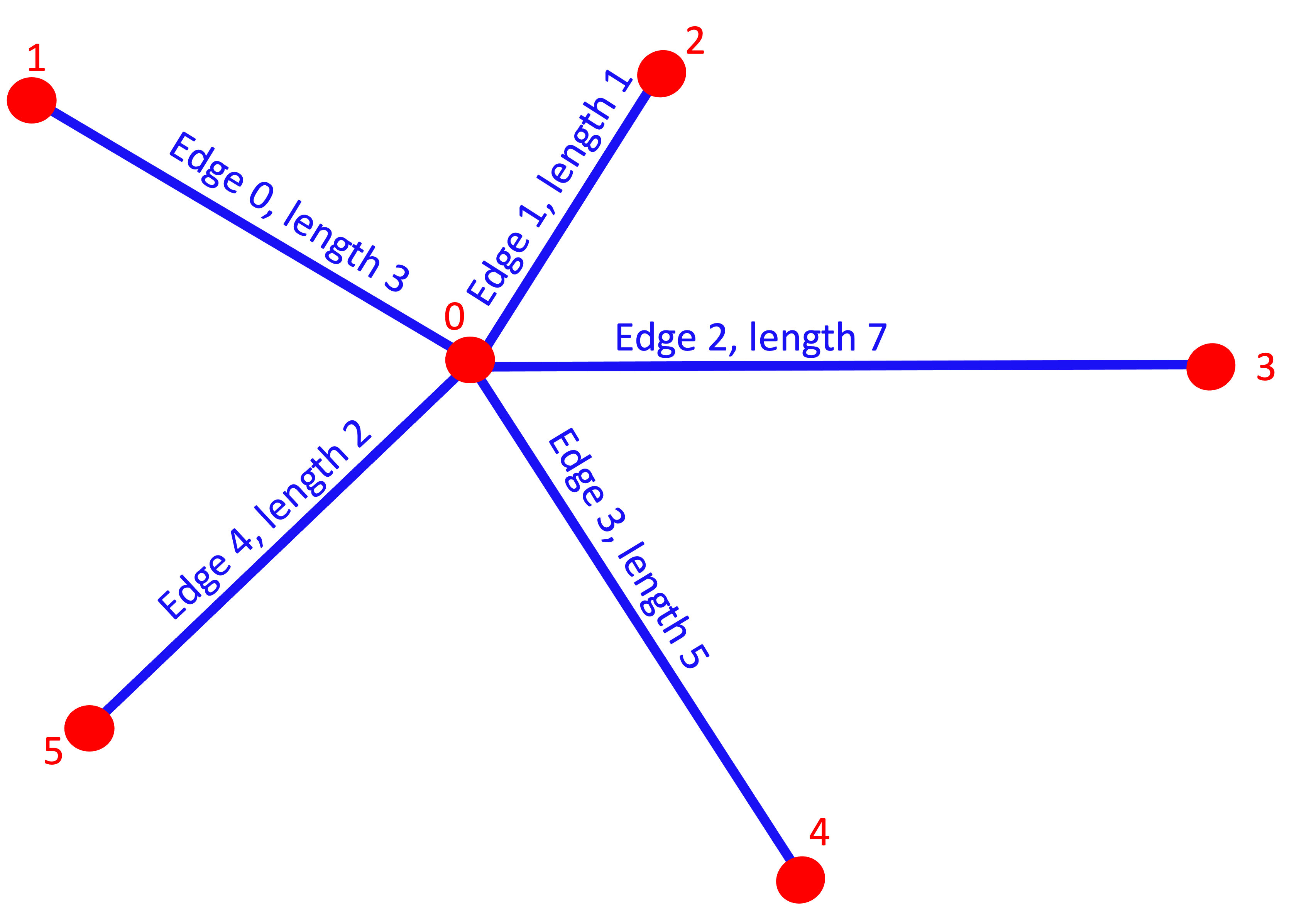}
	\end{minipage}}	\newline

\subfloat[Star graph with seven edges]{{\hspace{5cm}
	\begin{minipage}[c][0.4\textwidth]{
	   0.5\textwidth}
	   \includegraphics[width=\textwidth]{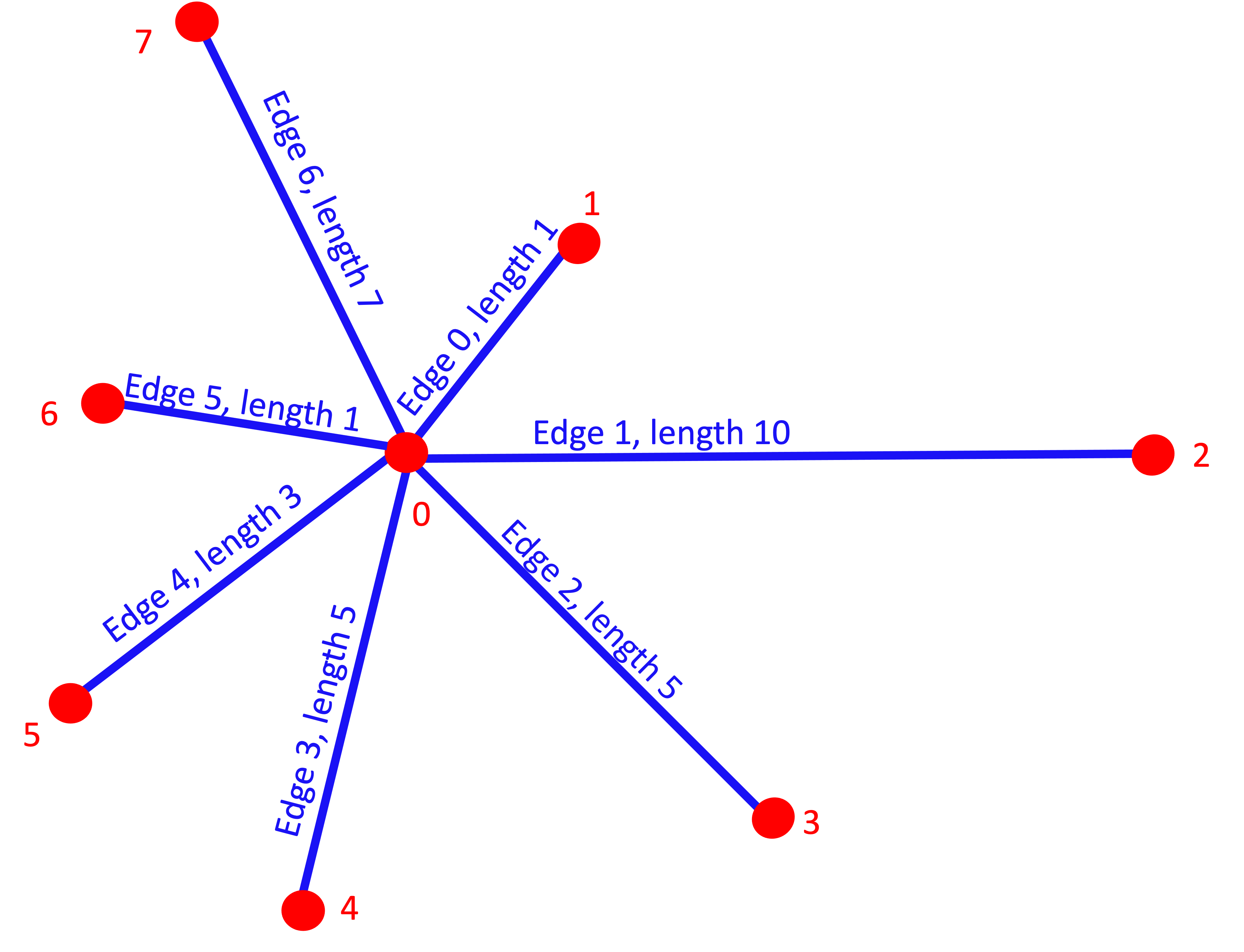}
	\end{minipage}}}
 \hfill 
\caption{Star graphs with three edges (top left), five edges (top right), and seven edges (bottom center).}
\label{star_graph_figure}
\end{figure}

Since previous work showed that FEM applied to metric graph can become unstable increases when the number of incidental edges to a node \cite{10.1093/imanum/drx029}, we dedicate this section to empirically studying the accuracy of the system of PINN models defined on a star graph when the number of edges incidental to the central node increases. To this end, we use a system of PINN models to solve the differential problems in Equations \eqref{elliptic}, \eqref{parabolic}, and \eqref{hyperbolic} on a star graph with a number of incidental edges to the central node equal to three, five, and seven. The details of the topological structures of the star graphs are provided in Figure \ref{star_graph_figure}. 

In Table \ref{lambdaerror_stargraph} we provide the optimized value of $\lambda$ and the validation error. Conditions at nodes are critical components of every differential problem defined on metric graphs, which justifies the need to properly tune the value of the penalization coefficient $\lambda$ through which the KN conditions are imposed in a weak sense. The results in Table \ref{lambdaerror_stargraph} show that BDMM is an effective technique to tune the penalization coefficient $\lambda$ used to combine the loss functions at the nodes with the loss functions on the edges. 
When the number of edges incidental to the central node increases, we notice that the error remains of the same order of magnitude. 

\begin{figure}
{\hspace{5cm}\includegraphics[width=0.35\textwidth]{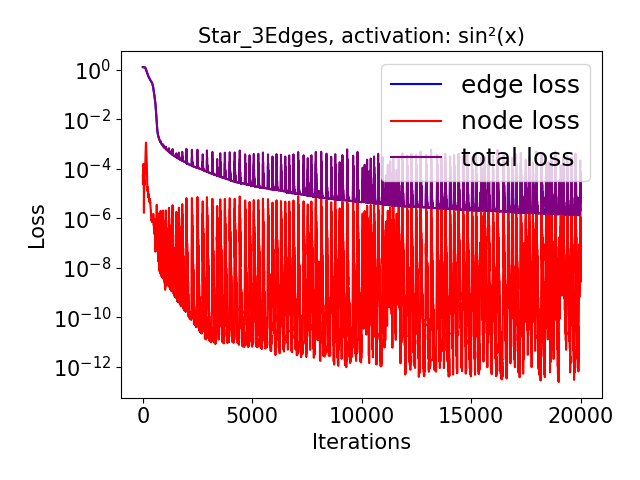}}\\
\includegraphics[width=0.33\textwidth]{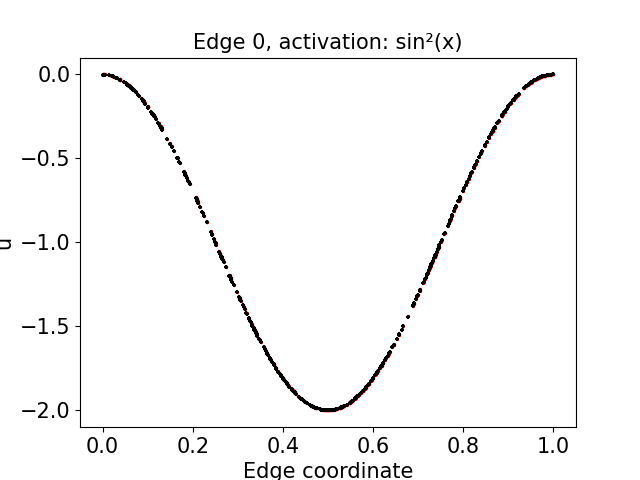}
\includegraphics[width=0.33\textwidth]{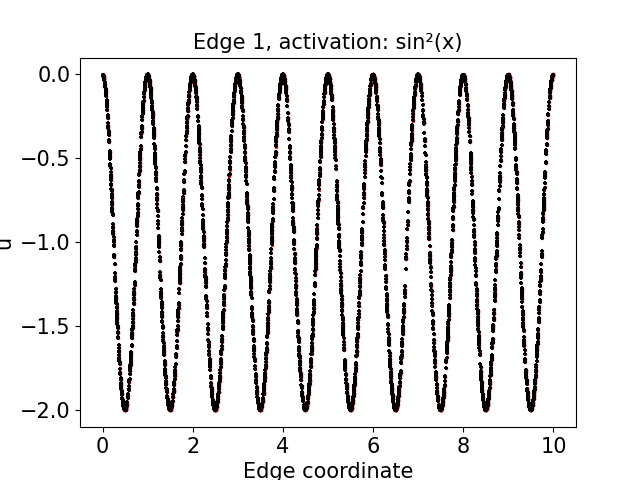}
\includegraphics[width=0.33\textwidth]{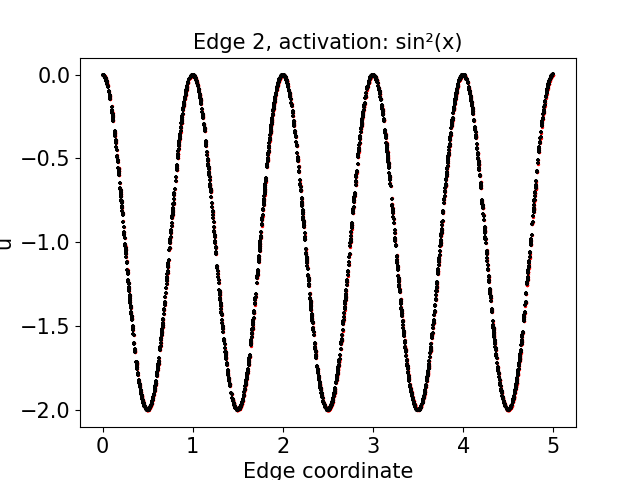}
\caption{Validation error (top figure) and comparison on each edge (figures on the second, third, and fourth line) of the exact solution (red) and PINN approximated solution (black) of Model output for the non-linear elliptic problem on the star graph with three edges}
\label{star3_elliptic}
\end{figure}

\begin{figure}[htbp!]
{\hspace{5cm}\includegraphics[width=0.35\textwidth]{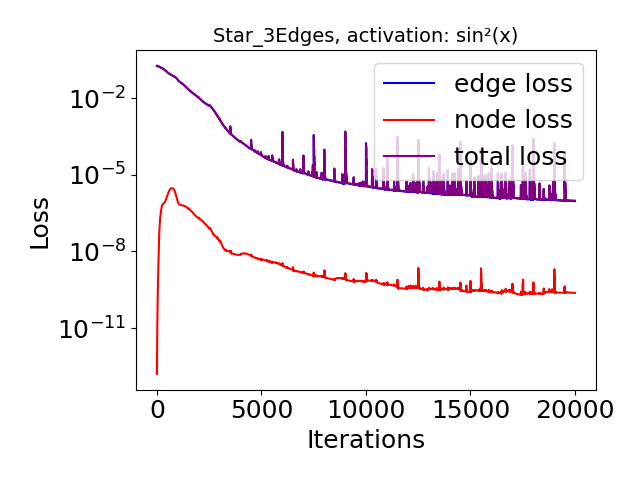}}\\
\vspace{-0.8cm} 
\adjustbox{trim={.25\width} {0\height} {.15\width} {0\height},clip}{\includegraphics[width=0.55\textwidth]{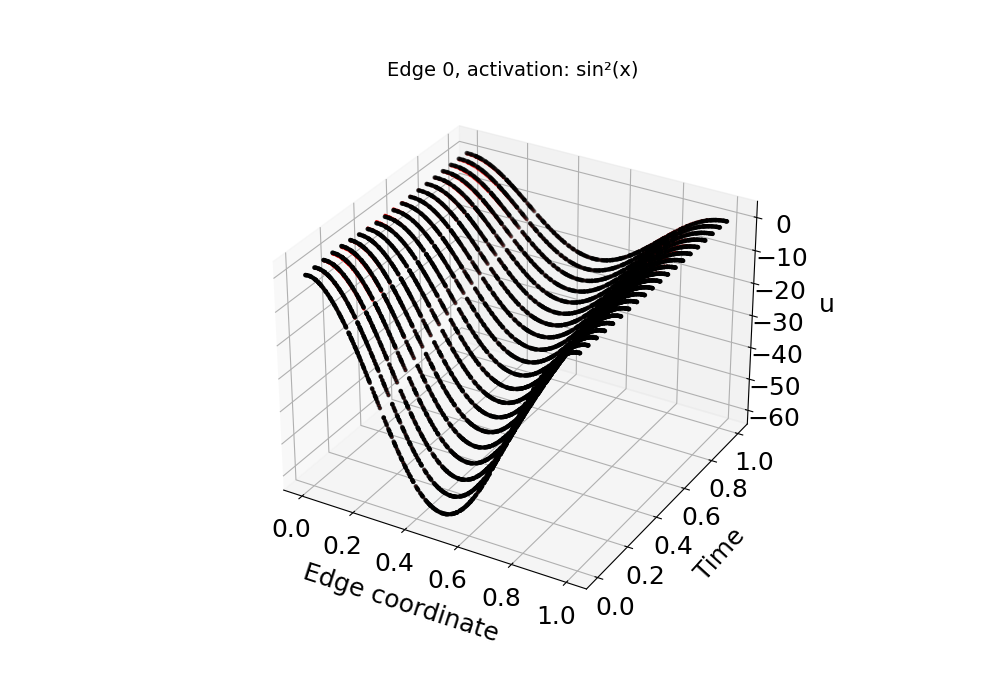}}
\adjustbox{trim={.25\width} {0\height} {.15\width} {0\height},clip}{\includegraphics[width=0.55\textwidth]{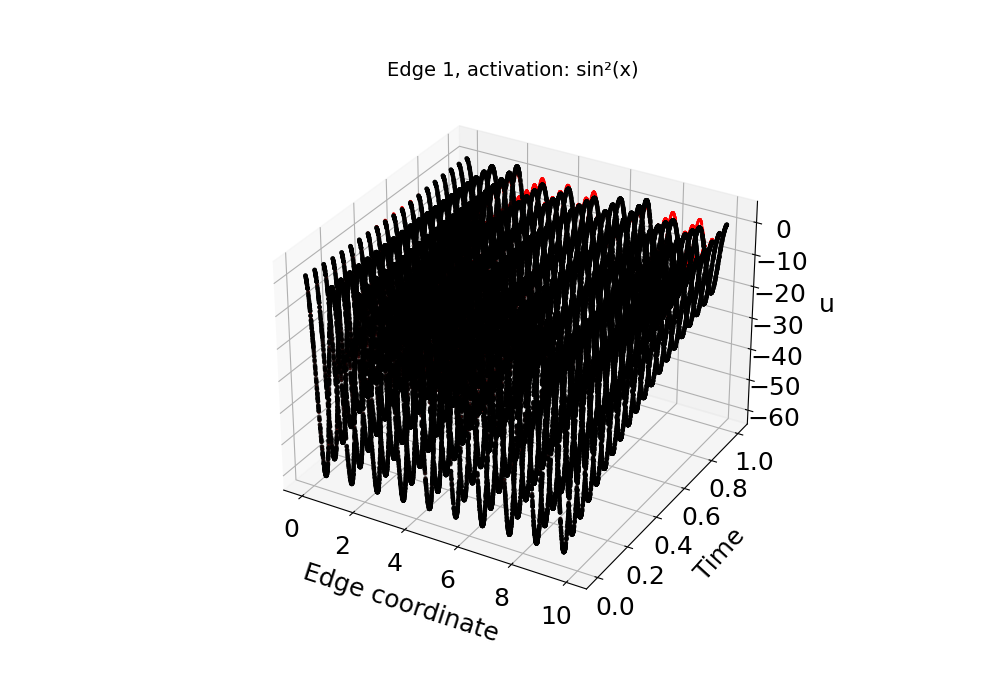}}
\adjustbox{trim={.25\width} {0\height} {.15\width} {0\height},clip}{\includegraphics[width=0.55\textwidth]{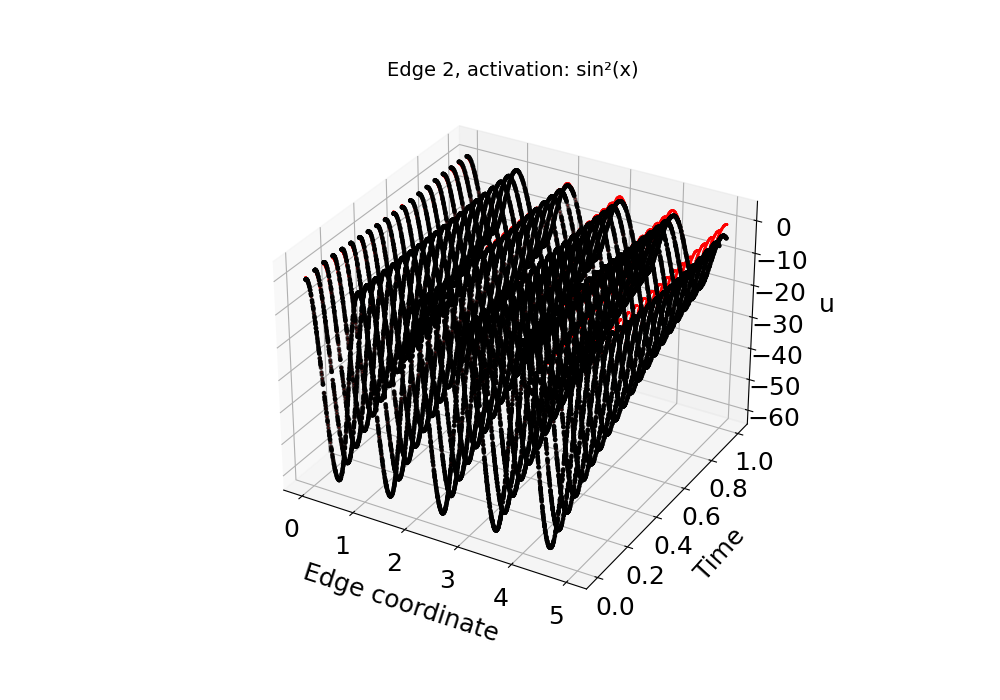}}
\vspace{0.4cm}
\caption{Validation error (top figure) and comparison on each edge (figures on the second line) of the exact solution (red) and PINN approximated solution (black) of the non-linear parabolic problem on the star graph with three edges}
\label{star3_parabolic}
\end{figure}

\begin{figure}[htbp!]
{\hspace{5cm}\includegraphics[width=0.35\textwidth]{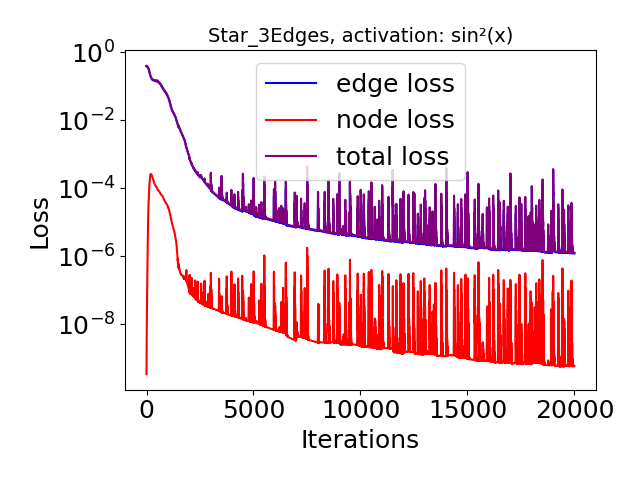}}\\
\vspace{-0.8cm}
\adjustbox{trim={.25\width} {0\height} {.15\width} {0\height},clip}{\includegraphics[width=0.55\textwidth]{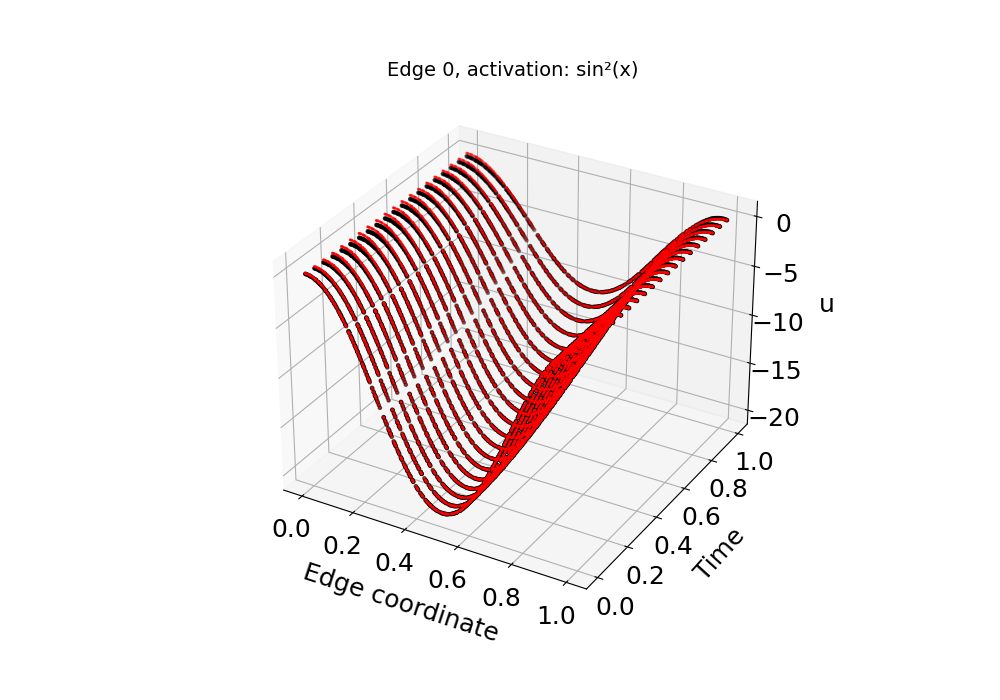}}
\adjustbox{trim={.25\width} {0\height} {.15\width} {0\height},clip}{\includegraphics[width=0.55\textwidth]{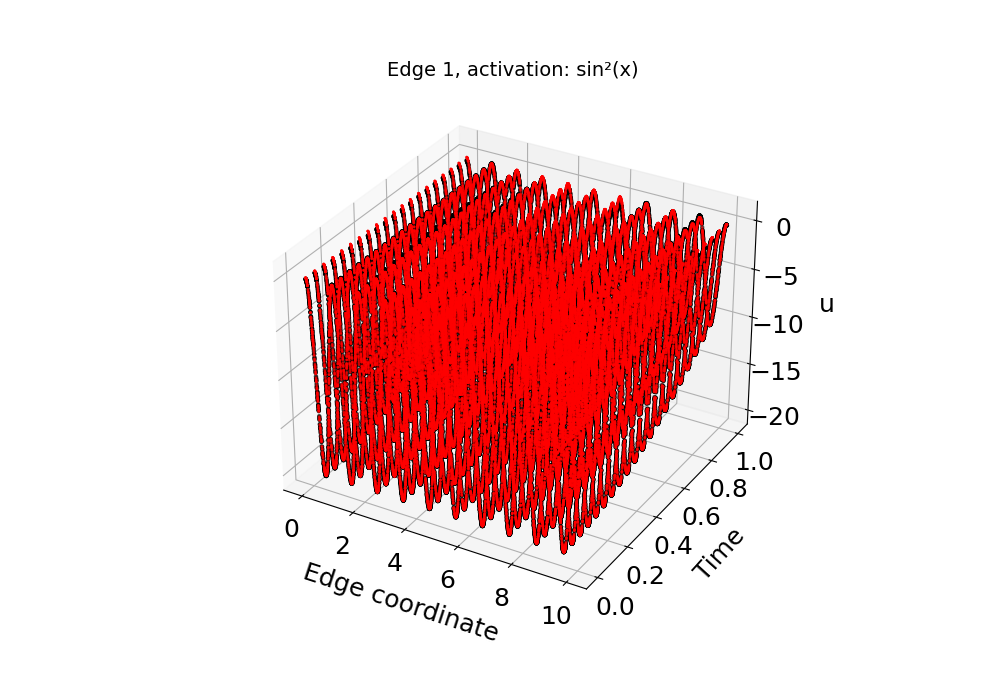}}
\adjustbox{trim={.25\width} {0\height} {.15\width} {0\height},clip}{\includegraphics[width=0.55\textwidth]{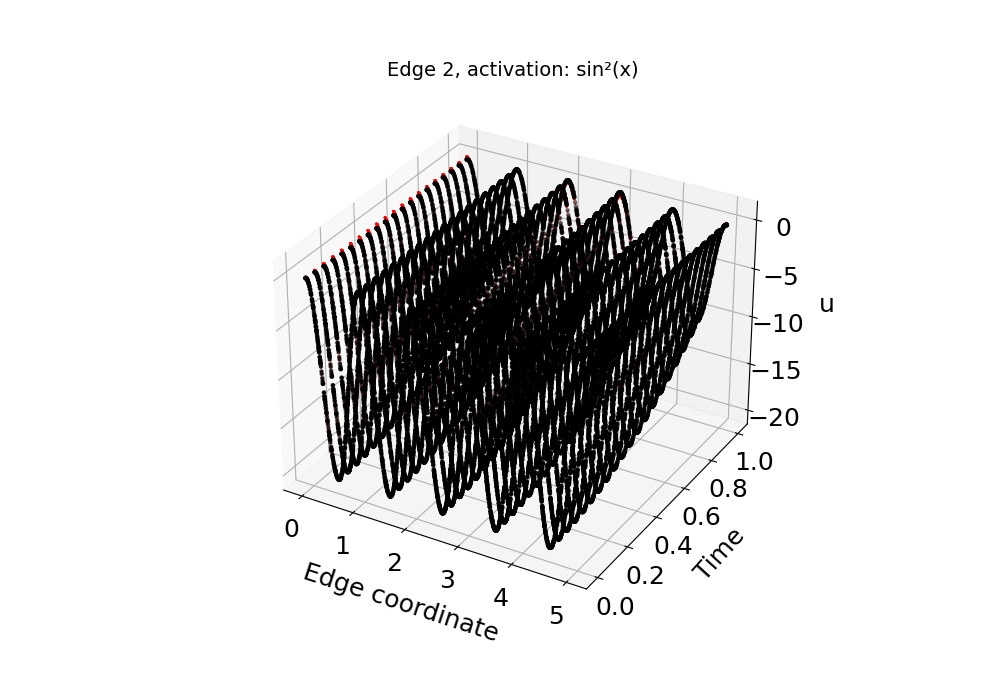}}
\vspace{0.4cm}
\caption{Validation error (top figure) and comparison on each edge (figures on the second line) of the exact solution (red) and PINN approximated solution (black) of the non-linear hyperbolic problem on the star graph with three edges}
\label{star3_hyperbolic}
\end{figure}

\begin{figure}
{\hspace{5cm}\includegraphics[width=0.35\textwidth]{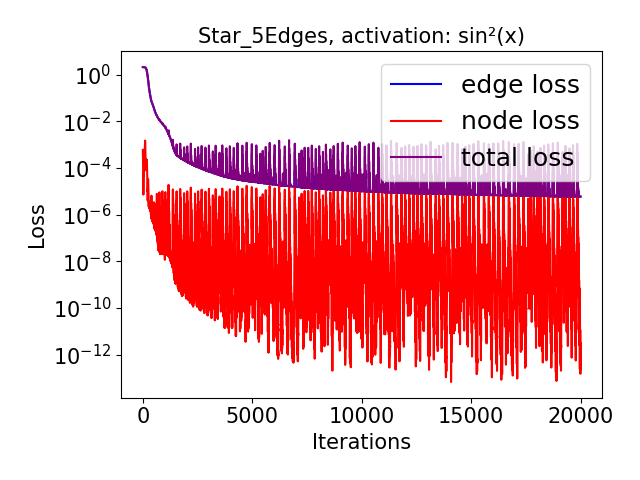}}\\
\includegraphics[width=0.33\textwidth]{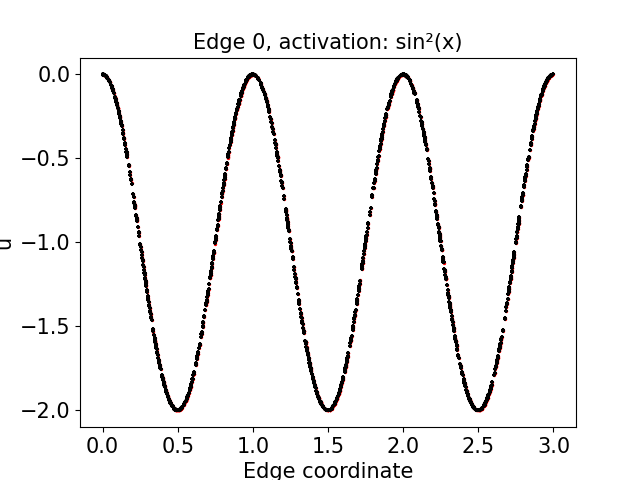}
\includegraphics[width=0.33\textwidth]{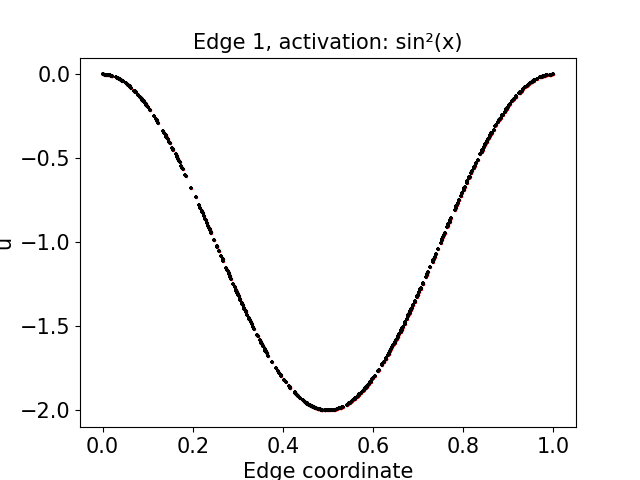}
\includegraphics[width=0.33\textwidth]{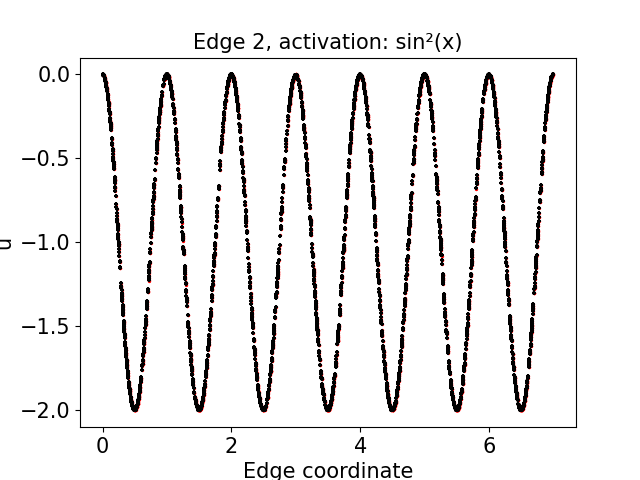}
\includegraphics[width=0.33\textwidth]{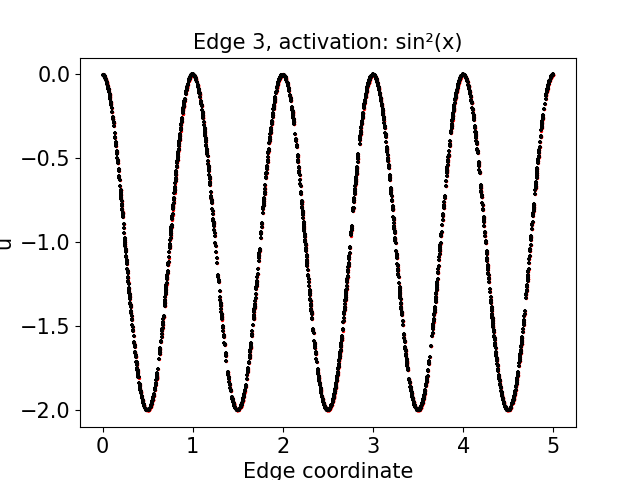}
\includegraphics[width=0.33\textwidth]{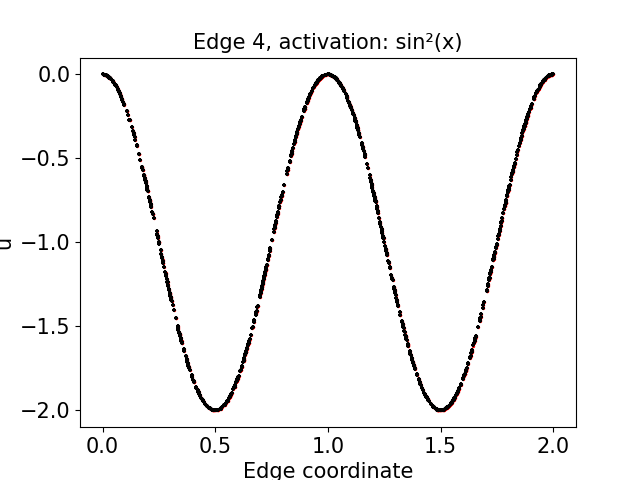}
\vspace{0.4cm}
\caption{Validation error (top figure) and comparison on each edge (figures on the second line) of the exact solution (red) and PINN approximated solution (black) of the non-linear elliptic problem on the star graph with five edges}
\label{star5_elliptic}
\end{figure}

\begin{figure}[htbp!]
{\hspace{5cm}\includegraphics[width=0.35\textwidth]{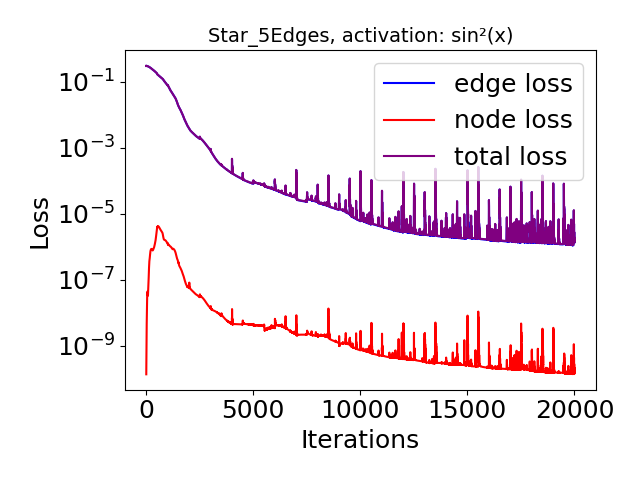}}\\
\vspace{-0.5cm}
\adjustbox{trim={.25\width} {0\height} {.15\width} {0\height},clip}{\includegraphics[width=0.55\textwidth]{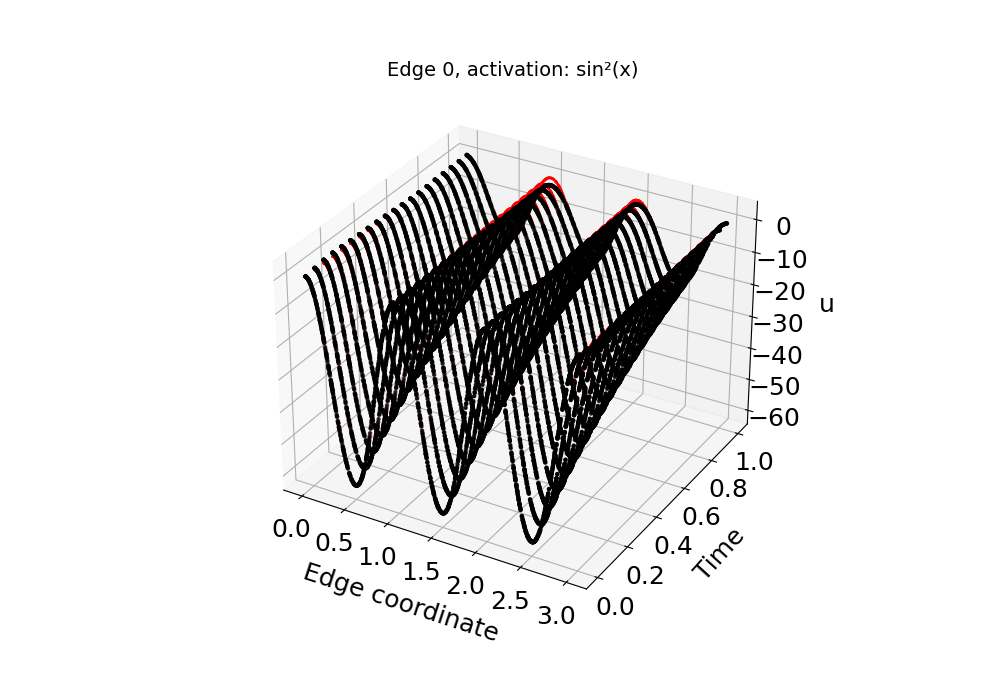}}
\adjustbox{trim={.25\width} {0\height} {.15\width} {0\height},clip}{\includegraphics[width=0.55\textwidth]{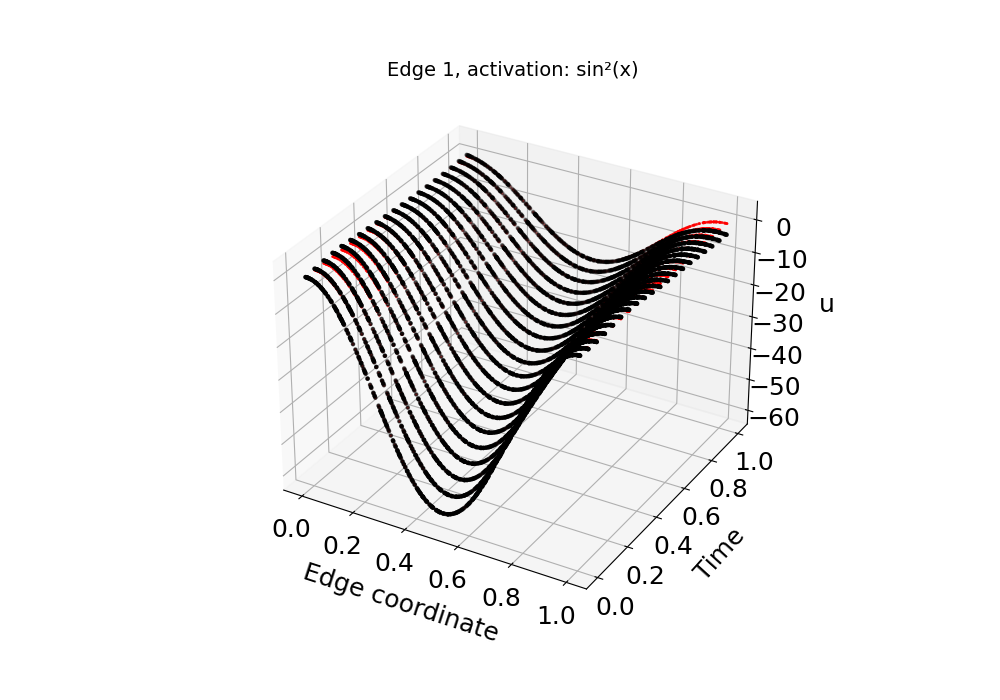}}
\adjustbox{trim={.25\width} {0\height} {.15\width} {0\height},clip}{\includegraphics[width=0.55\textwidth]{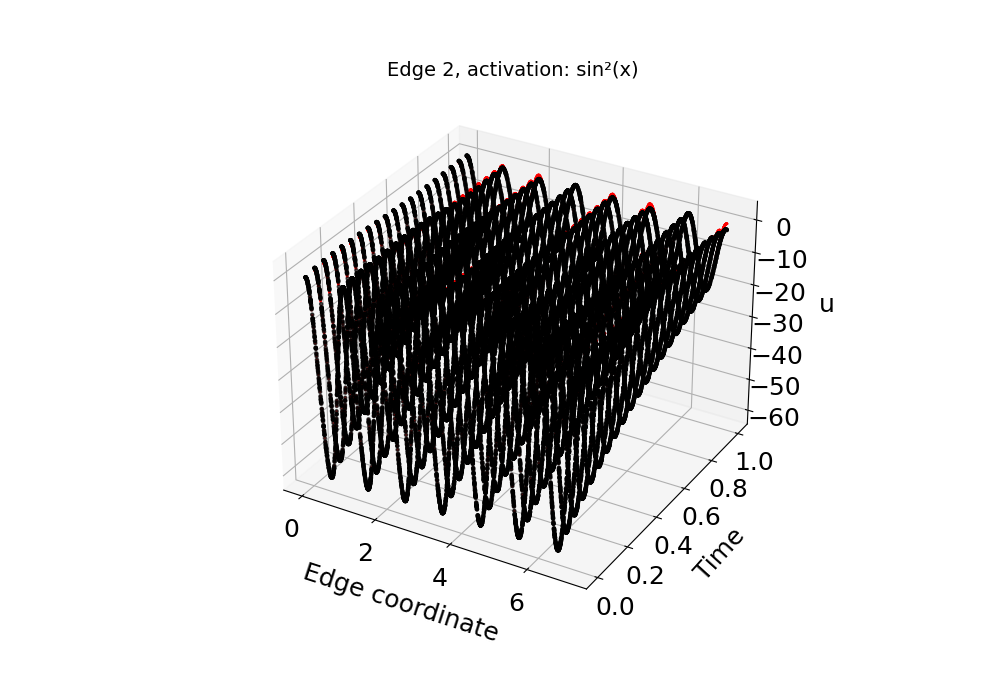}}
\vspace{-0.5cm}
\adjustbox{trim={.25\width} {0\height} {.15\width} {0\height},clip}{\includegraphics[width=0.55\textwidth]{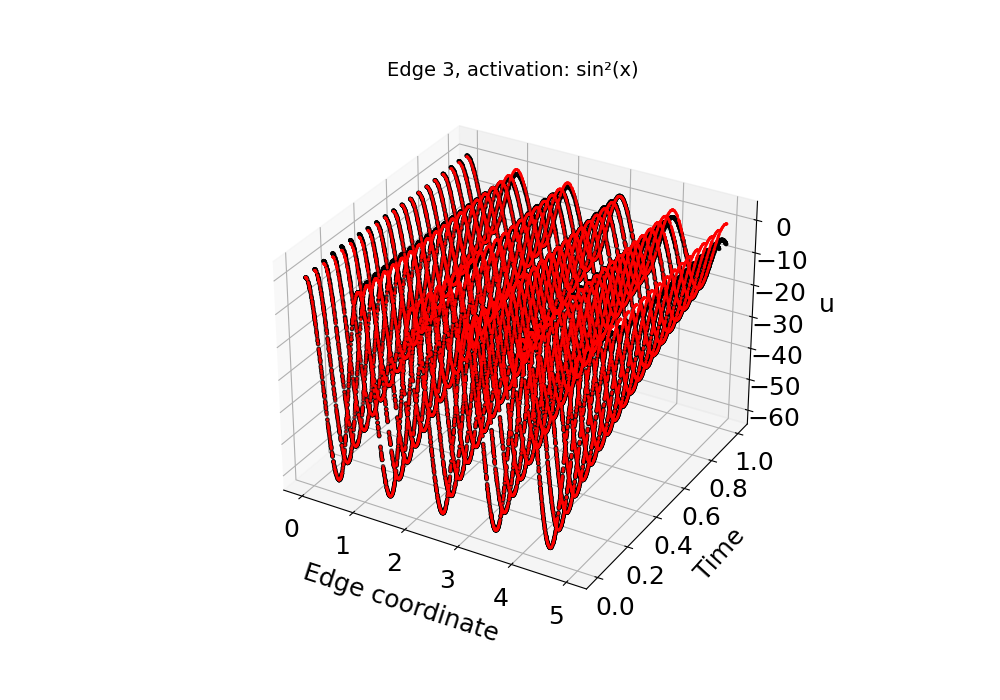}}
\adjustbox{trim={.25\width} {0\height} {.15\width} {0\height},clip}{\includegraphics[width=0.55\textwidth]{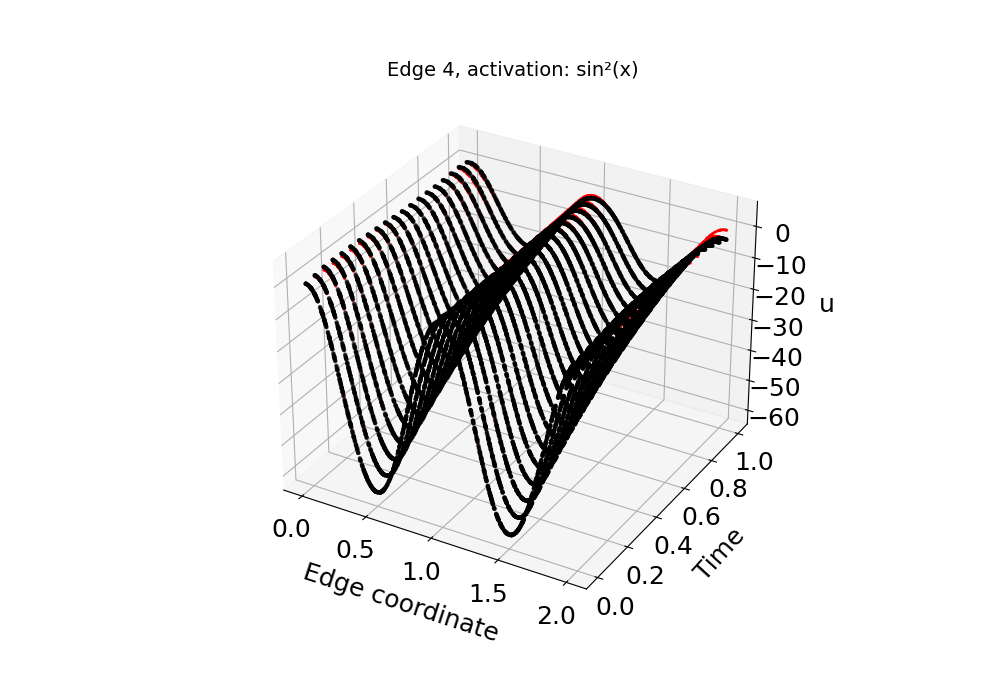}}
\caption{Validation error (top figure) and comparison on each edge (figures on the second and third line) of the exact solution (red) and PINN approximated solution (black) of the non-linear parabolic problem on the star graph with five edges}
\label{star5_parabolic}
\end{figure}

\begin{figure}[htbp!]
{\hspace{5cm}\includegraphics[width=0.35\textwidth]{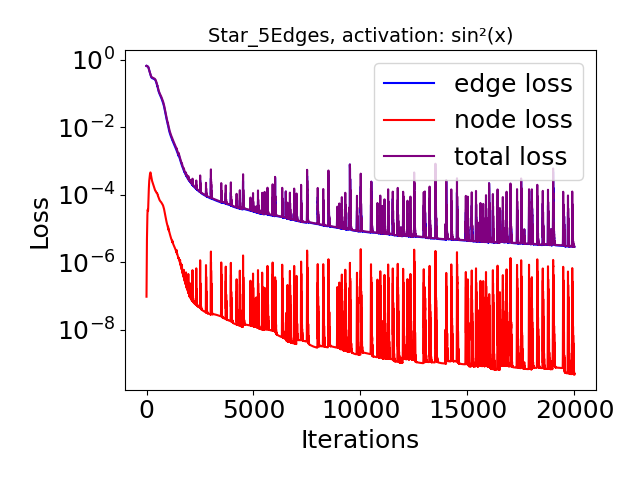}}\\
\vspace{-0.5cm}
\adjustbox{trim={.25\width} {0\height} {.15\width} {0\height},clip}{\includegraphics[width=0.55\textwidth]{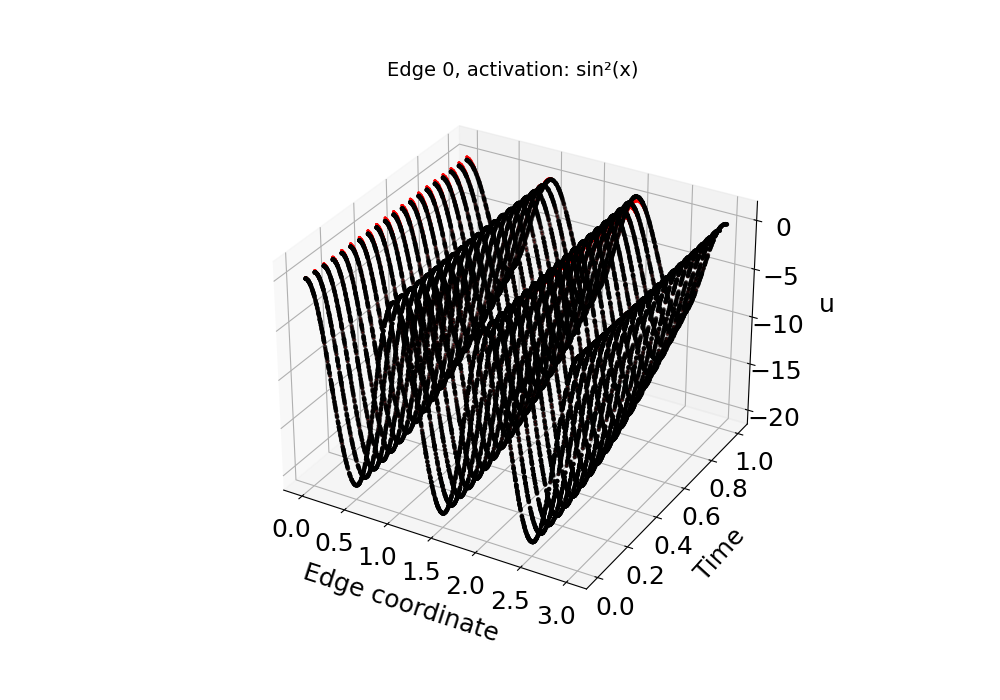}}
\adjustbox{trim={.25\width} {0\height} {.15\width} {0\height},clip}{\includegraphics[width=0.55\textwidth]{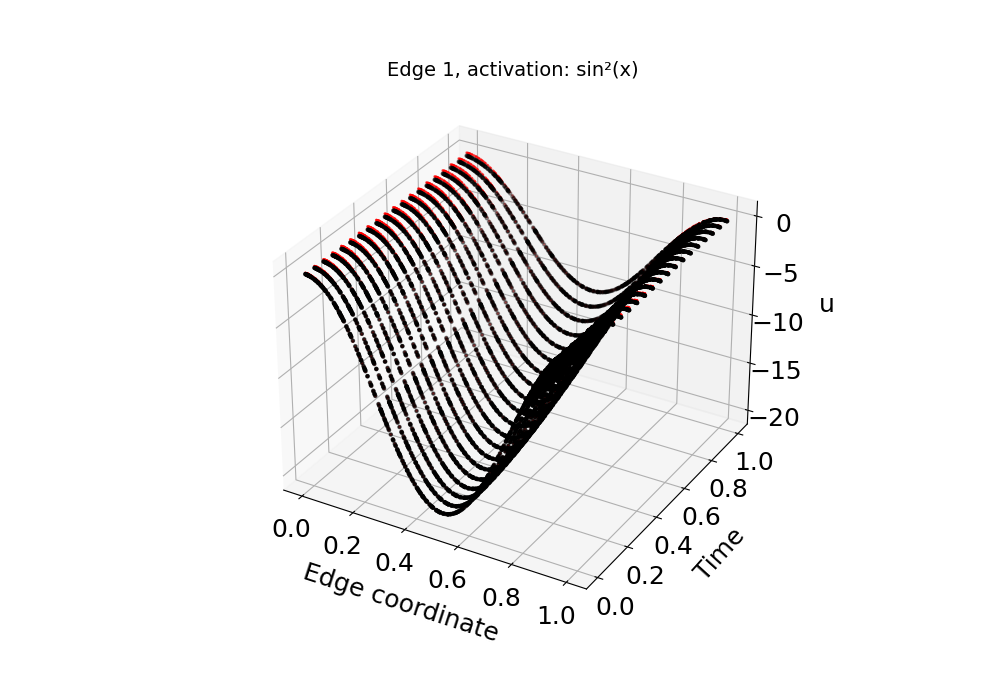}}
\adjustbox{trim={.25\width} {0\height} {.15\width} {0\height},clip}{\includegraphics[width=0.55\textwidth]{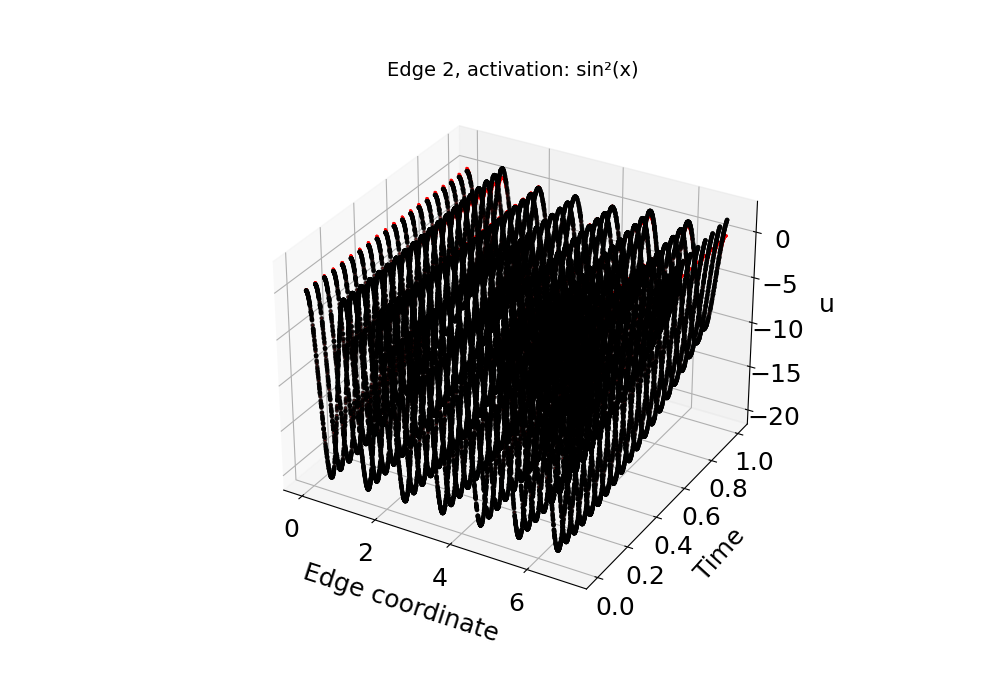}}
\vspace{-0.5cm}
\adjustbox{trim={.25\width} {0\height} {.15\width} {0\height},clip}{\includegraphics[width=0.55\textwidth]{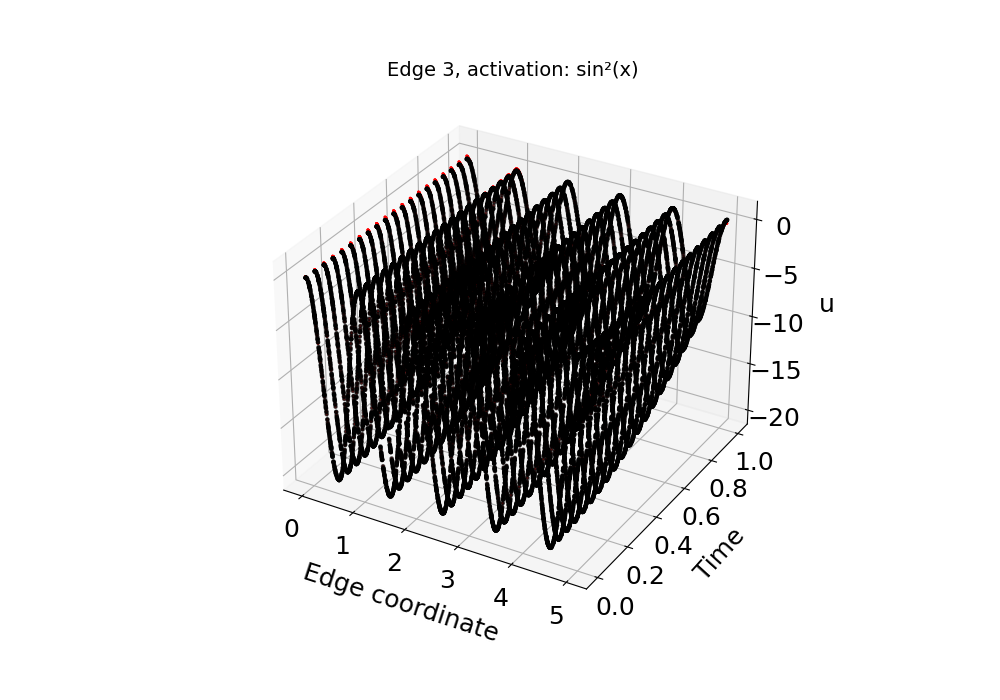}}
\adjustbox{trim={.25\width} {0\height} {.15\width} {0\height},clip}{\includegraphics[width=0.55\textwidth]{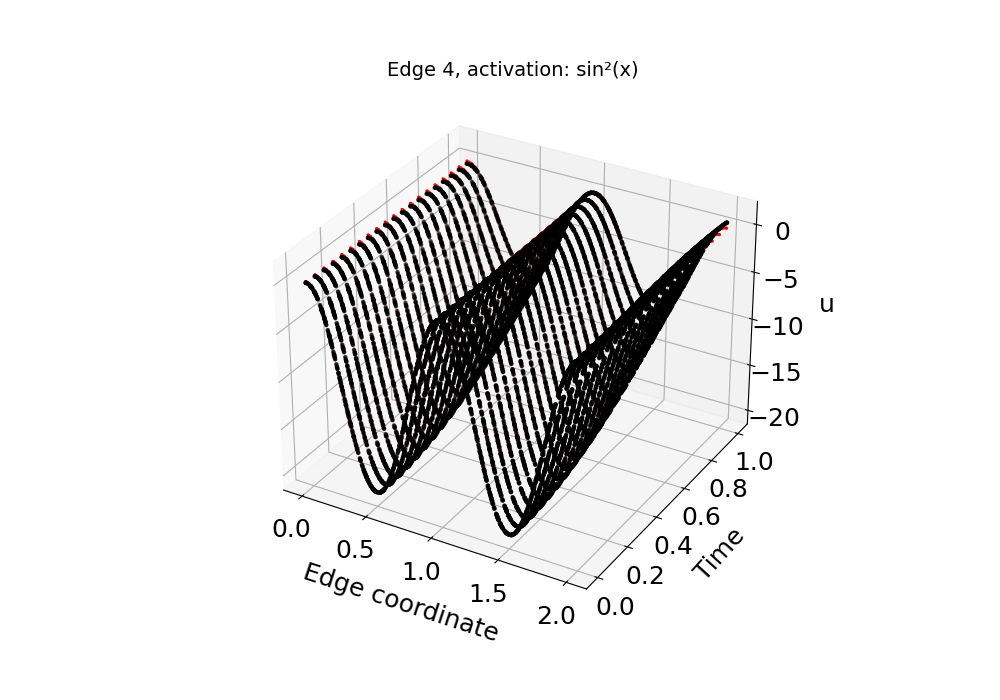}}
\caption{Validation error (top figure) and comparison on each edge (figures on the second and third line) of the exact solution (red) and PINN approximated solution (black) of the non-linear hyperbolic problem on the star graph with five edges}
\label{star5_hyperbolic}
\end{figure}

\begin{figure}
{\hspace{5cm}\includegraphics[width=0.35\textwidth]{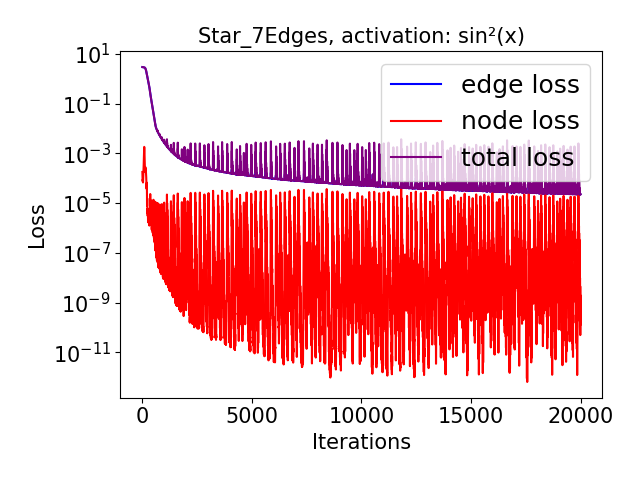}}\\
\includegraphics[width=0.33\textwidth]{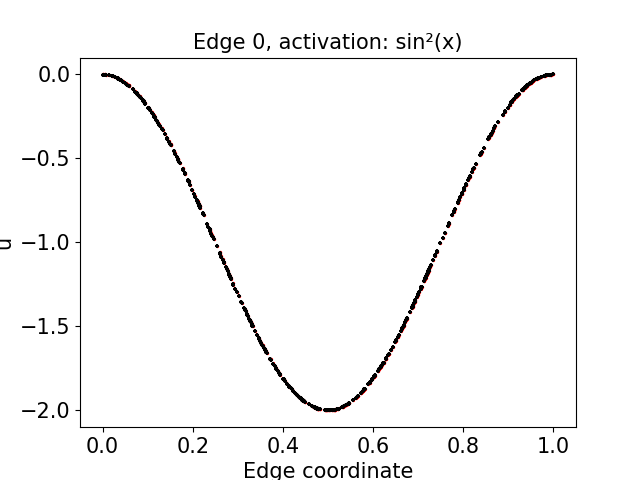}
\includegraphics[width=0.33\textwidth]{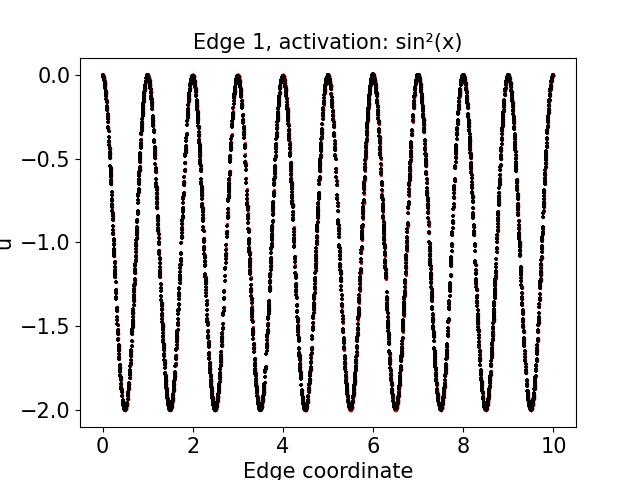}
\includegraphics[width=0.33\textwidth]{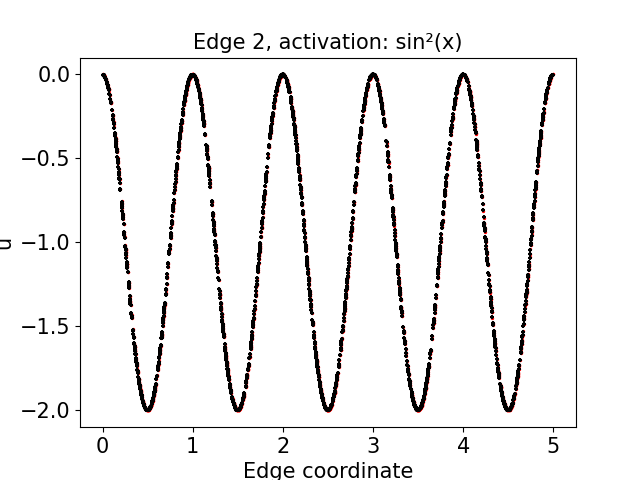}
\includegraphics[width=0.33\textwidth]{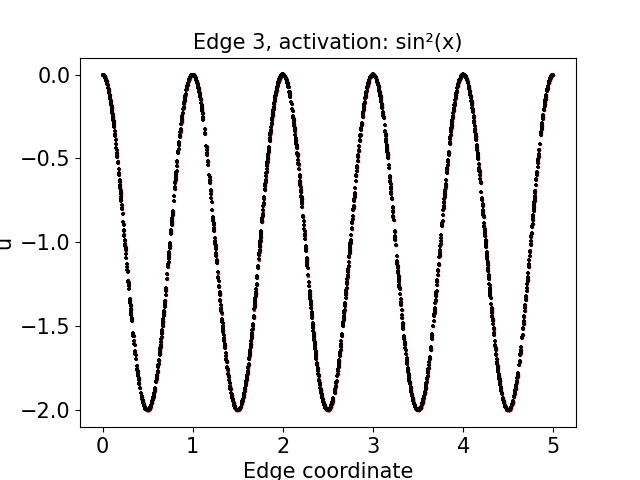}
\includegraphics[width=0.33\textwidth]{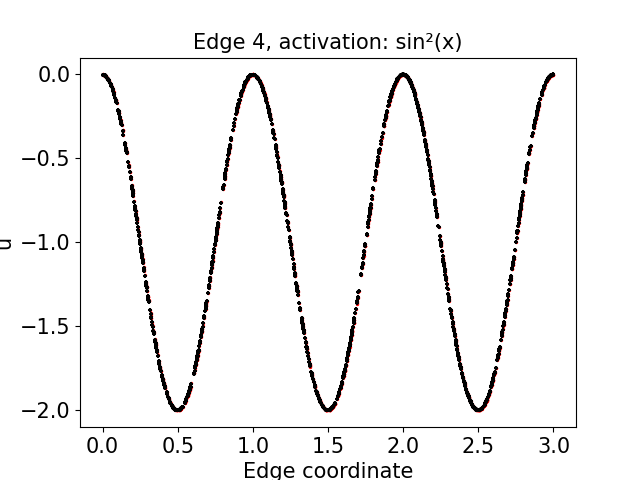}
\includegraphics[width=0.33\textwidth]{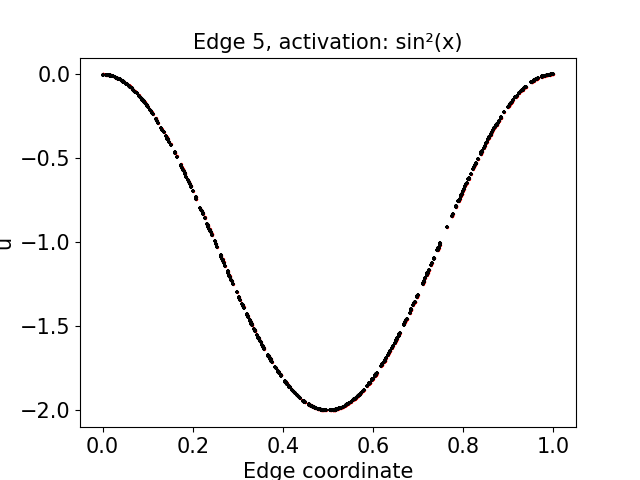}
\includegraphics[width=0.33\textwidth]{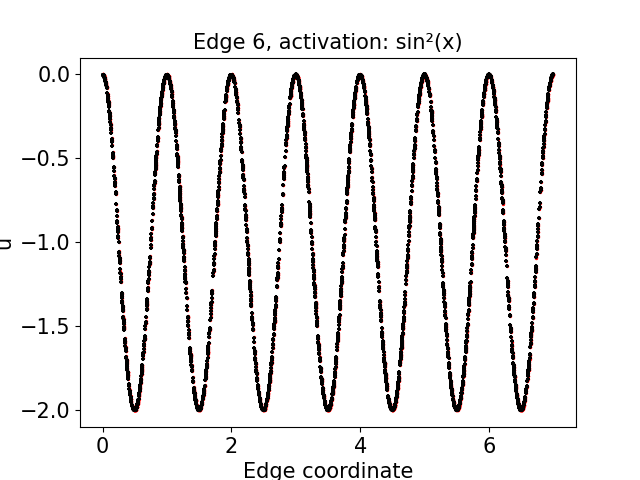}
\caption{Validation error (top figure) and comparison on each edge (figures on the second, third, and fourth line) of the exact solution (red) and PINN approximated solution (black) of the non-linear elliptic problem on the star graph with seven edges}
\label{star7_elliptic}
\end{figure}

\begin{figure}[htbp!]
{\hspace{5cm}\includegraphics[width=0.35\textwidth]{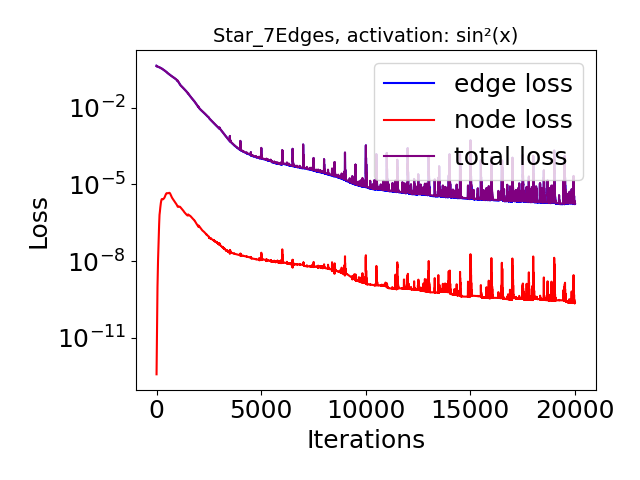}}\\
\vspace{-0.5cm}
\adjustbox{trim={.25\width} {0\height} {.15\width} {0\height},clip}{\includegraphics[width=0.55\textwidth]{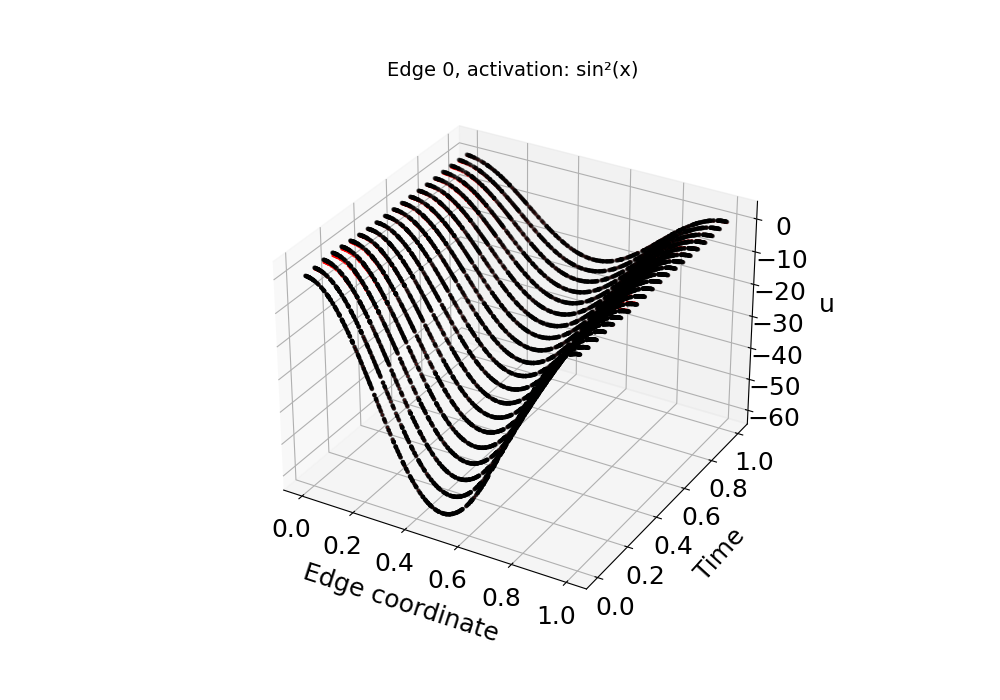}}
\adjustbox{trim={.25\width} {0\height} {.15\width} {0\height},clip}{\includegraphics[width=0.55\textwidth]{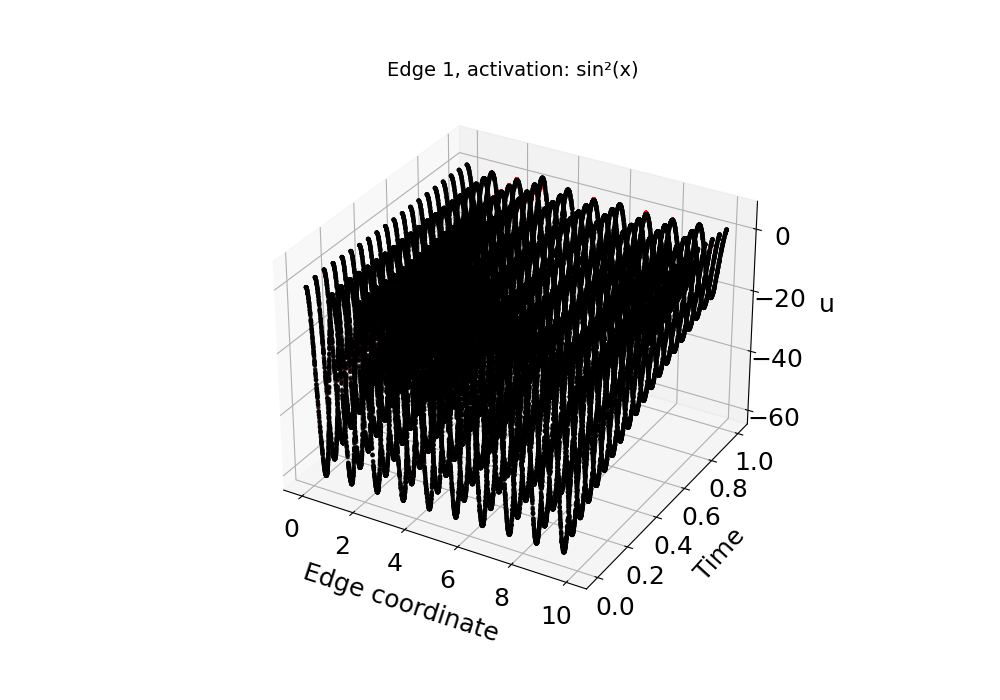}}
\adjustbox{trim={.25\width} {0\height} {.15\width} {0\height},clip}{\includegraphics[width=0.55\textwidth]{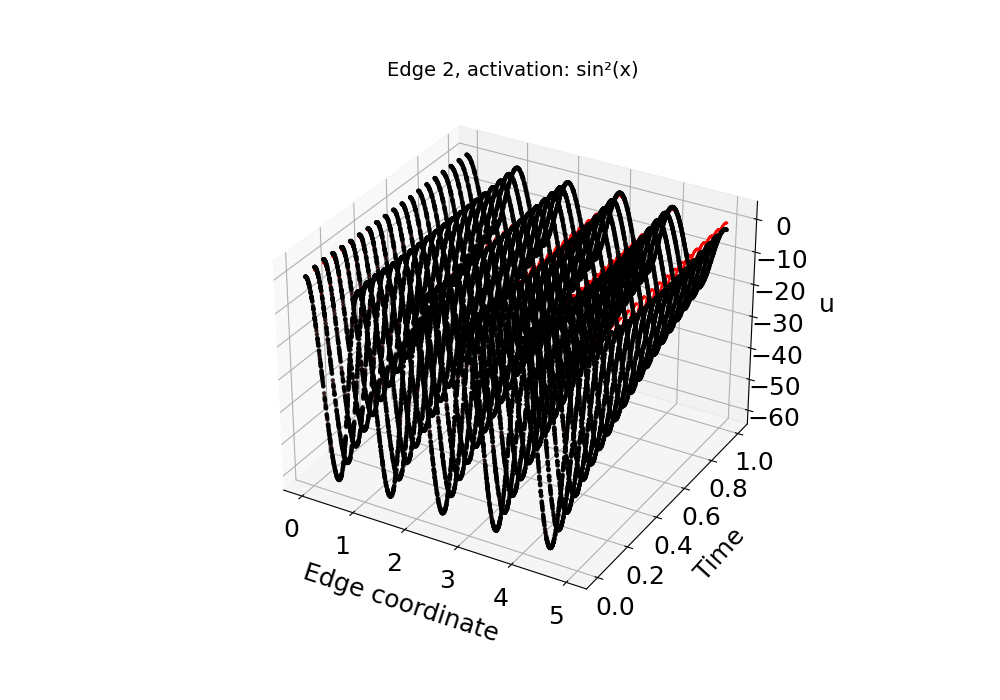}}
\vspace{-0.5cm}
\adjustbox{trim={.25\width} {0\height} {.15\width} {0\height},clip}{\includegraphics[width=0.55\textwidth]{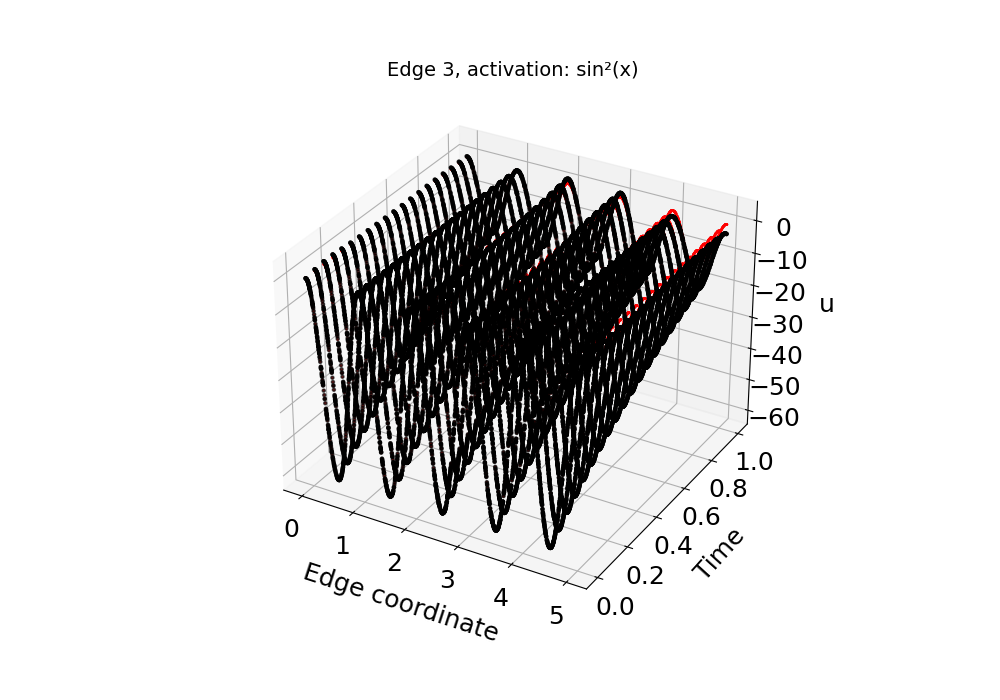}}
\adjustbox{trim={.25\width} {0\height} {.15\width} {0\height},clip}{\includegraphics[width=0.55\textwidth]{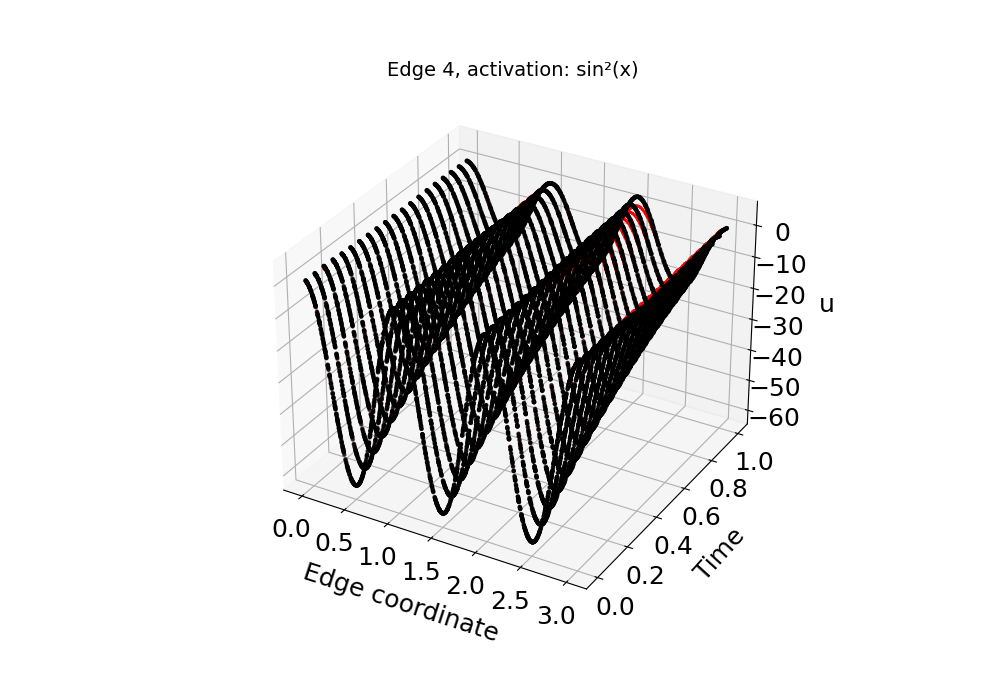}}
\adjustbox{trim={.25\width} {0\height} {.15\width} {0\height},clip}{\includegraphics[width=0.55\textwidth]{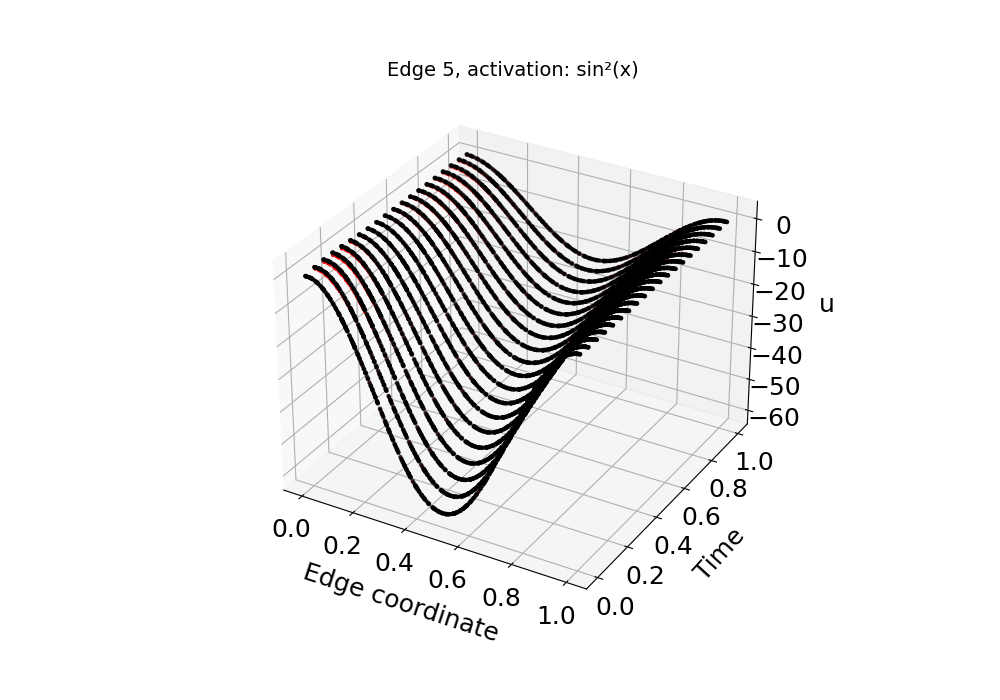}}
\vspace{-0.5cm}
\adjustbox{trim={.25\width} {0\height} {.15\width} {0\height},clip}{\includegraphics[width=0.55\textwidth]{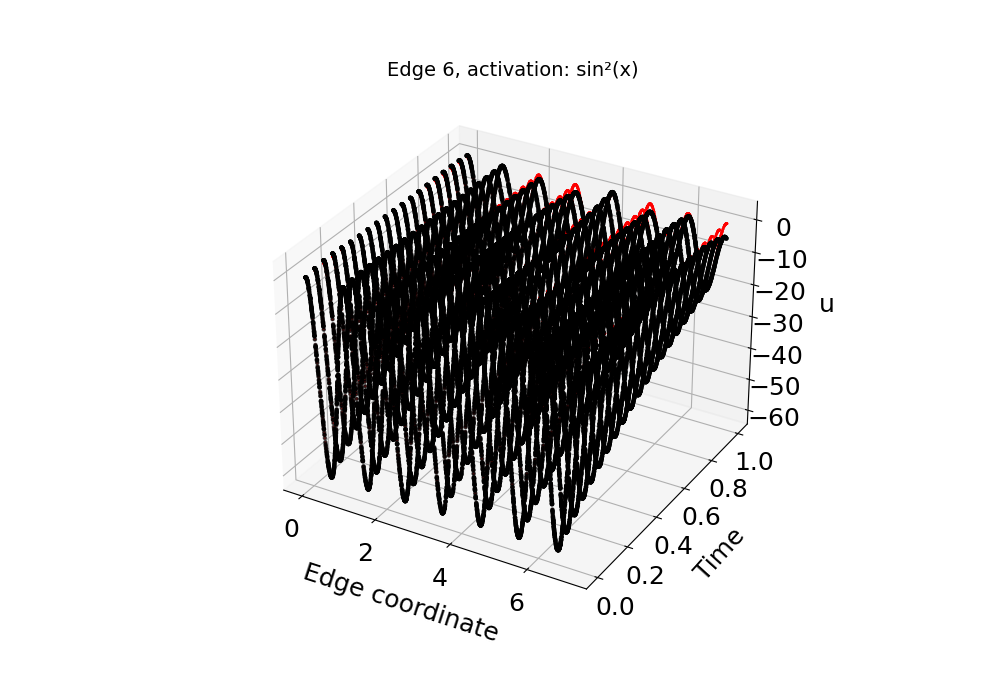}}
\caption{Validation error (top figure) and comparison on each edge (figures on the second, third, and fourth line) of the exact solution (red) and PINN approximated solution (black) of the non-linear parabolic problem on the star graph with seven edges}
\label{star7_parabolic}
\end{figure}

\begin{figure}[htbp!]
{\hspace{5cm}\includegraphics[width=0.35\textwidth]{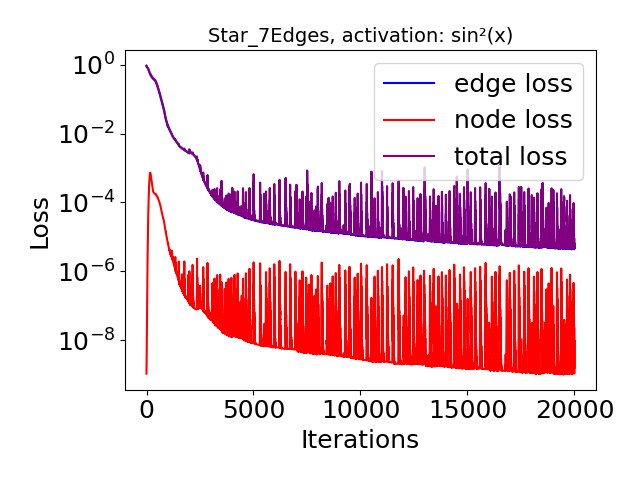}}\\
\vspace{-0.5cm}
\adjustbox{trim={.25\width} {0\height} {.15\width} {0\height},clip}{\includegraphics[width=0.55\textwidth]{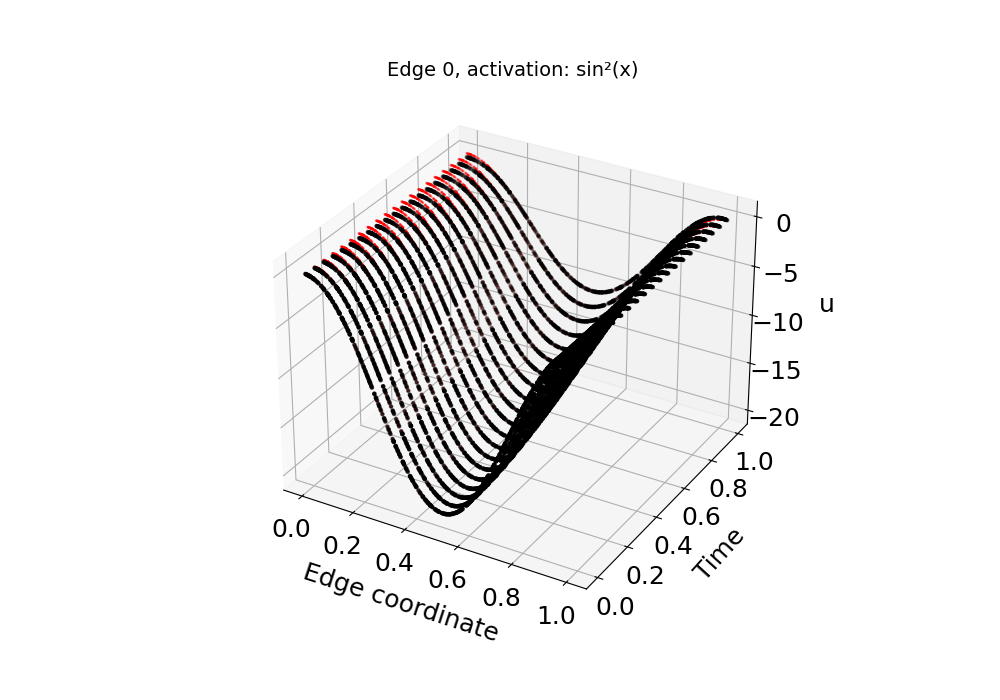}}
\adjustbox{trim={.25\width} {0\height} {.15\width} {0\height},clip}{\includegraphics[width=0.55\textwidth]{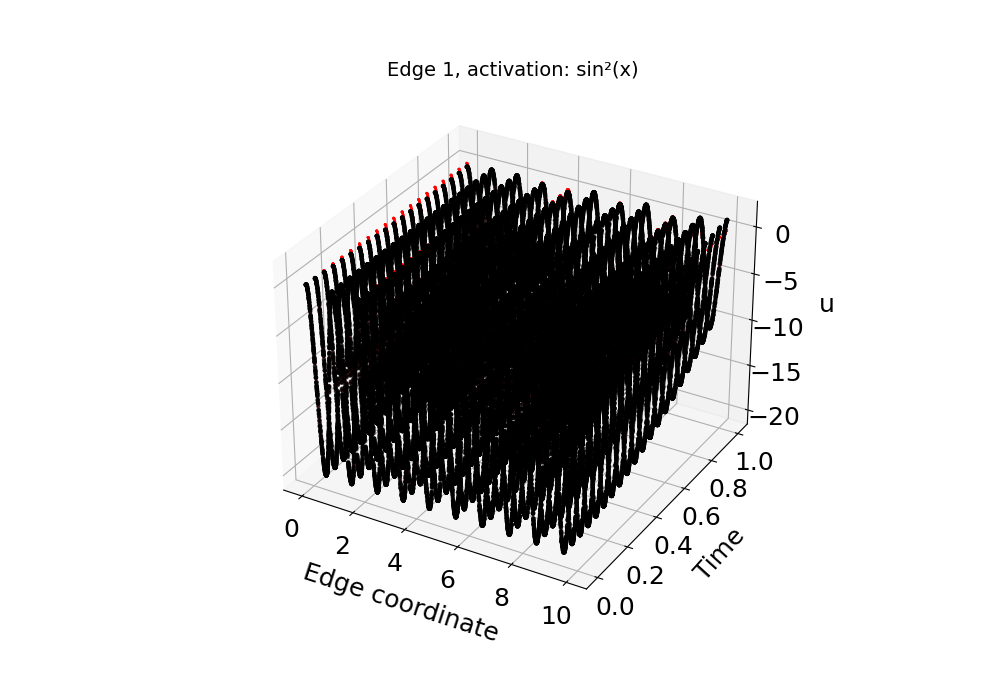}}
\adjustbox{trim={.25\width} {0\height} {.15\width} {0\height},clip}{\includegraphics[width=0.55\textwidth]{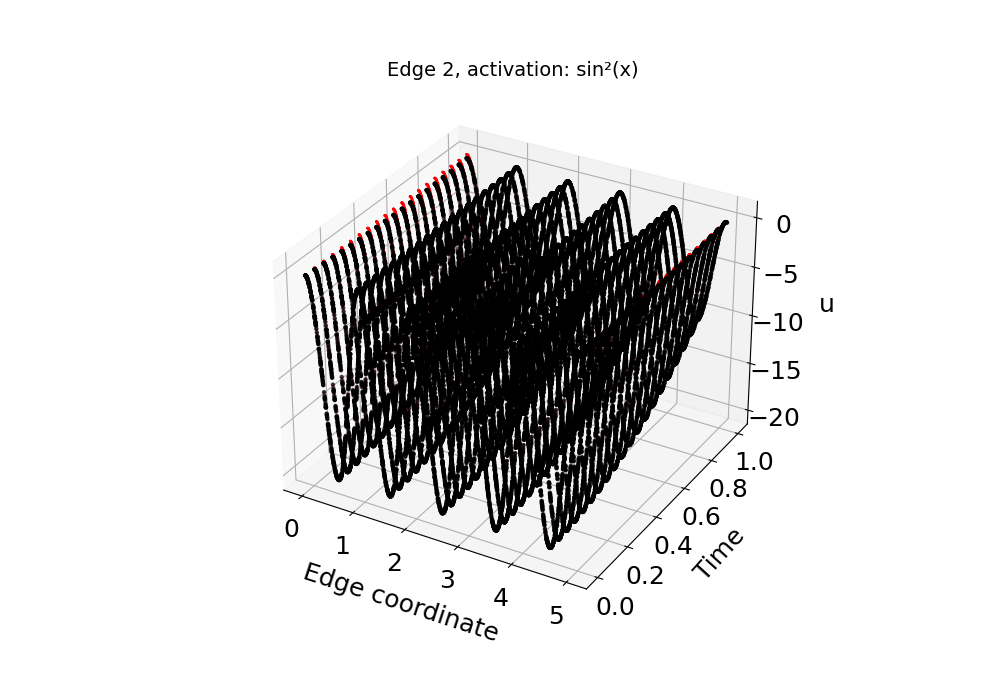}}
\vspace{-0.5cm}
\adjustbox{trim={.25\width} {0\height} {.15\width} {0\height},clip}{\includegraphics[width=0.55\textwidth]{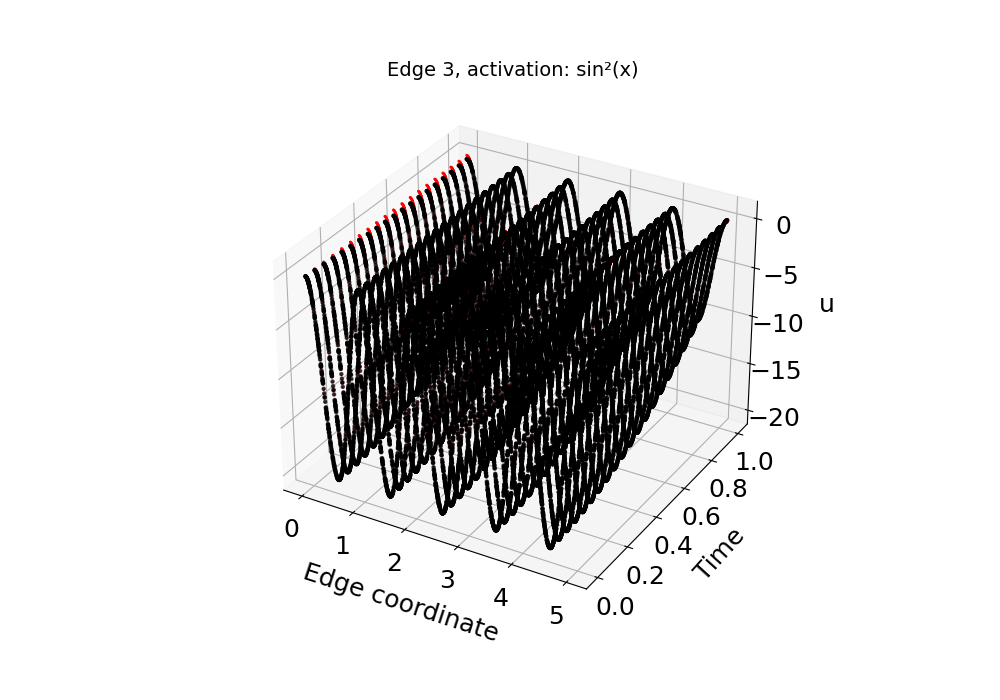}}
\adjustbox{trim={.25\width} {0\height} {.15\width} {0\height},clip}{\includegraphics[width=0.55\textwidth]{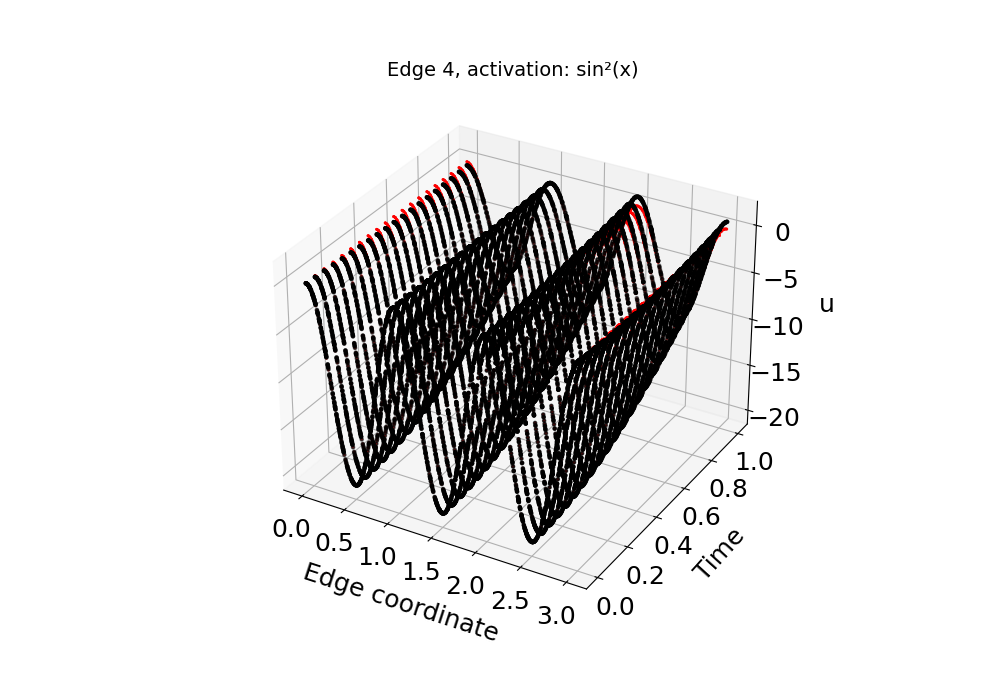}}
\adjustbox{trim={.25\width} {0\height} {.15\width} {0\height},clip}{\includegraphics[width=0.55\textwidth]{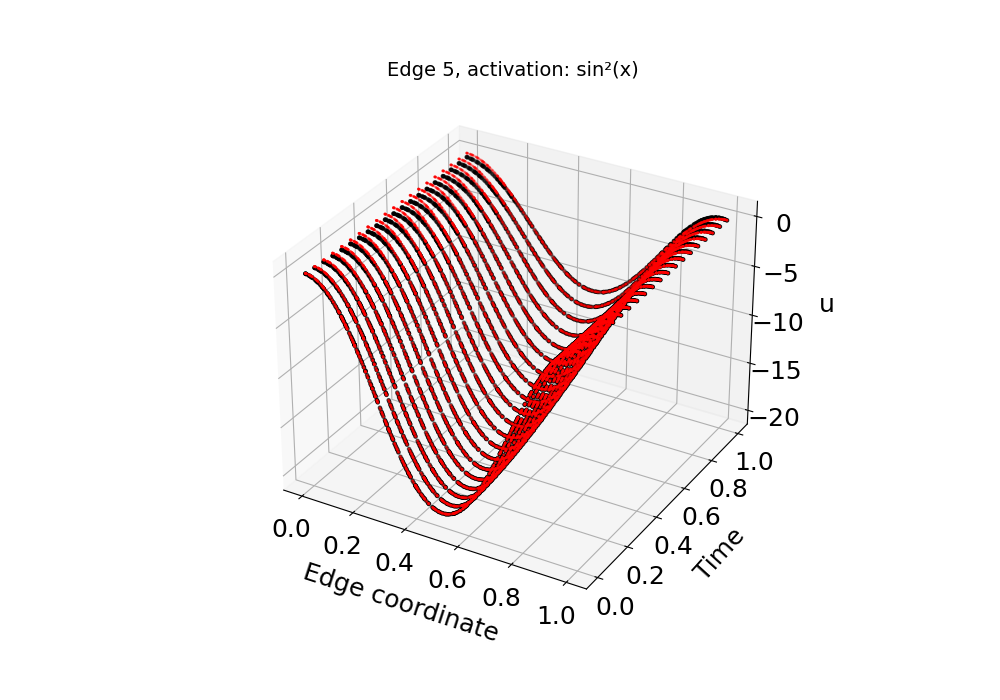}}
\vspace{-0.5cm}
\adjustbox{trim={.25\width} {0\height} {.15\width} {0\height},clip}{\includegraphics[width=0.55\textwidth]{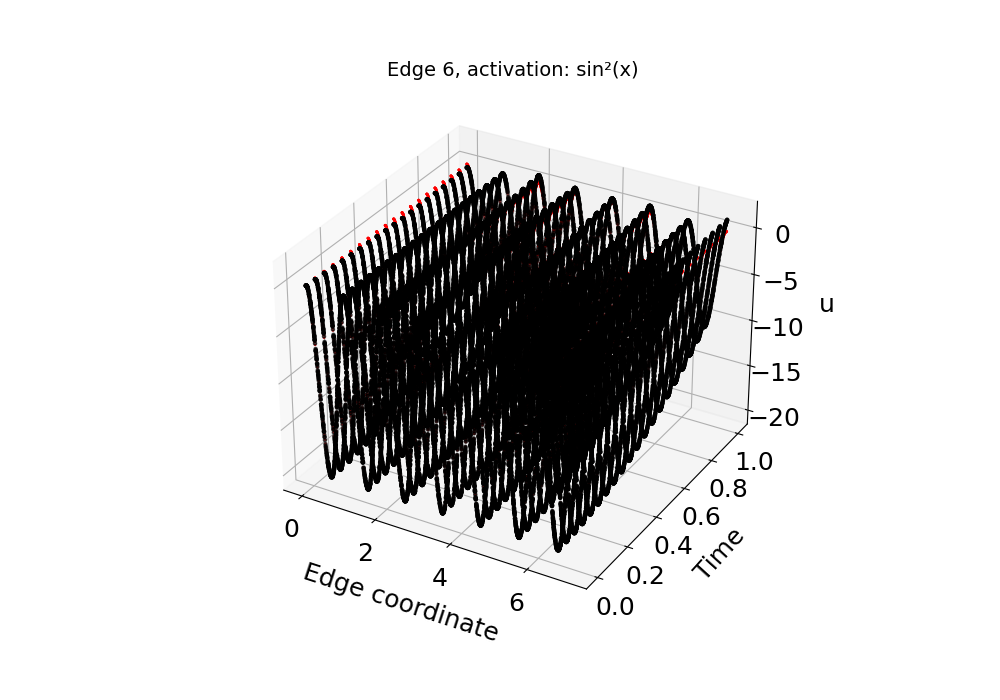}}
\caption{Validation error (top figure) and comparison on each edge (figures on the second, third, and fourth line) of the exact solution (red) and PINN approximated solution (black) of the non-linear hyperbolic problem on the star graph with seven edges}
\label{star7_hyperbolic}
\end{figure}

In Figures \ref{star3_elliptic}, \ref{star3_parabolic}, and \ref{star3_hyperbolic} we compare the exact solution and the PINN solution on each edge for the non-linear elliptic problem, non-linear parabolic problem, and non-linear hyperbolic problem on the star graph with three edges. 
In Figures \ref{star5_elliptic}, \ref{star5_parabolic}, and \ref{star5_hyperbolic} we compare the exact solution and the PINN solution on each edge for the non-linear elliptic problem, non-linear parabolic problem, and non-linear hyperbolic problem on the star graph with five edges. 
In Figures \ref{star7_elliptic}, \ref{star7_parabolic}, and \ref{star7_hyperbolic} we compare the exact solution and the PINN solution on each edge for the non-linear elliptic problem, non-linear parabolic problem, and non-linear hyperbolic problem on the star graph with seven edges. 
We notice that the PINN solutions match well with the exact solution on each edge even when the number of edges incidental to the central node increases. 
These results confirm the robustness of our DL approach in handling graphs where the degrees of the nodes varies across a significant range.

\begin{table}[!htbp] 
\centering
\begin{tabular}{|l|*{3}{|c|}}
\hline
\backslashbox{Type of problem}{activation func.}
&\makebox[3cm]{Star with three edges}&\makebox[3cm]{Star with five edges}&\makebox[3cm]{Star with seven edges}
\\\hline\hline
Non-linear elliptic problem &  $\lambda =51$ & $\lambda =74$, & $\lambda =99$
\\ 
& err= $7.738318e-04$ &  err=$1.979e-04$ & err=$9.329e-04$  \\
\hline
Non-linear parabolic problem & $\lambda = 201$
 & $\lambda=231$ & $\lambda=350$ \\
 & err=$2.3542e-01$ & err=$2.472e-01$  &  err=$2.133e-01$ \\
\hline Non-linear hyperbolic problem & $\lambda=101$ &
$\lambda=122$ & $\lambda=168$\\
 & err=$4.519e-02$ & err=$5.378e-02$ &err=$6.595e-02$ \\\hline
\end{tabular}
\caption{Final $\lambda$ and validation error on star graphs.} \label{lambdaerror_stargraph}
\end{table}
%%%%%%%%%%%%%%%%%%%%%%%%%%%%%%%%%%
%%%%%%%%%%%%%%%%%%%%%%%%%%%%%%%%%%%

%%%%%%%%%%%%%%%%%%%%

%%%%%%%%%%%%%%%%%%%%

%%%%%%%%%%%%%%%%%%%%%%%%%%%%%%%%%%%

%%%%%%%%%%%%%%%%%%%%
%%%%%%%%%%%%%%%%%%%%%%

%%%%%%%%%%%%%%%%%%%%

%%%%%%%%%%%%%%%%%%%%

\subsection{Complex metric graphs}
\label{general_graph_results}
\begin{figure}[ht]
\centering
	   \includegraphics[width=0.5\textwidth]{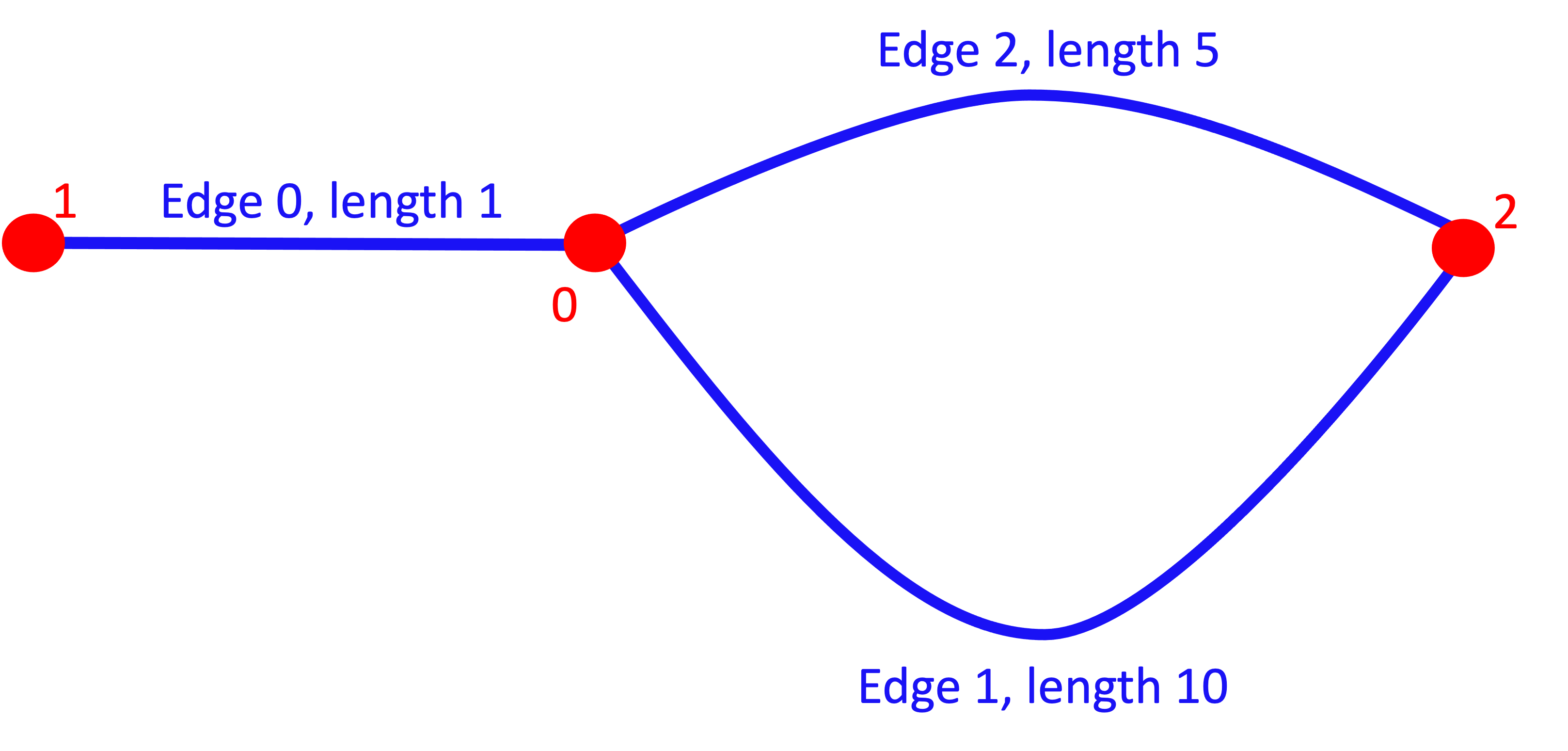}
	   \caption{Lasso graph.}
\label{lasso_graph}	   
\end{figure}

\begin{figure}[ht]
\centering
	   \includegraphics[width=0.6\textwidth]{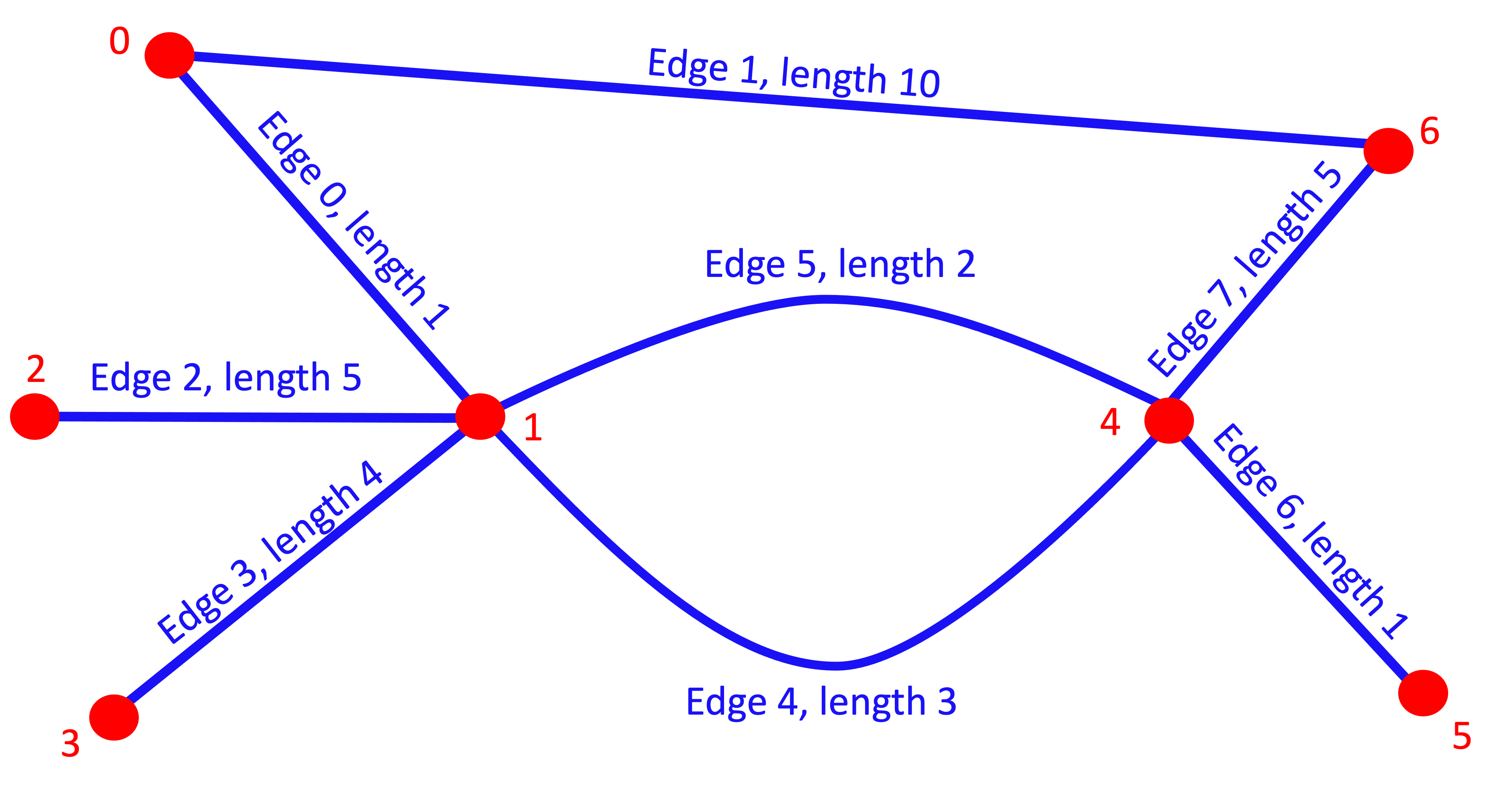}
\caption{General graph.}
\label{general_graph}
\end{figure}

To assess the performance of our DL method in handling metric graphs with different topologies, in this section we describe numerical results obtained by solving the differential problems in Equation \eqref{elliptic}, \eqref{parabolic}, and \eqref{hyperbolic} defined on the two metric graphs whose topology is illustrated in Figures \ref{lasso_graph} and \ref{general_graph}. 
According to the denomination introduced in \cite{Kurasov:13}, from now on we refer to the graph illustrated in Figure \ref{lasso_graph} as Lasso graph. For convenience, we will refer from now on to the graph illustrated in Figure \ref{general_graph} as general graph.

\begin{figure}[htbp!]
{\hspace{5cm}\includegraphics[width=0.35\textwidth]{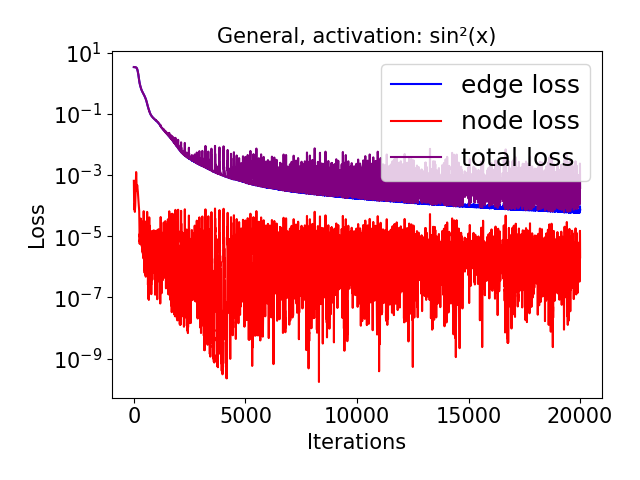}}\\
\includegraphics[width=0.33\textwidth]{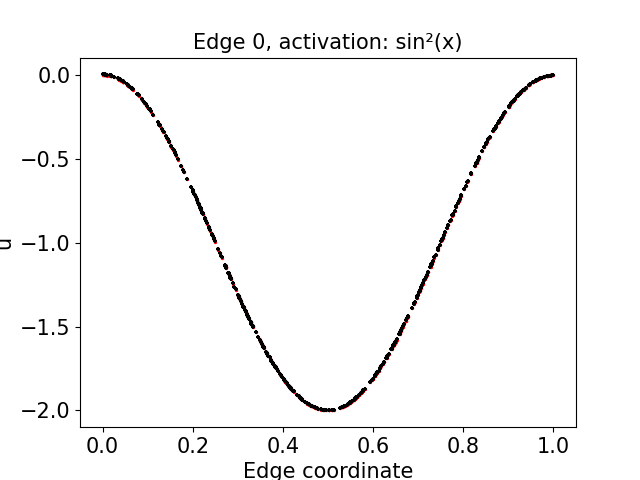}
\includegraphics[width=0.33\textwidth]{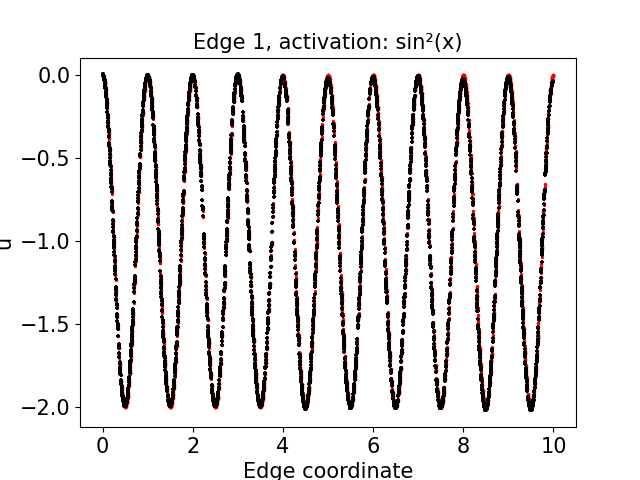}
\includegraphics[width=0.33\textwidth]{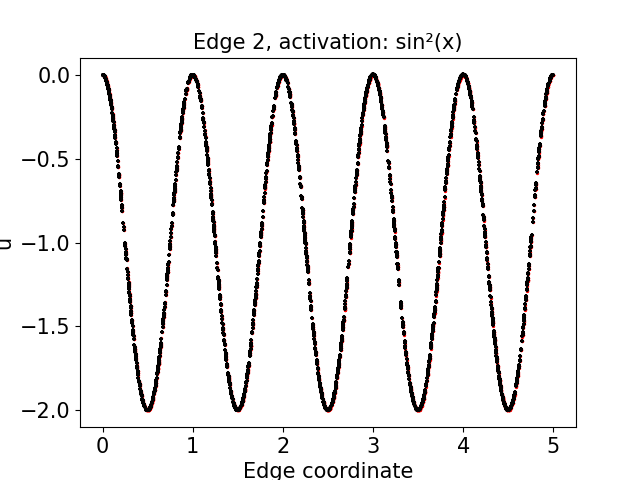}
\includegraphics[width=0.33\textwidth]{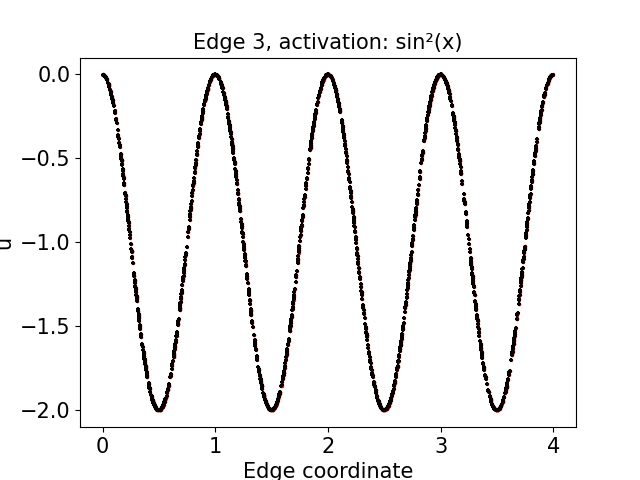}
\includegraphics[width=0.33\textwidth]{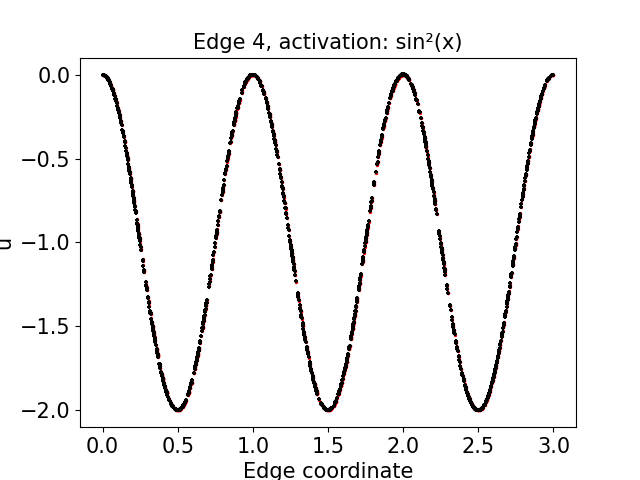}
\includegraphics[width=0.33\textwidth]{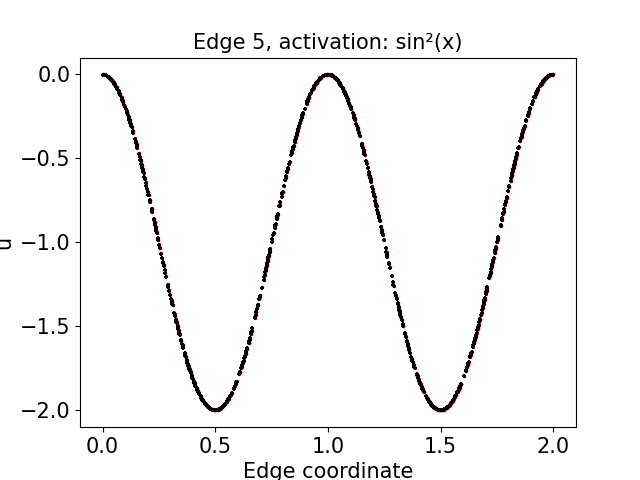}
\includegraphics[width=0.33\textwidth]{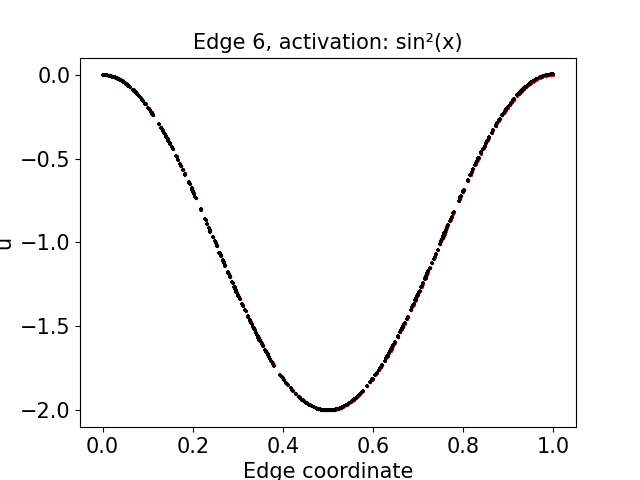}
\includegraphics[width=0.33\textwidth]{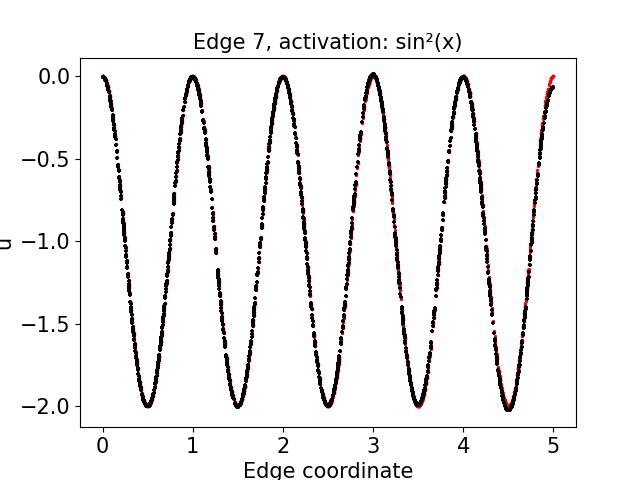}\\
\caption{Validation error (top figure) and comparison on each edge (figures on the second, third, and fourth line) of the exact solution (red) and PINN approximated solution (black) of the non-linear elliptic problem defined on the general graph.}
\label{general_graph_elliptic}
\end{figure}

\begin{figure}[htbp!]
{\hspace{5cm}\includegraphics[width=0.35\textwidth]{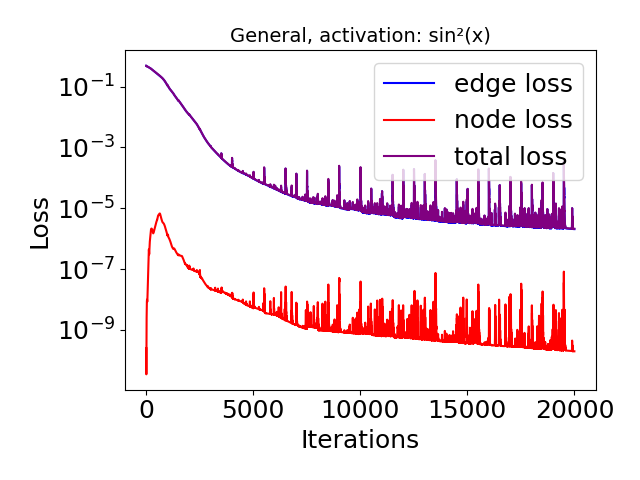} }\\
\vspace{-0.5cm}
\adjustbox{trim={.25\width} {0\height} {.15\width} {0\height},clip}{\includegraphics[width=0.55\textwidth]{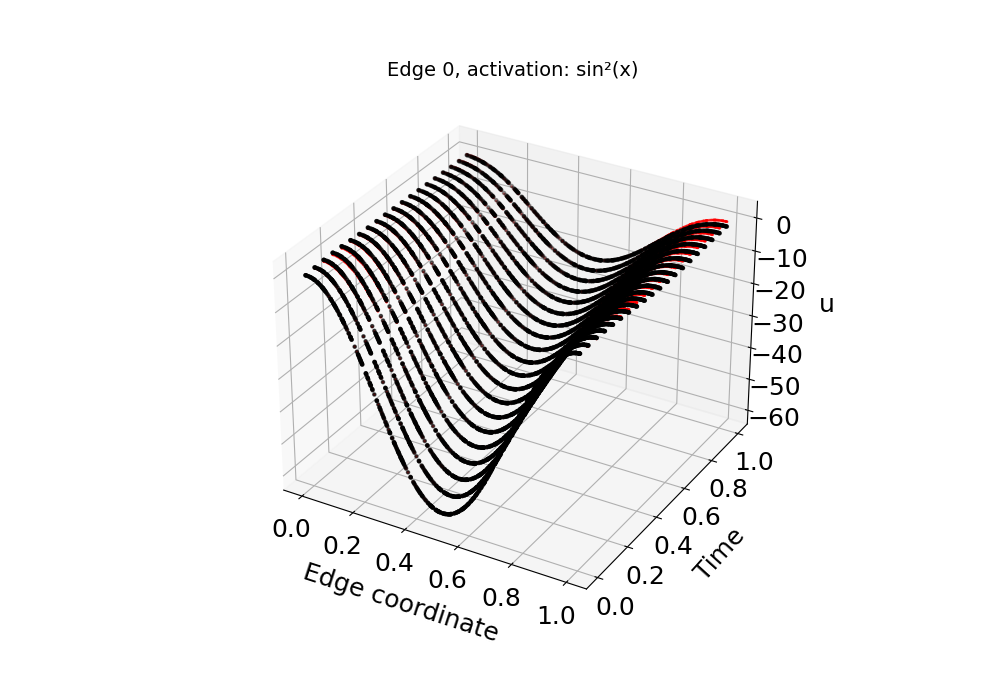}}
\adjustbox{trim={.25\width} {0\height} {.15\width} {0\height},clip}{\includegraphics[width=0.55\textwidth]{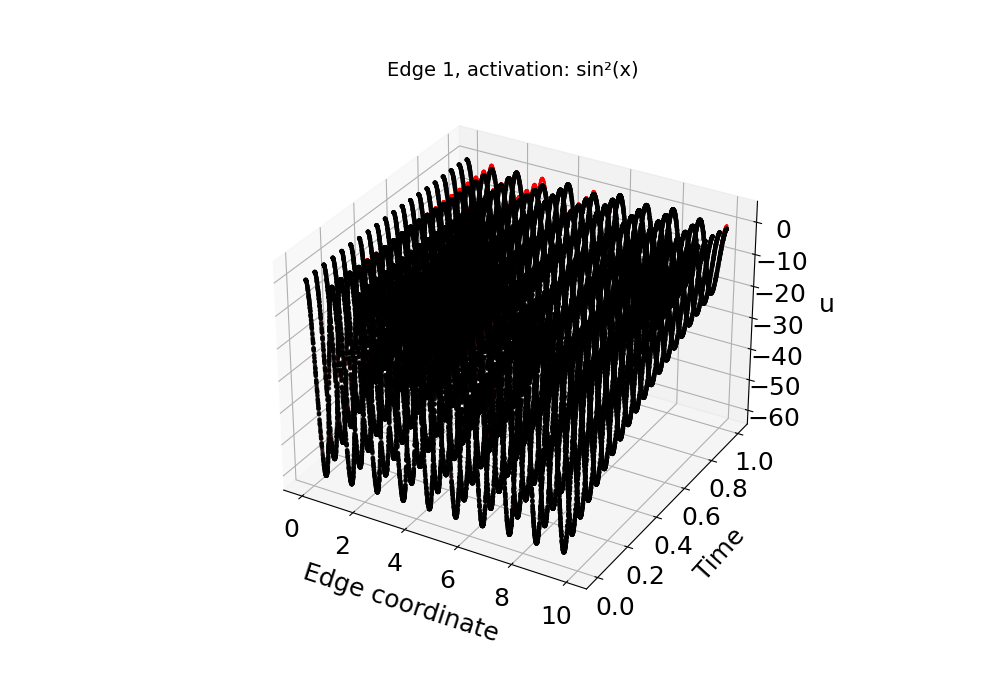}}
\adjustbox{trim={.25\width} {0\height} {.15\width} {0\height},clip}{\includegraphics[width=0.55\textwidth]{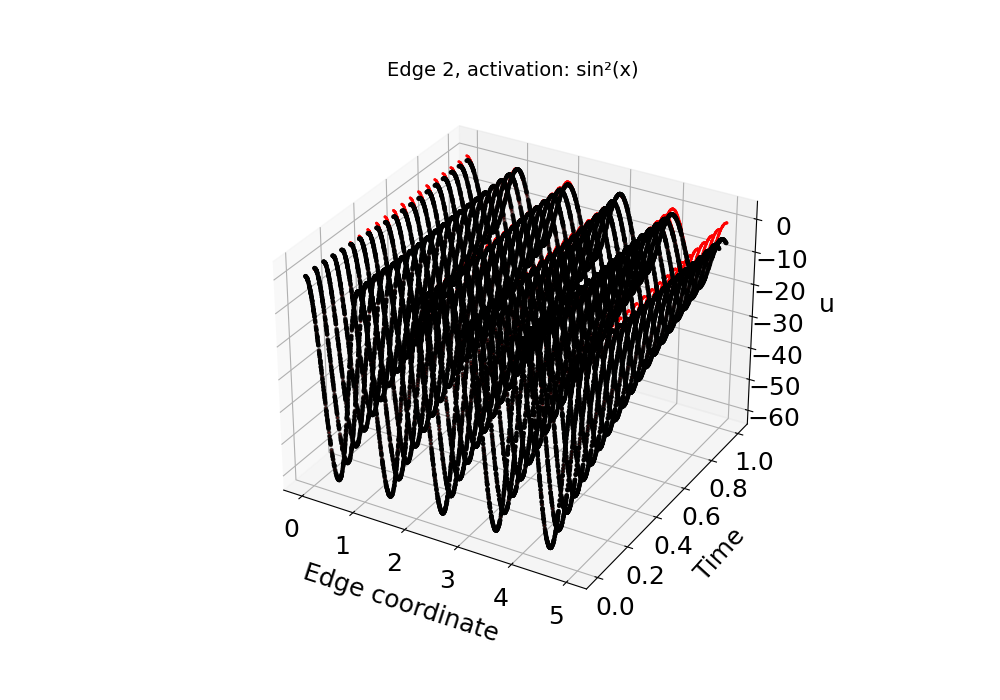}}
\vspace{-0.5cm}
\adjustbox{trim={.25\width} {0\height} {.15\width} {0\height},clip}{\includegraphics[width=0.55\textwidth]{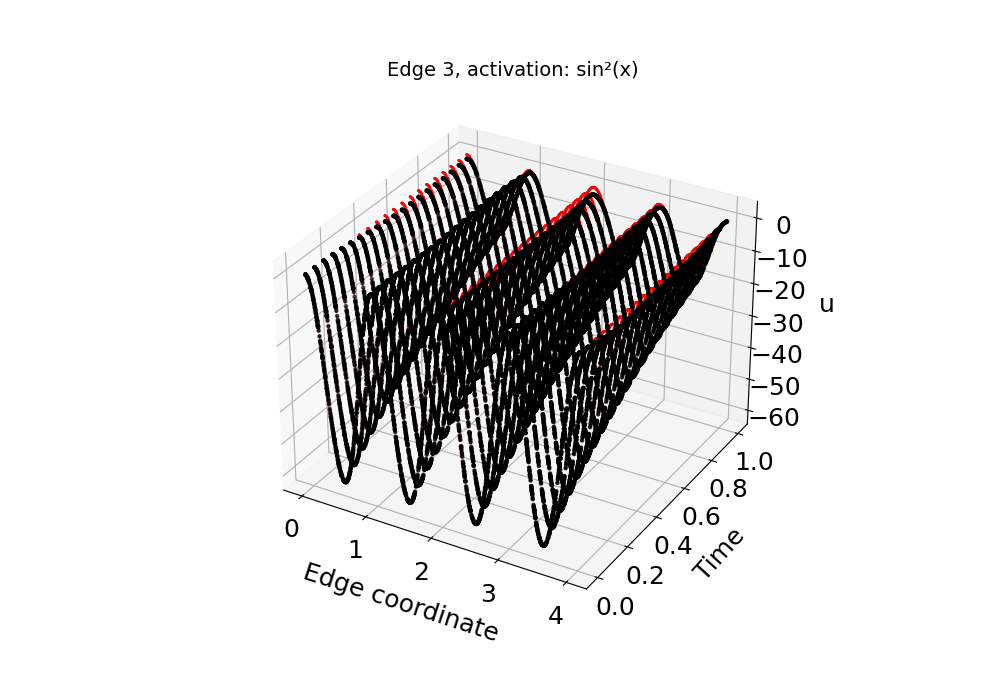}}
\adjustbox{trim={.25\width} {0\height} {.15\width} {0\height},clip}{\includegraphics[width=0.55\textwidth]{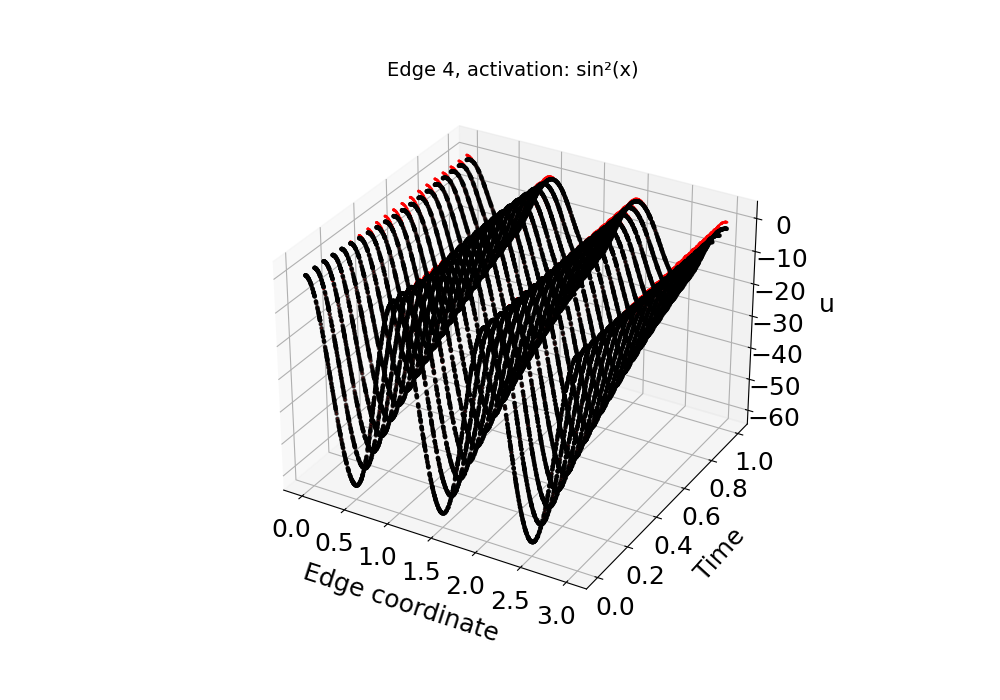}}
\adjustbox{trim={.25\width} {0\height} {.15\width} {0\height},clip}{\includegraphics[width=0.55\textwidth]{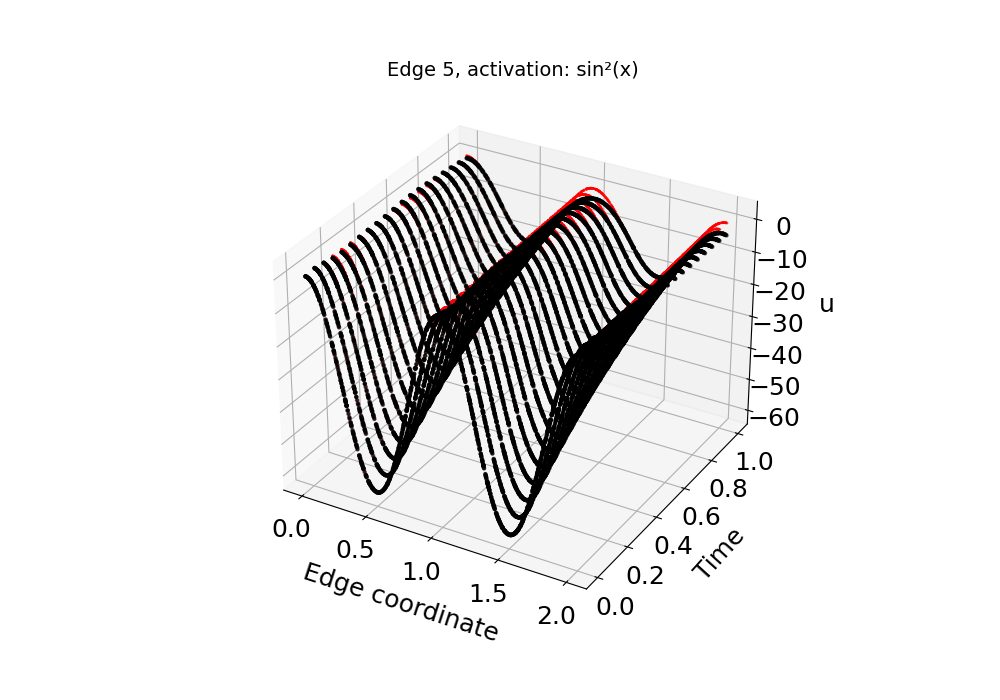}}
\vspace{-0.5cm}
\adjustbox{trim={.25\width} {0\height} {.15\width} {0\height},clip}{\includegraphics[width=0.55\textwidth]{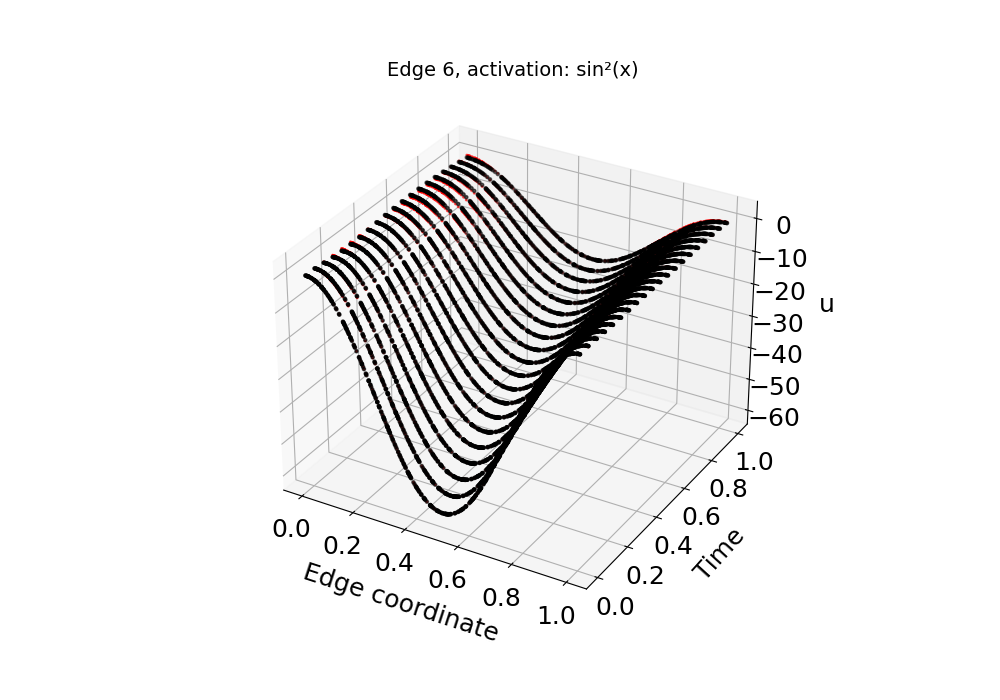}}
\adjustbox{trim={.25\width} {0\height} {.15\width} {0\height},clip}{\includegraphics[width=0.55\textwidth]{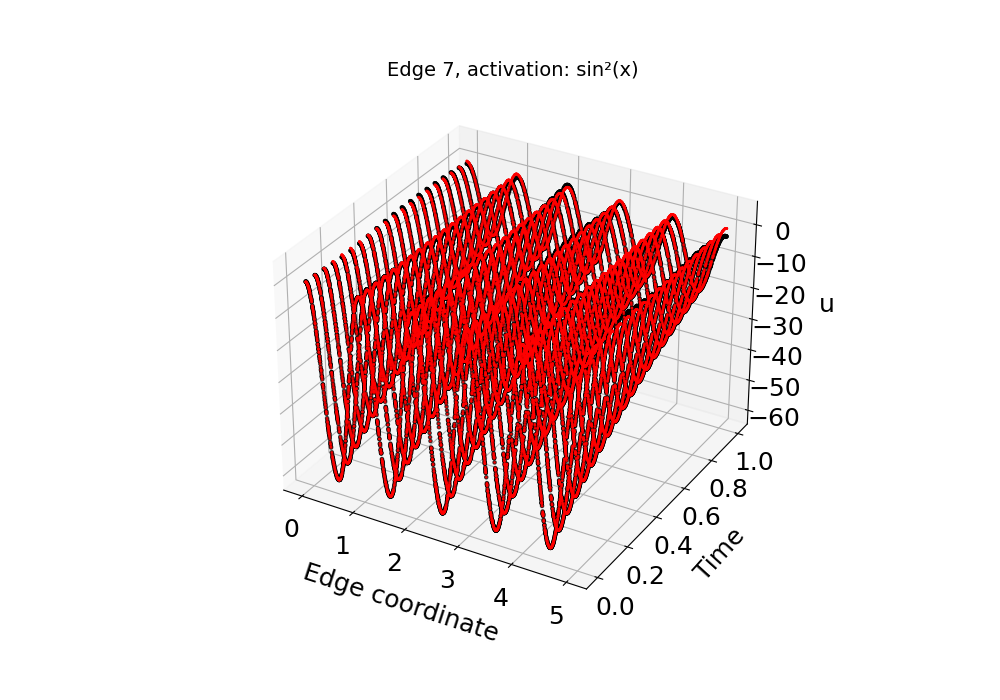}}
\caption{Validation error (top figure) and comparison on each edge (figures on the second, third, and fourth line) of the exact solution (red) and PINN approximated solution (black) of the non-linear parabolic problem defined on the general graph.}
\label{general_graph_parabolic}
\end{figure}

\begin{figure}[htbp!]
{\hspace{5cm}
\includegraphics[width=0.35\textwidth]{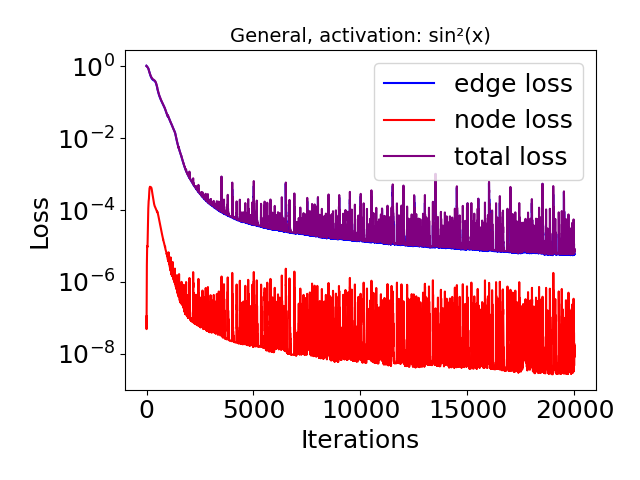}}\\
\vspace{-0.5cm}
\adjustbox{trim={.25\width} {0\height} {.15\width} {0\height},clip}{\includegraphics[width=0.55\textwidth]{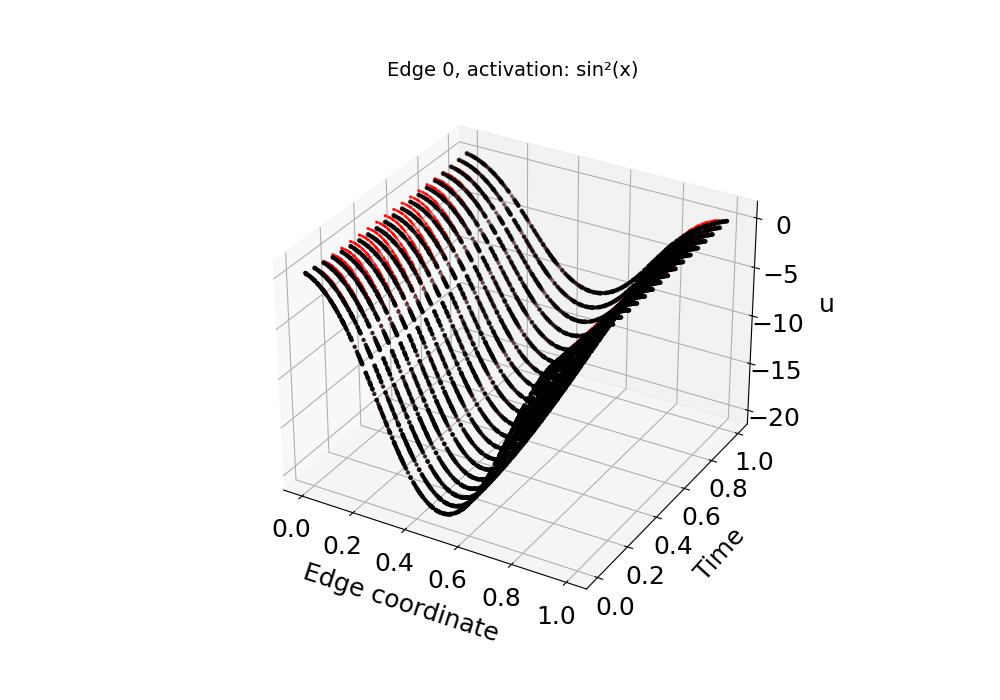}}
\adjustbox{trim={.25\width} {0\height} {.15\width} {0\height},clip}{\includegraphics[width=0.55\textwidth]{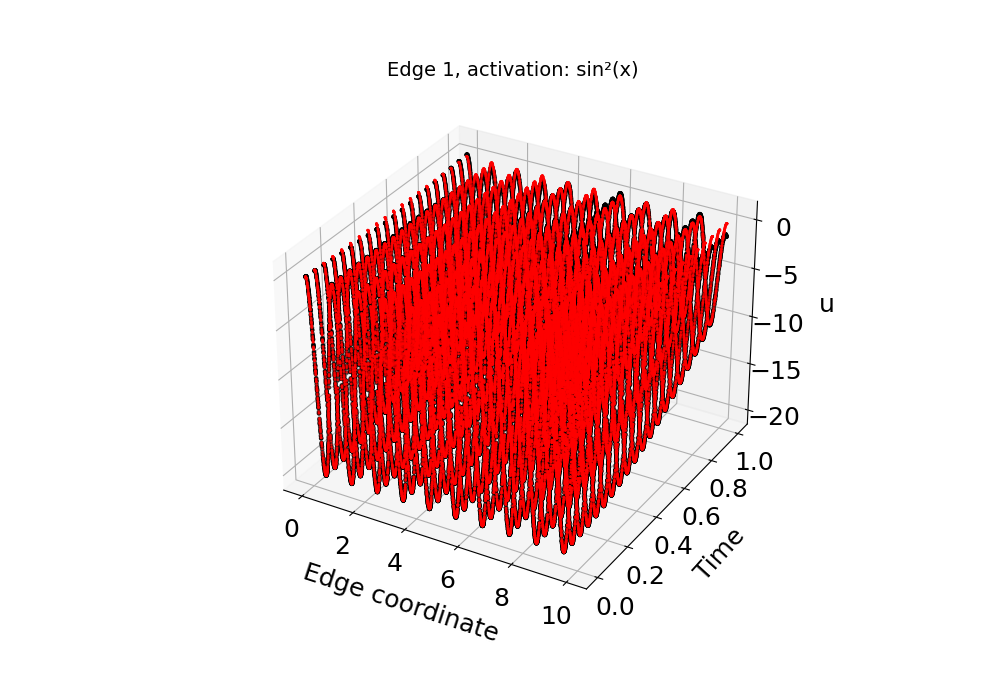}}
\adjustbox{trim={.25\width} {0\height} {.15\width} {0\height},clip}{\includegraphics[width=0.55\textwidth]{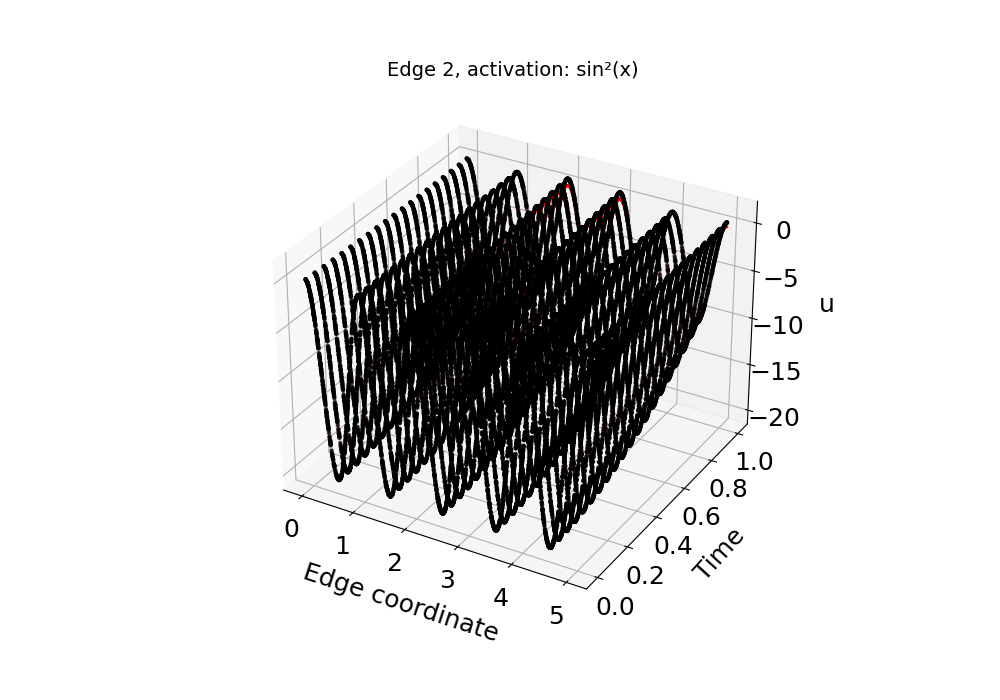}}
\vspace{-0.5cm}
\adjustbox{trim={.25\width} {0\height} {.15\width} {0\height},clip}{\includegraphics[width=0.55\textwidth]{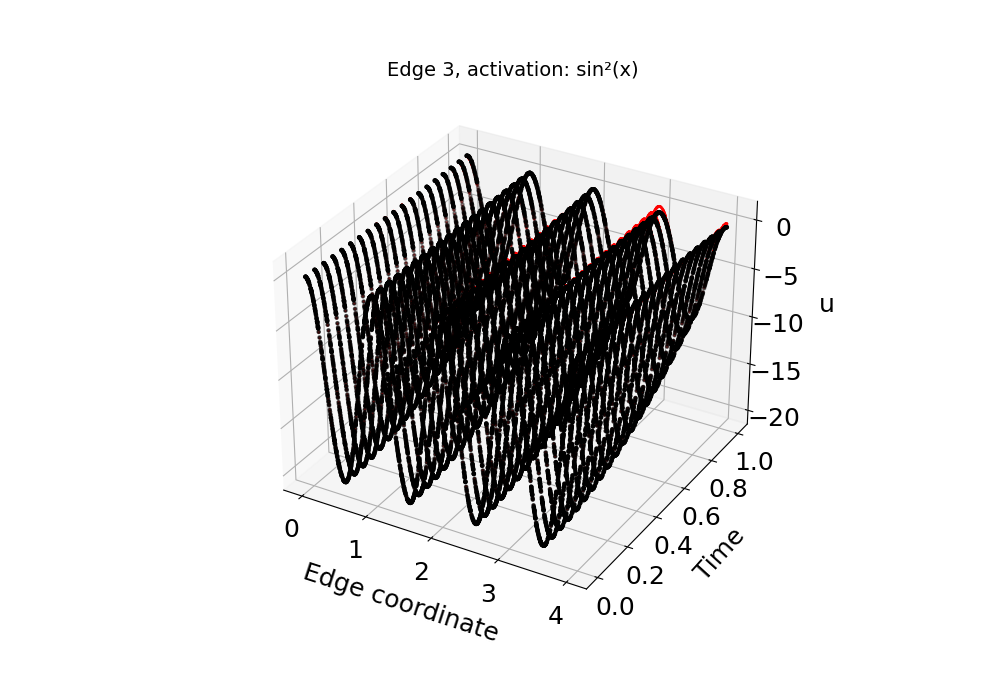}}
\adjustbox{trim={.25\width} {0\height} {.15\width} {0\height},clip}{\includegraphics[width=0.55\textwidth]{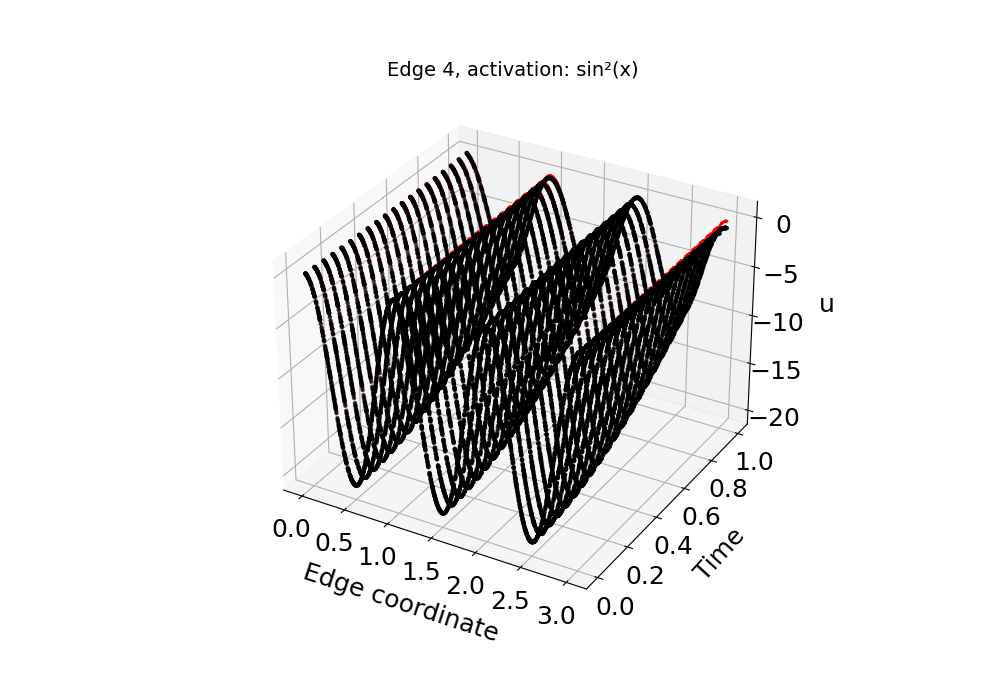}}
\adjustbox{trim={.25\width} {0\height} {.15\width} {0\height},clip}{\includegraphics[width=0.55\textwidth]{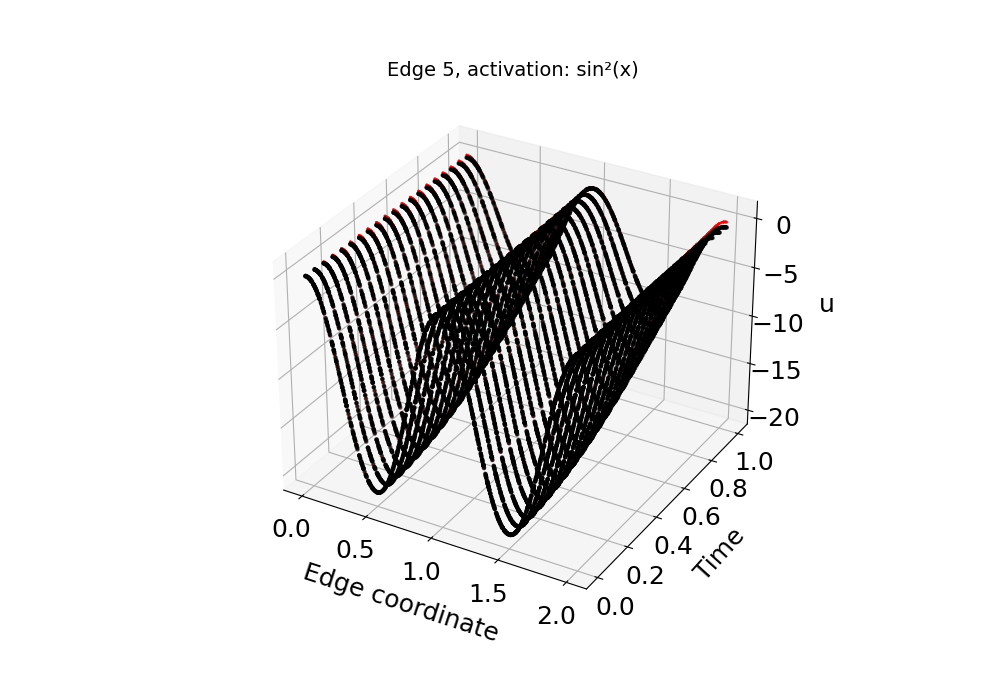}}
\vspace{-0.5cm}
\adjustbox{trim={.25\width} {0\height} {.1\width} {0\height},clip}{\includegraphics[width=0.55\textwidth]{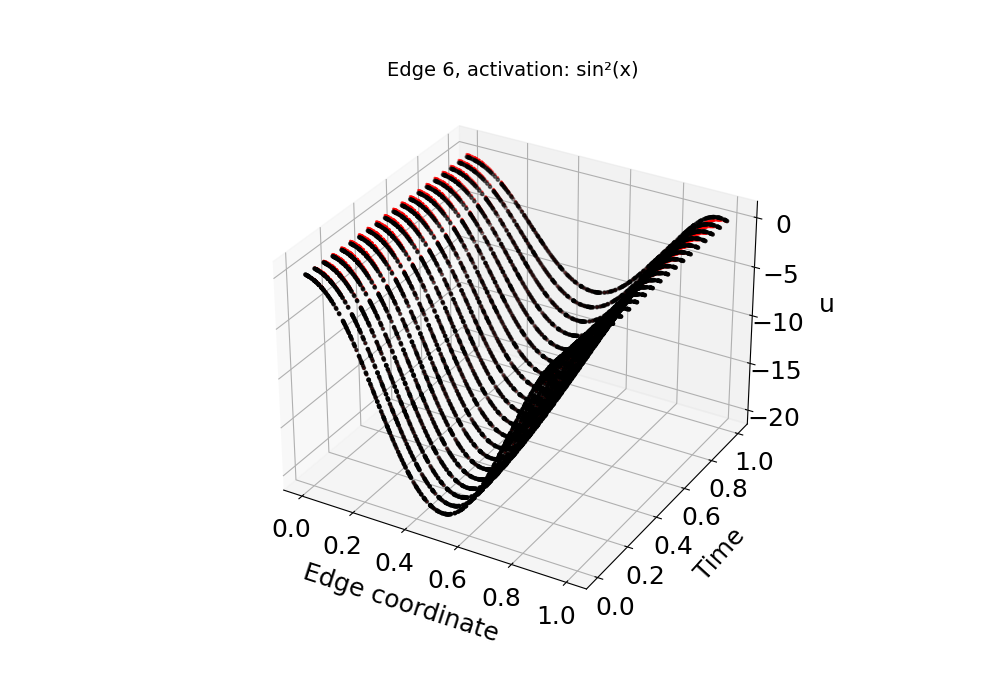}}
\adjustbox{trim={.25\width} {0\height} {.1\width} {0\height},clip}{\includegraphics[width=0.55\textwidth]{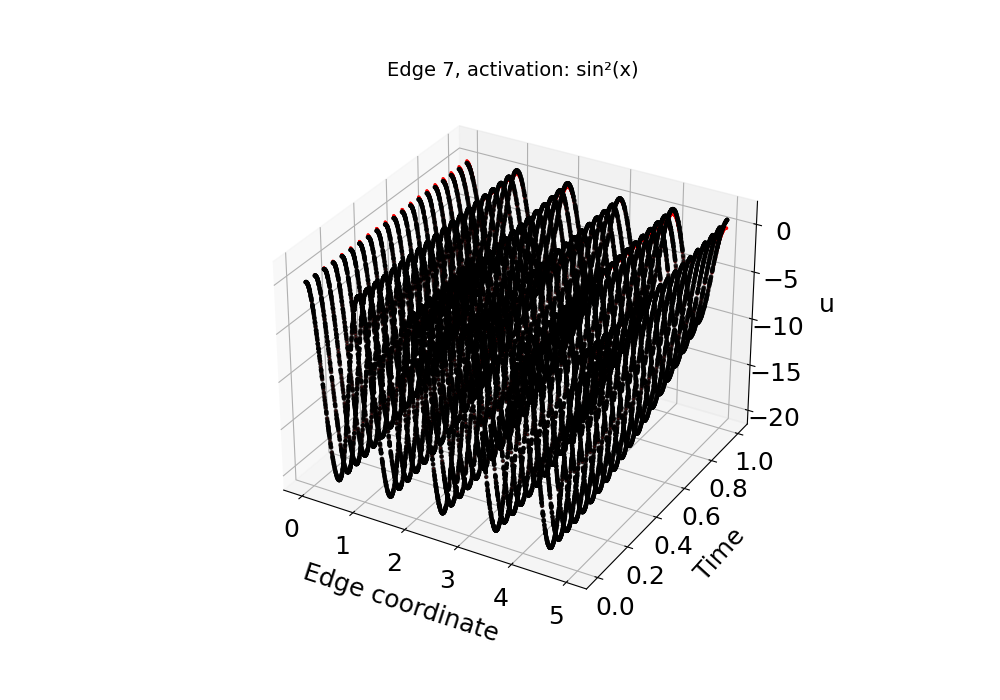}}
\caption{Validation error (top figure) and comparison on each edge (figures on the second, third, and fourth line) of the exact solution (red) and PINN approximated solution (black) of the non-linear hyperbolic problem defined on the general graph.}
\label{general_graph_hyperbolic}
\end{figure}

The final $\lambda$ and validation errors are listed in Table \ref{lambdaerror_general_graphs}. In Figures \ref{general_graph_elliptic}, \ref{general_graph_parabolic}, and \ref{general_graph_hyperbolic} we show the exact solution and the PINN solution for the differential problems defined on the general graph. Similar figures for the Lasso graph are included in the Appendix. The results show that our DL approach accurately approximates the solution to non-linear elliptic, non-linear parabolic, and non-linear hyperbolic problems defined on graphs with complex geometry, even when the length of the segment varies abruptly across different edges of the graph.

%%%%%%%%%%%%%%%%%%%%%%%%%%%%%
\begin{table}[!htbp] 
\centering
\begin{tabular}{|c|c|c|}
\hline
\backslashbox{edge length}{activation func.}
&\makebox[3em]{Lasso}&\makebox[3em]{General}
\\\hline\hline
Non-linear elliptic  &   $\lambda =94$, & $\lambda =151$
\\ 
&  err=$3.7055e-03$ & err=$3.4540e-03$  \\
\hline
Non-linear parabolic & $\lambda=153$ & $\lambda=176$ \\
 & err=$2.2302e-01$  &  err=$2.2766e-01$ \\
 \hline Non-linear hyperbolic & 
$\lambda=53$ & $\lambda=124$\\
 & err=$4.8285e-02$ &err=$4.5009e- 02$ \\\hline
\end{tabular}
\caption{Final $\lambda$ and validation error on lasso and general graphs.} 
\label{lambdaerror_general_graphs}
\end{table}

\FloatBarrier

\section{Conclusions and future developments} \label{conclusion}
We presented a novel DL approach to solve forward one-dimensional ODEs and PDEs on metric graphs that uses a system of PINN models, where each model solves the differential equation on a specific edge to which it has been assigned. The KN conditions are imposed in a weak form as an additive penalization term in the training loss function to preserve continuity of the solution and of its directional derivatives at the nodes. The penalization term in the training loss function to impose the KN conditions enforces coordination across PINN models that work on edges that share a node. 

While interpolation based methods such as FEM applied to noisy data both interpolate signal and noise, and stabilizations (e.g., Tikhonov regularization) dampen the noise effect while inevitably perturbing the nature of the problem, our approach avoids interpolation by using fitting DL models. Our DL approach provides the potential to solve inverse and control differential problems with noisy data without introducing any inductive bias in the formulation of the original problem.  

Differently from other DL approaches in the literature that use a single GNN model to solve the differential problem on the entire graph, which hinders an effective parallelization of the GNN training for graphs at large scale, our algorithm lends itself to parallelization by allowing different NN models to be trained concurrently on different distributed computing resources.   

The numerical results show that our DL approach is robust in solving differential forward problems of different nature (elliptic, parabolic, and hyperbolic), and handling varying degrees for the nodes in the metric graph. 

\noindent As future work, we plan to continue and expand on the existing work by pursuing several research directions:  
\begin{itemize}
\item We will distribute our approach on HPC resources by assigning each PINN model to a dedicated graphic processing unit (GPU) or tensor processing unit (TPU). To this goal, solving the differential problems by minimizing a global training loss function that aggregates information from all edges and nodes on the entire graph introduces severe bottlenecks for parallelization. To overcome this limitation, we will replace the global training loss function with multiple local training loss functions, each defined on a single edge and respective nodes. This will allow to replace one global communications with multiple local communications across processes, thus significantly reducing the presence of severe bottlenecks for parallelization. 
\item We will extend the applicability of our approach by using systems of PINNs to solve differential problems where the solution is a non-periodic function. To this goal, we will refer to \cite{cheridito2021efficient} where the authors showed significant broadening in the applicability of DL models to learn functions that belong to different families when the nonlinear activation functions are allowed to change across hidden layers. In particular, we will adopt this approach by combining the use of periodic and non-periodic activation functions in different layers of the PINN model.  
\item We will assess the performance (and potential benefit) of alternative approaches to handle time. In particular, we will treat time and space separately, using finite difference schemes (e.g., Runge-Kutta) for time stepping and we will compare the accuracy of this split scheme that seperately handles time and space with our all-at-once approach. 
\item We will extend our DL approach (currently used to solve forward problems) to solve also inverse and control problems defined on metric graphs.  
\end{itemize}

\section{Acknowledgements}
Massimiliano Lupo Pasini thanks Dr. Vladimir Protopopescu for his valuable feedback in the preparation of this manuscript.
Massimiliano Lupo Pasini's work was supported in part by the Office of Science of the Department of Energy and by the Artificial Intelligence Initiative as part of the Laboratory Directed Research and Development (LDRD) Program of Oak Ridge National Laboratory, managed by UT-Battelle, LLC, for the US Department of Energy under contract DE-AC05-00OR22725.

Yuanyuan Zhao thanks Dr. Sergei Avdonin and Dr. Nina Avdonina for valuable discussions in the preparation of this manuscript.  Yuanyuan Zhao's works was supported by the National Science Foundation Graduate Research Fellowship and the INTERN supplement funding under Grant No. 1242789.

This research used cloud resources of the Compute and Data Environment for Science (CADES) at the Oak Ridge National Laboratory, which is supported by the Office of Science of the U.S. Department of Energy under Contract No. DE-AC05-00OR22725.

\bibliographystyle{unsrt}
\bibliography{references}

\begin{thebibliography}{10}

\bibitem{STEINBACH2007345}
M.~C. Steinbach.
\newblock On pde solution in transient optimization of gas networks.
\newblock {\em Journal of Computational and Applied Mathematics},
  203(2):345--361, 2007.
\newblock Special Issue: The first Indo-German Conference on PDE, Scientific
  Computing and Optimization in Applications.

\bibitem{DOMSCHKE20151003}
P.~Domschke, O.~Kolb, and J.~Lang.
\newblock Adjoint-based error control for the simulation and optimization of
  gas and water supply networks.
\newblock {\em Applied Mathematics and Computation}, 259:1003--1018, 2015.

\bibitem{Egger_2020}
H.~Egger and L.~Schöbel-Kröhn.
\newblock Chemotaxis on networks: analysis and numerical approximation.
\newblock {\em {ESAIM}: Mathematical Modelling and Numerical Analysis},
  54(4):1339--1372, jun 2020.

\bibitem{M2AN_2014__48_1_231_0}
G.~Bretti, Roberto Natalini, and M.~Ribot.
\newblock A hyperbolic model of chemotaxis on a network: a numerical study.
\newblock {\em ESAIM: Mathematical Modelling and Numerical Analysis -
  Mod\'elisation Math\'ematique et Analyse Num\'erique}, 48(1):231--258, 2014.

\bibitem{1556-1801_2017_3_381}
R.~Borsche, A.~Klar, and T.~N.~H. Pham.
\newblock Nonlinear flux-limited models for chemotaxis on networks.
\newblock {\em Networks and Heterogeneous Media},
  12(1556-1801\_2017\_3\_381):381--401, 2017.

\bibitem{Herty2010}
M.~Herty, J.~Mohring, and V.~Sachers.
\newblock A new model for gas flow in pipe networks.
\newblock {\em Mathematical Methods in the Applied Sciences}, 33(7):845--855,
  2011.

\bibitem{piccoli_traffic_network_2006}
M.~Garavello and B.~Piccoli.
\newblock {\em Traffic flow on networks}, volume~1.
\newblock American Institute of Mathematical Sciences, 2006.

\bibitem{carbon_nanostructures}
P.~Kuchment and O.~Post.
\newblock On the spectra of carbon nano-structures.
\newblock {\em Communications in Mathematical Physics volume}, 3275:805--826,
  2007.

\bibitem{OPPENHEIMER2000223}
S.~F. Oppenheimer.
\newblock A convection–diffusion problem in a network.
\newblock {\em Applied Mathematics and Computation}, 112(2):223--240, 2000.

\bibitem{1937-5093_2011_4_1081}
M.~Herty and C.~Ringhofer.
\newblock Averaged kinetic models for flows on unstructured networks.
\newblock {\em Kinetic and Related Models},
  4(1937-5093\_2011\_4\_1081):1081--1096, 2011.

\bibitem{GARCIA2015120}
L.~García, J.~Barreiro-Gomez, E.~Escobar, D.~Téllez, N.~Quijano, and
  N.~Ocampo-Martinez.
\newblock Modeling and real-time control of urban drainage systems: A review.
\newblock {\em Advances in Water Resources}, 85:120--132, 2015.

\bibitem{DUCA2021109324}
A.~Duca.
\newblock Bilinear quantum systems on compact graphs: Well-posedness and global
  exact controllability.
\newblock {\em Automatica}, 123:109324, 2021.

\bibitem{social_network}
B.~Du, X.~Lian, and X.~Cheng.
\newblock Partial differential equation modeling with {Dirichlet} boundary
  conditions on social networks.
\newblock {\em Boundary Value Problems}, (50):2035--2052, 2018.

\bibitem{Cheng2018}
X.~Cheng and J.M.A. Scherpen.
\newblock Clustering approach to model order reduction of power networks with
  distributed controllers.
\newblock {\em Advances in Computational Mathematics volume}, 44:1917--1939,
  2018.

\bibitem{Lagnese1993}
J.~E. Lagnese, G.~Leugering, and E.~J.~P.~G. Schmidt.
\newblock Modelling of dynamic networks of thin thermoelastic beams.
\newblock {\em Mathematical Methods in the Applied Sciences}, 16(5):327--358,
  1993.

\bibitem{KIIK20151871}
J.-C. Kiik, P.~Kurasov, and M.~Usman.
\newblock On vertex conditions for elastic systems.
\newblock {\em Physics Letters A}, 379(34):1871--1876, 2015.

\bibitem{Berkolaiko_2022}
G.~Berkolaiko and M.~Ettehad.
\newblock Three-dimensional elastic beam frames: Rigid joint conditions in
  variational and differential formulation.
\newblock {\em Studies in Applied Mathematics}, 148(4):1586--1623, feb 2022.

\bibitem{WALLACE20172035}
M.~Wallace, R.~Feres, and G.~Yablonsky.
\newblock Reaction–diffusion on metric graphs: from {3D} to {1D}.
\newblock {\em Computers \& Mathematics with Applications}, 73(9):2035--2052,
  2017.

\bibitem{GUGAT202259}
M.~Gugat and M.~Herty.
\newblock Chapter 2 - modeling, control, and numerics of gas networks.
\newblock In Emmanuel Trélat and Enrique Zuazua, editors, {\em Numerical
  Control: Part A}, volume~23 of {\em Handbook of Numerical Analysis}, pages
  59--86. Elsevier, 2022.

\bibitem{Gugat2011FlowCI}
M.~Gugat, M.~Herty, and V.~Schleper.
\newblock Flow control in gas networks: Exact controllability to a given
  demand.
\newblock {\em Mathematical Methods in the Applied Sciences}, 34(7):745--757,
  2011.

\bibitem{7402932}
A.~Zlotnik, M.~Chertkov, and S.~Backhaus.
\newblock Optimal control of transient flow in natural gas networks.
\newblock In {\em 2015 54th IEEE Conference on Decision and Control (CDC)},
  pages 4563--4570, 2015.

\bibitem{family_exponentials}
S.~A. Avdonin and S~A. Ivanov.
\newblock {\em Families of Exponentials. The Method of Moments in
  Controllability Problems for Distributed Parameter Systems}.
\newblock Cambridge University Press, 1995.

\bibitem{avdonin2022exact}
Sergei Avdonin and Yuanyuan Zhao.
\newblock Exact controllability of the wave equation on graphs.
\newblock {\em Applied Mathematics \& Optimization}, 85(2):1--44, 2022.

\bibitem{AVDONIN201952}
S.~A. Avdonin.
\newblock Control, observation and identification problems for the wave
  equation on metric graphs.
\newblock {\em IFAC-PapersOnLine}, 52(2):52--57, 2019.
\newblock 3rd IFAC Workshop on Control of Systems Governed by Partial
  Differential Equations CPDE 2019.

\bibitem{alam2021control}
M~Alam, S~Avdonin, and N~Avdonina.
\newblock Control problems for the telegraph and wave equation networks.
\newblock In {\em Journal of Physics: Conference Series}, volume 1847, page
  012015. IOP Publishing, 2021.

\bibitem{Kurasov_2005}
P.~Kurasov and M.~Nowaczyk.
\newblock Inverse spectral problem for quantum graphs.
\newblock {\em Journal of Physics A: Mathematical and General},
  38(22):4901--4915, may 2005.

\bibitem{https://doi.org/10.48550/arxiv.2103.16727}
E.~Sadurni and T.~H Seligman.
\newblock Inverse problems in quantum graphs and accidental degeneracy.
\newblock {\em ArXiv}, (arxiv.2103.16727), 2021.

\bibitem{vibration3040028}
S.~Avdonin and J.~Edward.
\newblock An inverse problem for quantum trees with delta-prime vertex
  conditions.
\newblock {\em Vibration}, 3(4):448--463, 2020.

\bibitem{https://doi.org/10.48550/arxiv.2202.00944}
E.~Blåsten, P.~Exner, H.~Isozaki, M.~Lassas, and J.~Lu.
\newblock Inverse problems for locally perturbed lattices -- discrete
  hamiltonian and quantum graph.
\newblock {\em ArXiv}, (arxiv.2202.00944), 2022.

\bibitem{GuanYangWu+2021+577+585}
S.-Y. Guan, C.-F. Yang, and D.-J. Wu.
\newblock A partial inverse problem for quantum graphs with a loop.
\newblock {\em Journal of Inverse and Ill-posed Problems}, 29(4):577--585,
  2021.

\bibitem{https://doi.org/10.48550/arxiv.1505.00185}
J.~Solomon.
\newblock {PDE} approaches to graph analysis.
\newblock {\em ArXiv}, arxiv.1505.00185, 2015.

\bibitem{MERCIER2008174}
D.~Mercier and V.~Régnier.
\newblock Spectrum of a network of {Euler}–{Bernoulli} beams.
\newblock {\em Journal of Mathematical Analysis and Applications},
  337(1):174--196, 2008.

\bibitem{KOTTOS199976}
T.~Kottos and U.~Smilansky.
\newblock Periodic orbit theory and spectral statistics for quantum graphs.
\newblock {\em Annals of Physics}, 274(1):76--124, 1999.

\bibitem{VONBELOW1985309}
J.~{von Below}.
\newblock A characteristic equation associated to an eigenvalue problem on
  c2-networks.
\newblock {\em Linear Algebra and its Applications}, 71:309--325, 1985.

\bibitem{https://doi.org/10.1007/s00023-017-0601-2}
R.~Band and G.~Lévy.
\newblock Quantum graphs which optimize the spectral gap.
\newblock {\em Annales Henri Poincaré}, 18:3269--3323, 2017.

\bibitem{doi:10.1137/14097865X}
M.~Laurent and T.~Piovesan.
\newblock Conic approach to quantum graph parameters using linear optimization
  over the completely positive semidefinite cone.
\newblock {\em SIAM Journal on Optimization}, 25(4):2461--2493, 2015.

\bibitem{mesh_refinement_gas_pdes}
P.~Domschke, A.~Dua, J.~.J Stolwijk, J.~Lang, and V.~Mehrmann.
\newblock Adaptive refinement strategies for the simulation of gas flow in
  networks using a model hierarchy.
\newblock {\em Electronic Transactions on Numerical Analysis}, 48:97--113,
  2018.
\newblock Online available: https://epub.oeaw.ac.at/?arp=0x0038867b - Last
  access:9.7.2022.

\bibitem{10.1007/978-3-030-43651-3_45}
H.~Egger and N.~Philippi.
\newblock A hybrid discontinuous galerkin method for transport equations on
  networks.
\newblock In Robert Kl{\"o}fkorn, Eirik Keilegavlen, Florin~A. Radu, and
  J{\"u}rgen Fuhrmann, editors, {\em Finite Volumes for Complex Applications IX
  - Methods, Theoretical Aspects, Examples}, pages 487--495, Cham, 2020.
  Springer International Publishing.

\bibitem{GRUNDEL202260}
S.~Grundel and M.~Herty.
\newblock Hyperbolic discretization of simplified {Euler} equation via
  {Riemann} invariants.
\newblock {\em Applied Mathematical Modelling}, 106:60--72, 2022.

\bibitem{10.1007/978-3-319-27517-8_1}
S.~Grundel, N.~Hornung, and S.~Roggendorf.
\newblock Numerical aspects of model order reduction for gas transportation
  networks.
\newblock In Slawomir Koziel, Leifur Leifsson, and Xin-She Yang, editors, {\em
  Simulation-Driven Modeling and Optimization}, pages 1--28, Cham, 2016.
  Springer International Publishing.

\bibitem{PESENSON2005203}
I.~Pesenson.
\newblock Polynomial splines and eigenvalue approximations on quantum graphs.
\newblock {\em Journal of Approximation Theory}, 135(2):203--220, 2005.

\bibitem{Wybo2015}
W.~A. Wybo, D.~Boccalini, B.~Torben-Nielsen, and M.~O. Gewaltig.
\newblock A sparse reformulation of the green's function formalism allows
  efficient simulations of morphological neuron models.
\newblock {\em Neural Computation}, 27(12):2587--2622, 2015.

\bibitem{10.1093/imanum/drx029}
M.~Arioli and M.~Benzi.
\newblock {A finite element method for quantum graphs}.
\newblock {\em IMA Journal of Numerical Analysis}, 38(3):1119--1163, 06 2017.

\bibitem{Leugering2017}
L.~García, J.~Barreiro-Gomez, E.~Escobar, D.~Téllez, N.~Quijano, and
  C.~Ocampo-Martinez.
\newblock Non-overlapping domain decomposition for optimal control problems
  governed by semi-linear models for gas flow in networks.
\newblock {\em Control and Cybernetics}, 46(3):191--225, 2017.

\bibitem{Stoll2021}
M.~Stoll and M.~Winkler.
\newblock Optimal {D}irichlet control of partial differential equations on
  networks.
\newblock {\em Electronic Transactions on Numerical Analysis}, 54:392--419,
  2021.

\bibitem{ghattas_willcox_2021}
Omar Ghattas and Karen Willcox.
\newblock Learning physics-based models from data: perspectives from inverse
  problems and model reduction.
\newblock {\em Acta Numer.}, 30:445–554, 2021.

\bibitem{weather_piml}
Albert A. Wu J-L. Jiang C. Esmaeilzadeh S. Azizzadenesheli K. Wang R.
  Chattopadhyay A. Singh A. Manepalli A. Chirila D. Yu R. Walters R. White B.
  Xiao H. Tchelepi H. A. Marcus P. Anandkumar A. Hassanzadeh~P. Kashinath~K.,
  Mustafa~M. and Prabhat.
\newblock Physics-informed neural networks: A deep learning framework for
  solving forward and inverse problems involving nonlinear partial differential
  equations.
\newblock {\em 2021 Philosophical Transactions of the Royal Society A},
  379(20200093):20200093, 2021.

\bibitem{piml}
G.~E. Karniadakis, I.~G. Kevrekidis, L.~Lu, P.~Perdikaris, S.~Wang, and
  L.~Yang.
\newblock Physics-informed machine learning.
\newblock {\em Nature Reviews Physics volume}, 3:422--440, 2021.

\bibitem{https://doi.org/10.48550/arxiv.2204.00538}
M.~Lupo~Pasini and S.~Perotto.
\newblock Hierarchical model reduction driven by machine learning for
  parametric advection-diffusion-reaction problems in the presence of noisy
  data.
\newblock {\em ArXiv}, (arxiv.2204.00538), 2022.

\bibitem{raissi2019physics}
M.~Raissi, P.~Perdikaris, and G.~E. Karniadakis.
\newblock Physics-informed neural networks: A deep learning framework for
  solving forward and inverse problems involving nonlinear partial differential
  equations.
\newblock {\em Journal of Computational Physics}, 378:686--707, 2019.

\bibitem{Chen:20}
Y.~Chen, L.~Lu, G.~E. Karniadakis, and L.~Dal~Negro.
\newblock Physics-informed neural networks for inverse problems in nano-optics
  and metamaterials.
\newblock {\em Opt. Express}, 28(8):11618--11633, Apr 2020.

\bibitem{https://doi.org/10.48550/arxiv.2202.11821}
A.~D. Jagtap, Z.~Mao, N.~Adams, and G.~E. Karniadakis.
\newblock Physics-informed neural networks for inverse problems in supersonic
  flows.
\newblock {\em ArXiv}, (https://doi.org/10.48550/arxiv.2202.11821), 2022.

\bibitem{graphNN_pde}
V.~Iakovlev, M.~Heinonen, and H.~L{\"a}hdesm{\"a}ki.
\newblock Learning continuous-time pdes from sparse data with graph neural
  networks.
\newblock In {\em International Conference on Learning Representations}, 2021.
\newblock International Conference on Learning Representations, ICLR ;
  Conference date: 25-04-2021 Through 29-04-2021.

\bibitem{https://doi.org/10.48550/arxiv.2108.01938}
M.~Eliasof, E.~Haber, and E.~Treister.
\newblock {PDE-GCN}: Novel architectures for graph neural networks motivated by
  partial differential equations.
\newblock {\em ArXiv}, arxiv.2108.01938, 2021.

\bibitem{alexander2020}
A.~M.~Sadiku C.~Alexander.
\newblock {\em Fundamentals of Electric Circuits, Seventh Edition}.
\newblock McGraw Hill, 2020.

\bibitem{hornik1989multilayer}
K.~Hornik, M.~Stinchcombe, and H.~White.
\newblock Multilayer feedforward networks are universal approximators.
\newblock {\em Neural Networks}, 2(5):359--366, 1989.

\bibitem{funahashi1989approximate}
K.-I. Funahashi.
\newblock On the approximate realization of continuous mappings by neural
  networks.
\newblock {\em Neural Networks}, 2(3):183--192, 1989.

\bibitem{leshno1993multilayer}
M.~Leshno, V.~Ya Lin, A.~Pinkus, and S.~Schocken.
\newblock Multilayer feedforward networks with a nonpolynomial activation
  function can approximate any function.
\newblock {\em Neural Networks}, 6(6):861--867, 1993.

\bibitem{mlp}
F.~Rosenblatt.
\newblock Principles of neurodynamics: perceptrons and the theory of brain
  mechanisms.
\newblock {\em Nature}, 542(7639):75--79, 2017.

\bibitem{goodfellow}
I.~Goodfellow, Y.~Bengio, and A.~Courville.
\newblock {\em Deep Learning}.
\newblock The MIT Press, Cambridge, Massachusetts, 2016.

\bibitem{pinns2019}
M.~Raissi, P.~Perdikaris, and G.E. Karniadakis.
\newblock Physics-informed neural networks: a deep learning framework for
  solving forward and inverse problems involving nonlinear partial differential
  equations.
\newblock {\em Journal of Computational Physics}, 378:686--707, February 2019.

\bibitem{platt1987constrained}
J.~Platt and A.~Barr.
\newblock Constrained differential optimization.
\newblock In D.~Anderson, editor, {\em Neural Information Processing Systems}.
  American Institute of Physics, 1987.

\bibitem{fey_2019}
M.~Fey and J.~E. Lenssen.
\newblock Fast graph representation learning with {PyTorch Geometric}.
\newblock In {\em ICLR Workshop on Representation Learning on Graphs and
  Manifolds}, 2019.

\bibitem{torch_geometric}
{PyTorch Geometric}.
\newblock \url{https://pytorch-geometric.readthedocs.io/en/latest/}.

\bibitem{adam}
D.~P. Kingma and J.~Ba.
\newblock Adam: a method for stochastic optimization.
\newblock {\em arXiv:1412.6980 [cs]}, January 2017.
\newblock arXiv: 1412.6980.

\bibitem{ziyin2020neural}
L.~Ziyin, T.~Hartwig, and M.~Ueda.
\newblock Neural networks fail to learn periodic functions and how to fix it.
\newblock {\em Advances in Neural Information Processing Systems},
  33:1583--1594, 2020.

\bibitem{Kurasov:13}
P.~Kurasov.
\newblock Inverse scattering for lasso graph.
\newblock {\em Journal of Mathematical Physics}, 54(042103), Apr 2013.

\bibitem{cheridito2021efficient}
Patrick Cheridito, Arnulf Jentzen, and Florian Rossmannek.
\newblock Efficient approximation of high-dimensional functions with neural
  networks.
\newblock {\em IEEE Transactions on Neural Networks and Learning Systems},
  2021.

\end{thebibliography}

\clearpage

\section*{Appendix}

%%%%%%%%%%%%%%%%%%%%%%%%%%%%%%%%%%%%%%%%%%%%
\begin{figure}
\vspace{-0.2cm}
\includegraphics[width=0.45\textwidth]{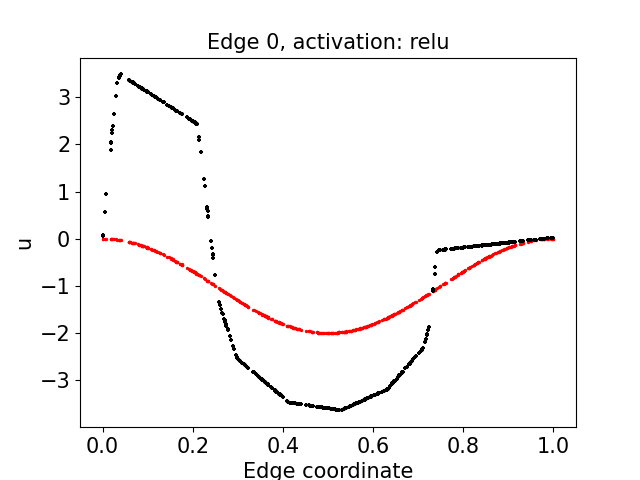}
\includegraphics[width=0.45\textwidth]{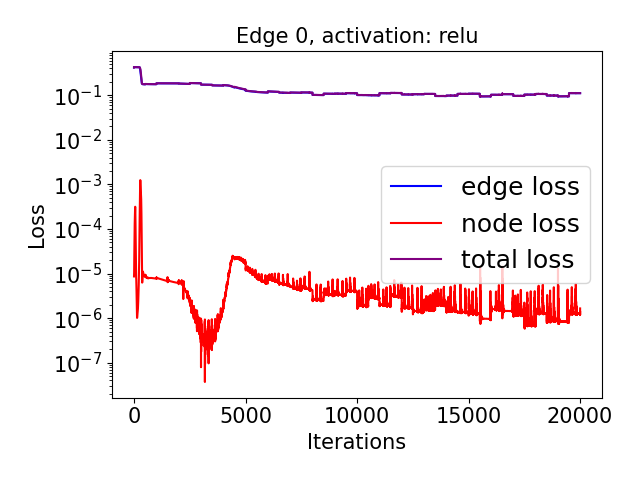}\\
\vspace{-0.2cm}
\includegraphics[width=0.45\textwidth]{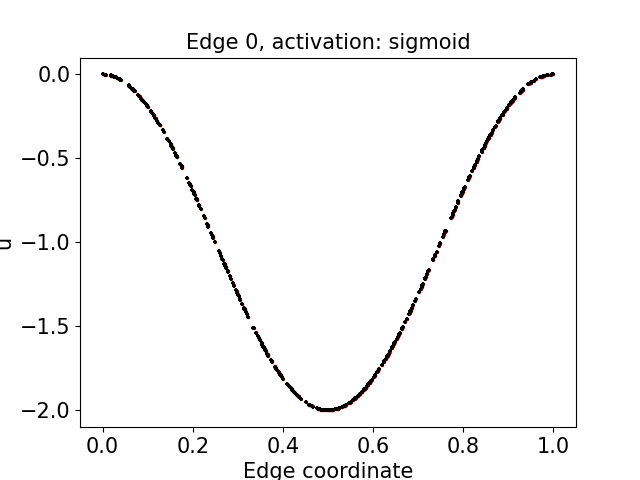}
\includegraphics[width=0.45\textwidth]{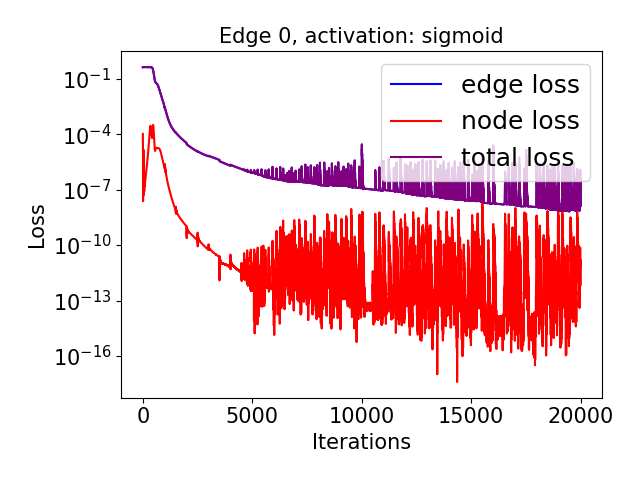}\\
\vspace{-0.2cm}
\includegraphics[width=0.45\textwidth]{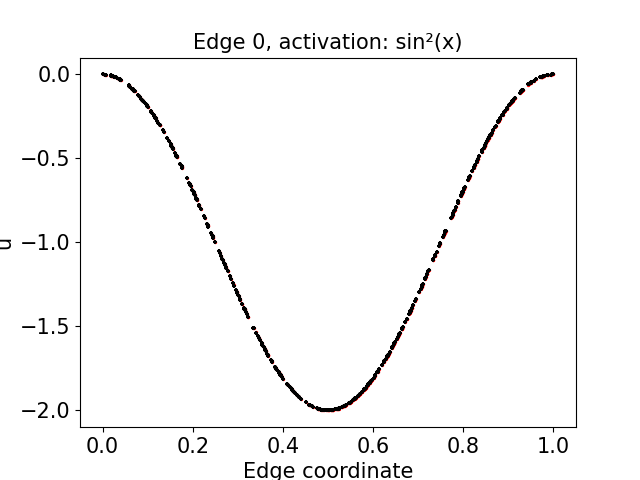}
\includegraphics[width=0.45\textwidth]{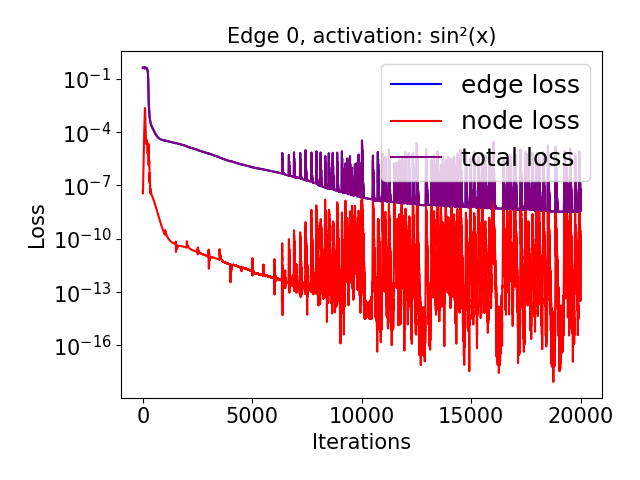}\\
\vspace{-0.2cm}
\includegraphics[width=0.45\textwidth]{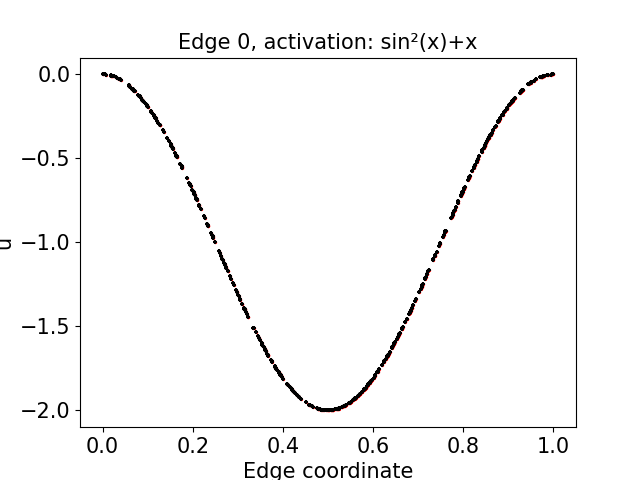}
\includegraphics[width=0.45\textwidth]{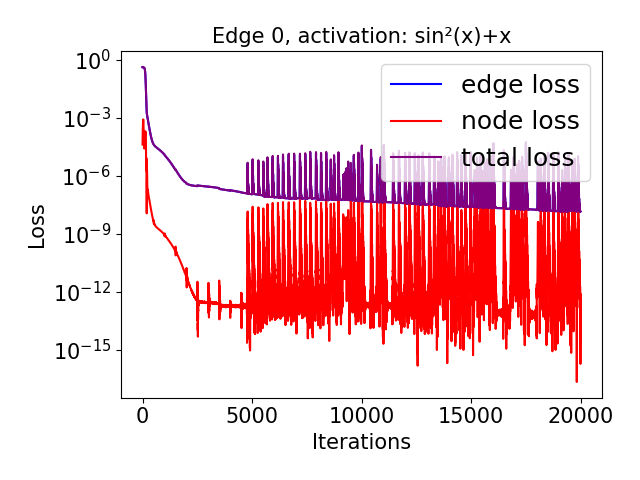}\\
\vspace{-0.2cm}
\caption{Model output for the non-linear elliptic problem on one edge of length $1$ using different activation functions}\label{hel1}

\end{figure}

%%%%%%%%%%%%%%%%%%%%%%%%%%%%%%%%%%%%%%%%%%%
\begin{figure}[!htbp]
\vspace{-0.2cm}
\includegraphics[width=0.45\textwidth]{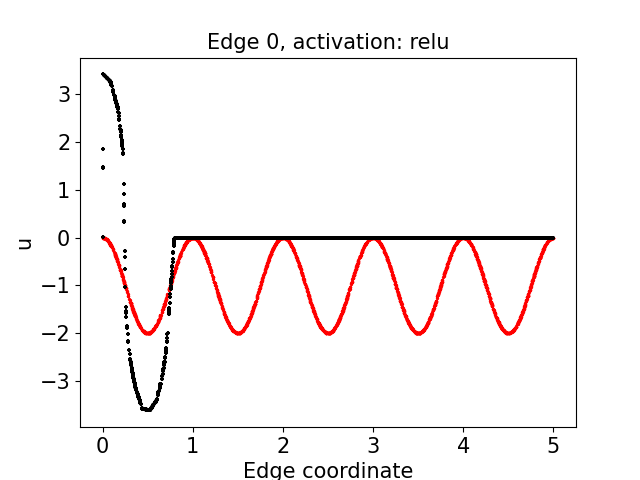}
\includegraphics[width=0.45\textwidth]{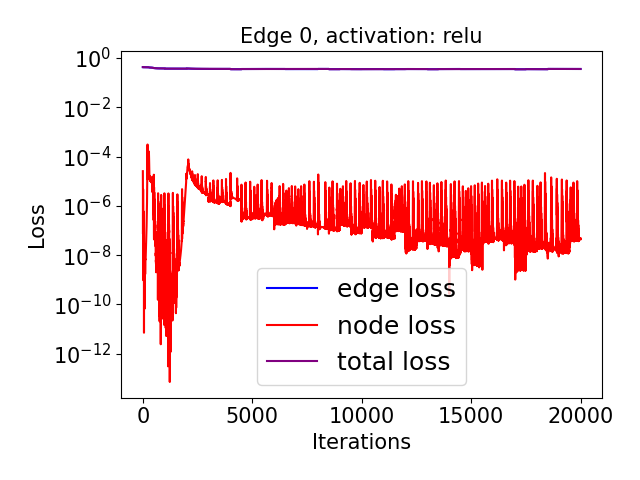}\\
\vspace{-0.2cm}
\includegraphics[width=0.45\textwidth]{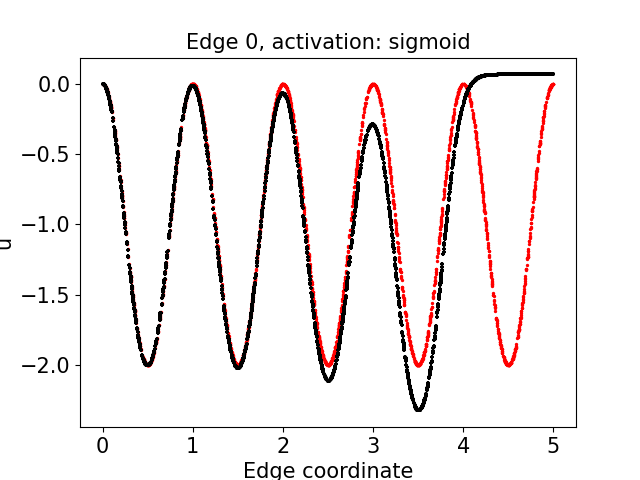}
\includegraphics[width=0.45\textwidth]{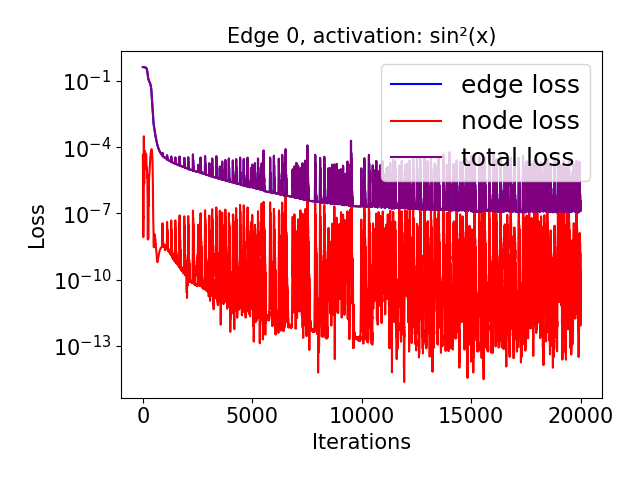}\\
\vspace{-0.2cm}
\includegraphics[width=0.45\textwidth]{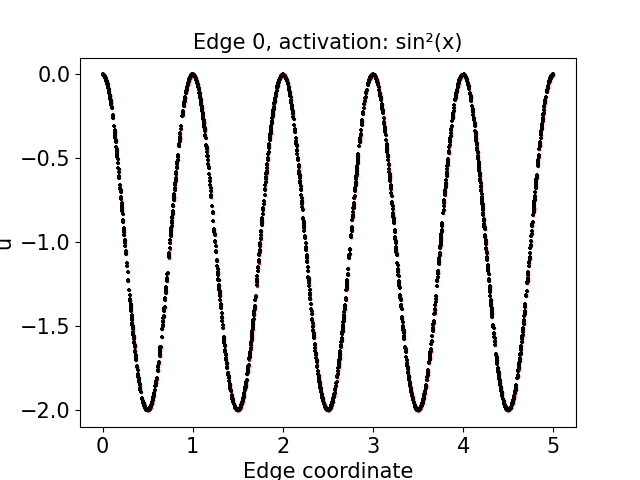}
\includegraphics[width=0.45\textwidth]{edge5_s2_fre=1_lalr=1000.000000_edge_0_loss_25.png} \\
\vspace{-0.2cm}
\includegraphics[width=0.45\textwidth]{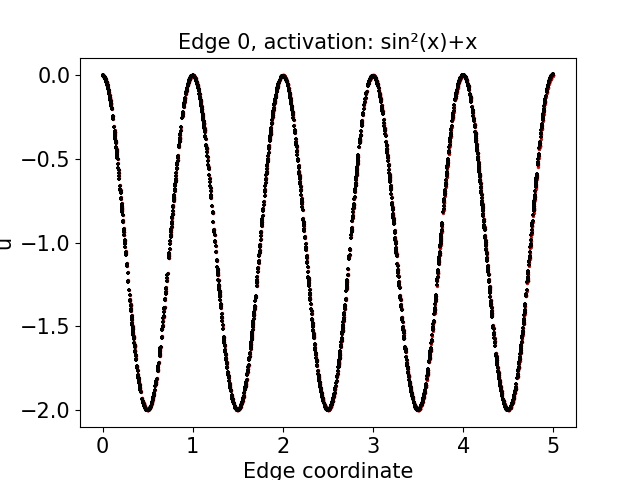}
\includegraphics[width=0.45\textwidth]{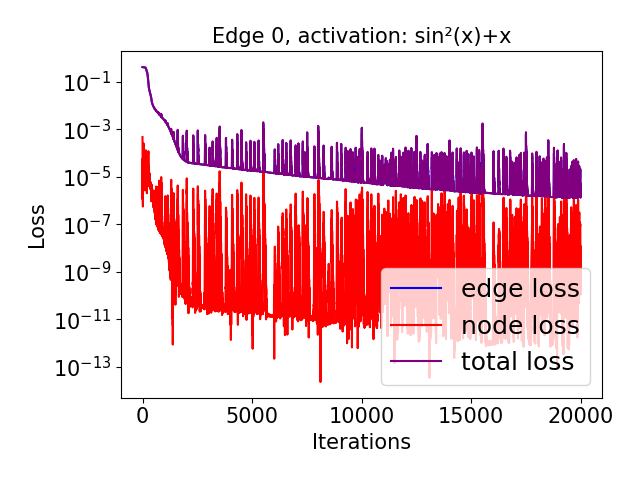}\\
\vspace{-0.2cm}
\caption{Model output for the non-linear elliptic problem on one edge of length $5$ using different activation functions}\label{hel5}
\end{figure}

%%%%%%%%%%%%%%%%%%%%%%%%%%%%%%
\begin{figure}[!htbp]
\vspace{-0.2cm}
\includegraphics[width=0.45\textwidth]{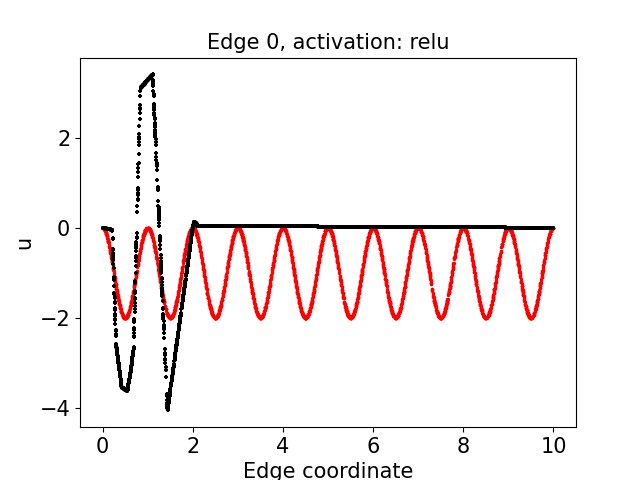}
\includegraphics[width=0.45\textwidth]{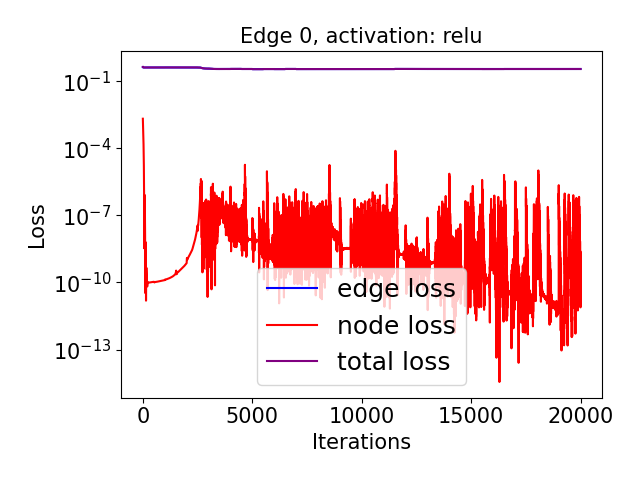}\\
\vspace{-0.2cm}
\includegraphics[width=0.45\textwidth]{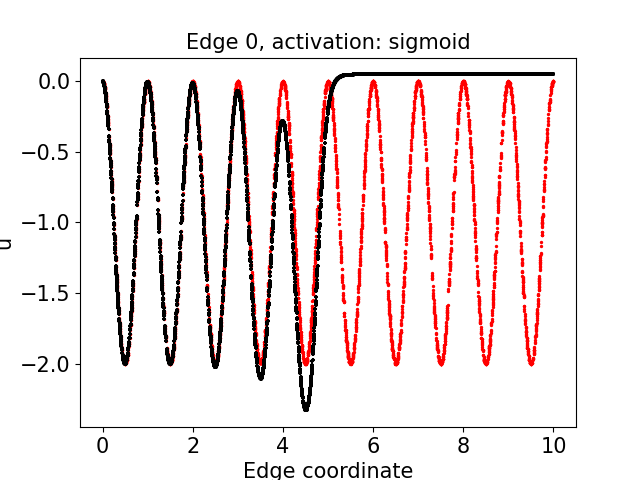}
\includegraphics[width=0.45\textwidth]{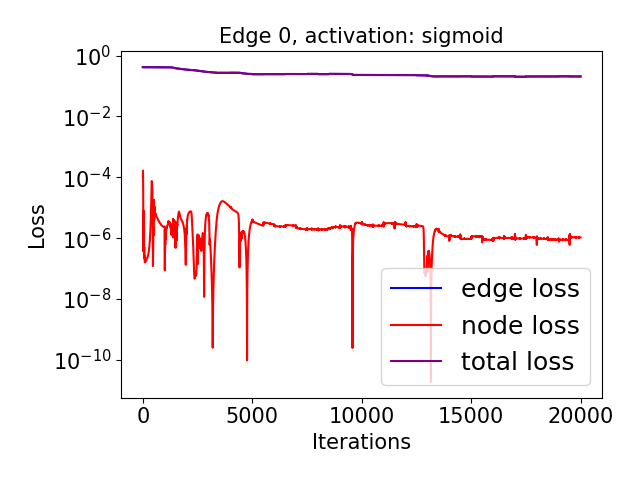}\\
\vspace{-0.2cm}
\includegraphics[width=0.45\textwidth]{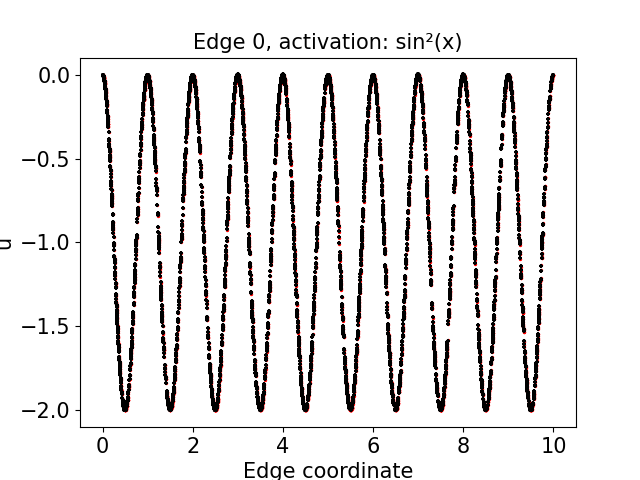}
\includegraphics[width=0.45\textwidth]{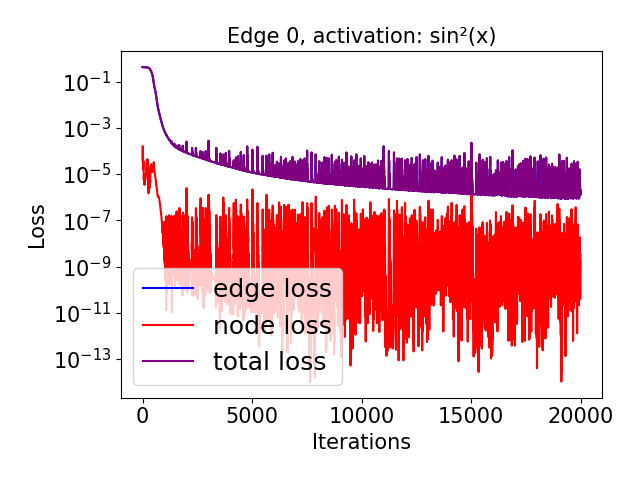}\\
\vspace{-0.2cm}
\includegraphics[width=0.45\textwidth]{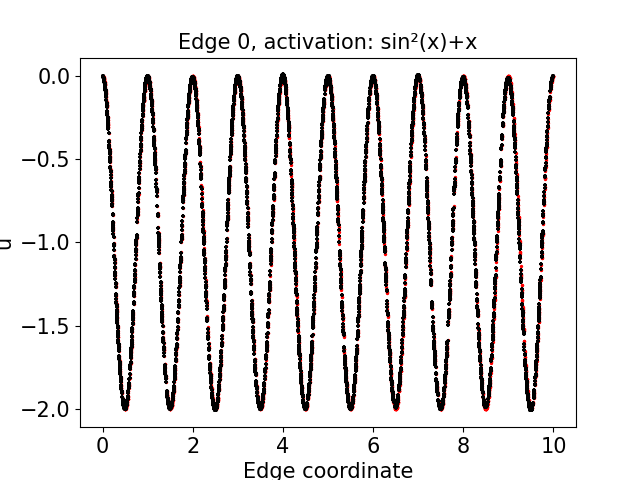}
\includegraphics[width=0.45\textwidth]{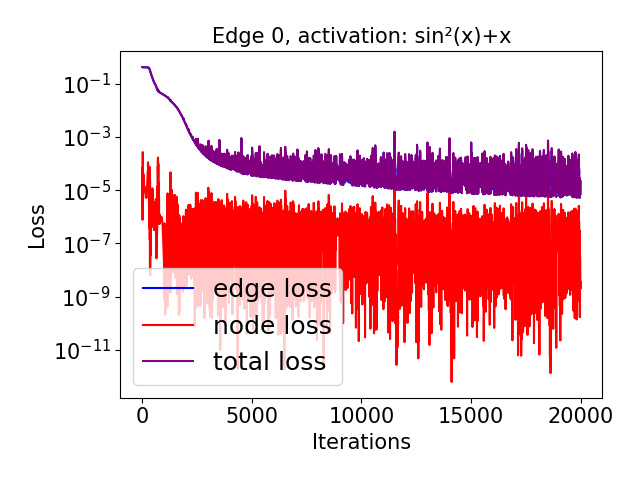}\\
\vspace{-0.2cm}
\caption{Model output for the non-linear elliptic problem on one edge of length $10$ using different activation functions}  \label{hel10}

\end{figure}

\begin{figure}[htbp!]
\vspace{-0.5cm}
\hspace{1cm}\includegraphics[width=0.5\textwidth]{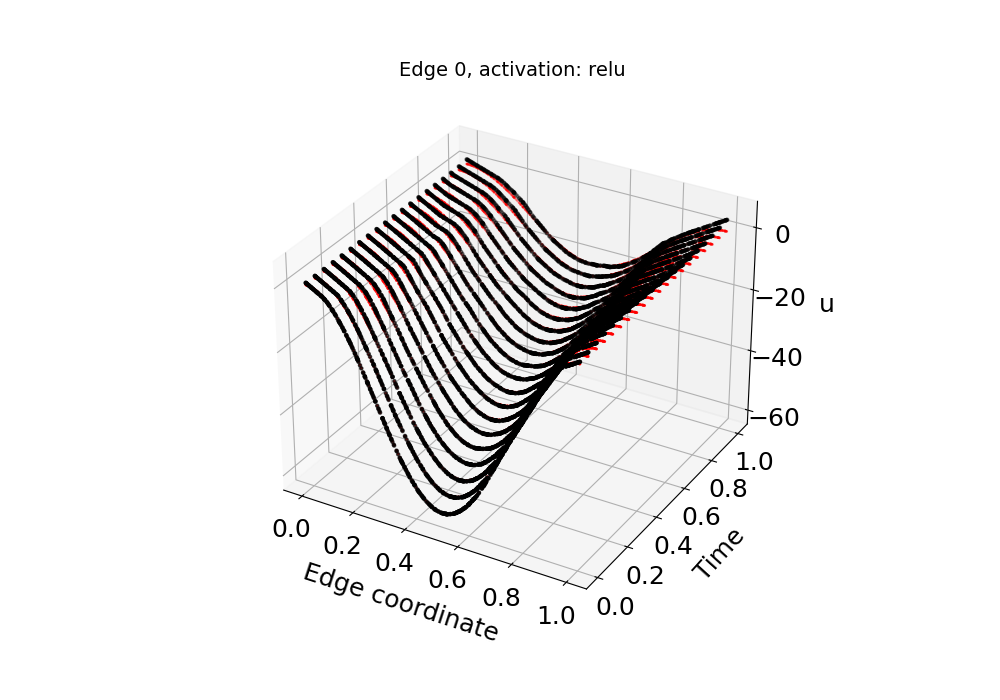}
\raisebox{0.2\height}{\includegraphics[width=0.3\textwidth]{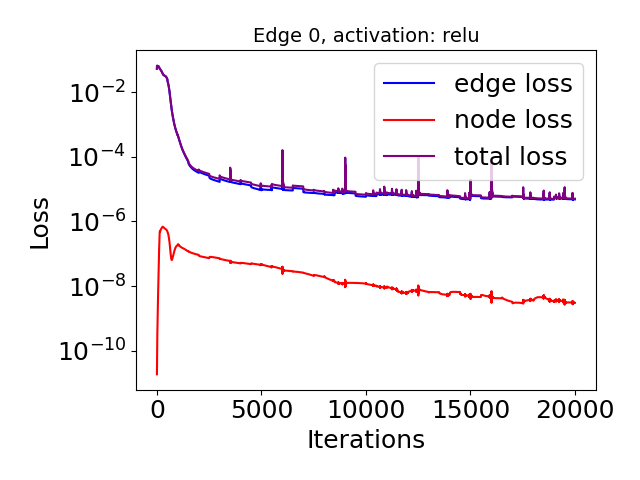}}\\
\vspace{-0.5cm}
\hspace{1cm}\includegraphics[width=0.5\textwidth]{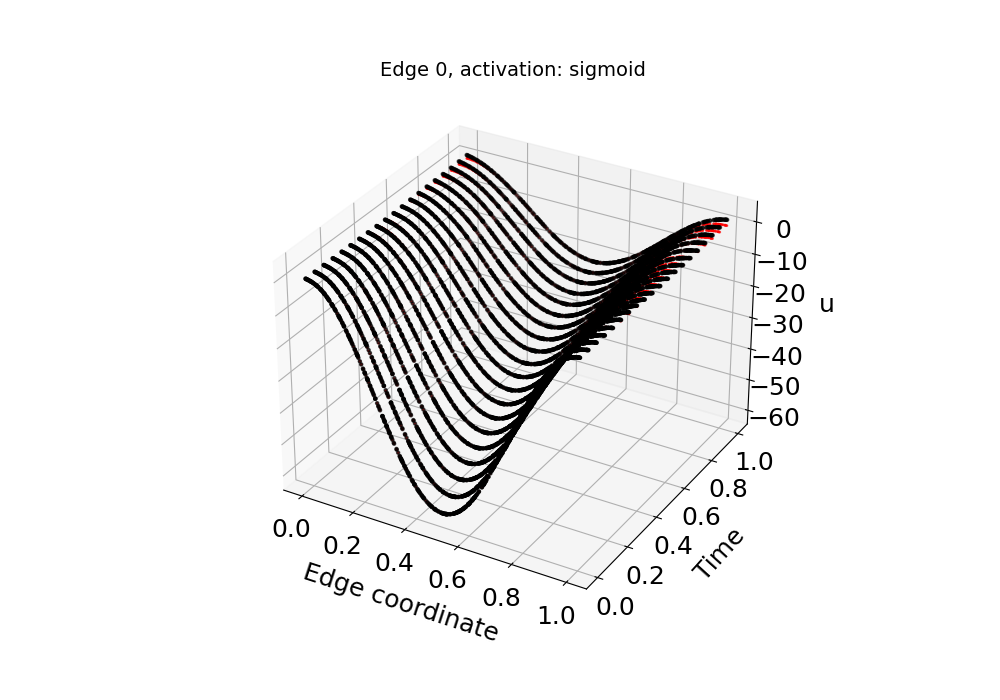}
\raisebox{0.2\height}{\includegraphics[width=0.3\textwidth]{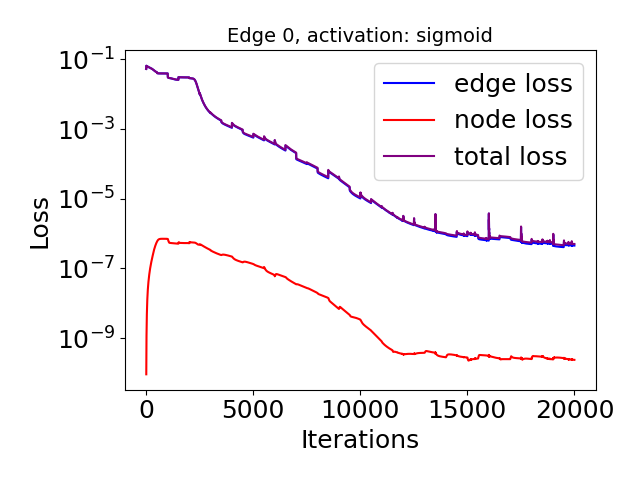}}\\
\vspace{-0.5cm}
\hspace{1cm}\includegraphics[width=0.5\textwidth]{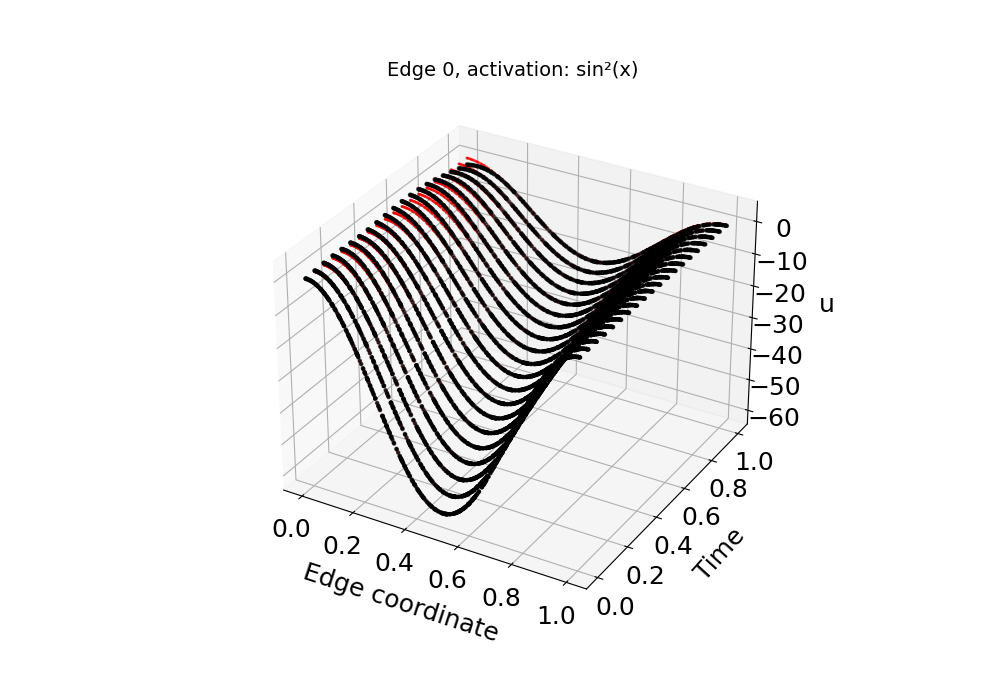}
\raisebox{0.2\height}{\includegraphics[width=0.3\textwidth]{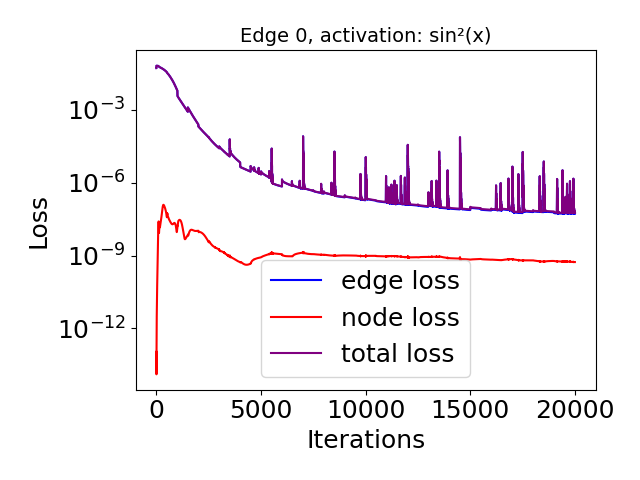}} \\
\vspace{-0.5cm}
\hspace{1cm}\includegraphics[width=0.5\textwidth]{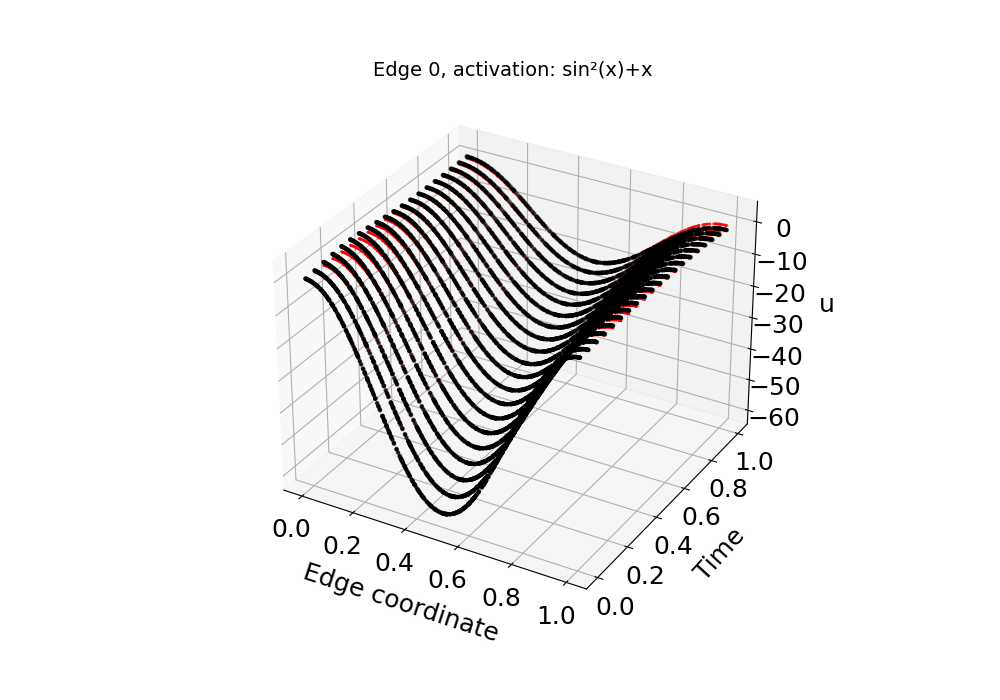}
\raisebox{0.2\height}{\includegraphics[width=0.3\textwidth]{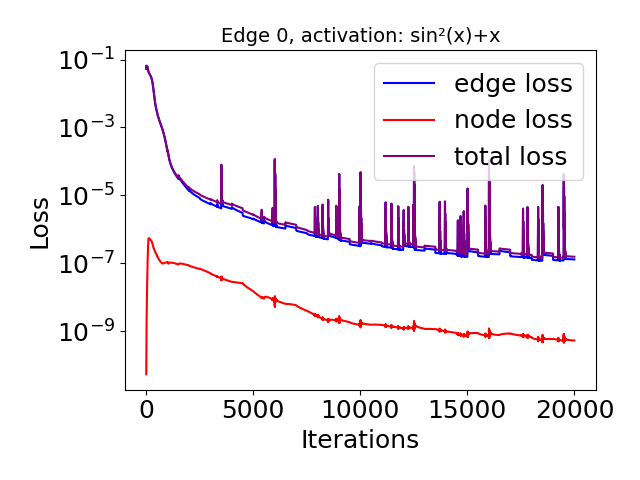}}\\
\caption{Model output for the non-linear parabolic problem on one edge of length $1$ using different activation functions} \label{para1}
\end{figure}

%%%%%%%%%%%%%%%%%%%%%%%%%%%%%%%%%%%%%%%%%%%%%%%%%%%%

\begin{figure}[!htbp]
\vspace{-0.5cm}
\hspace{1cm}
\includegraphics[width=0.5\textwidth]{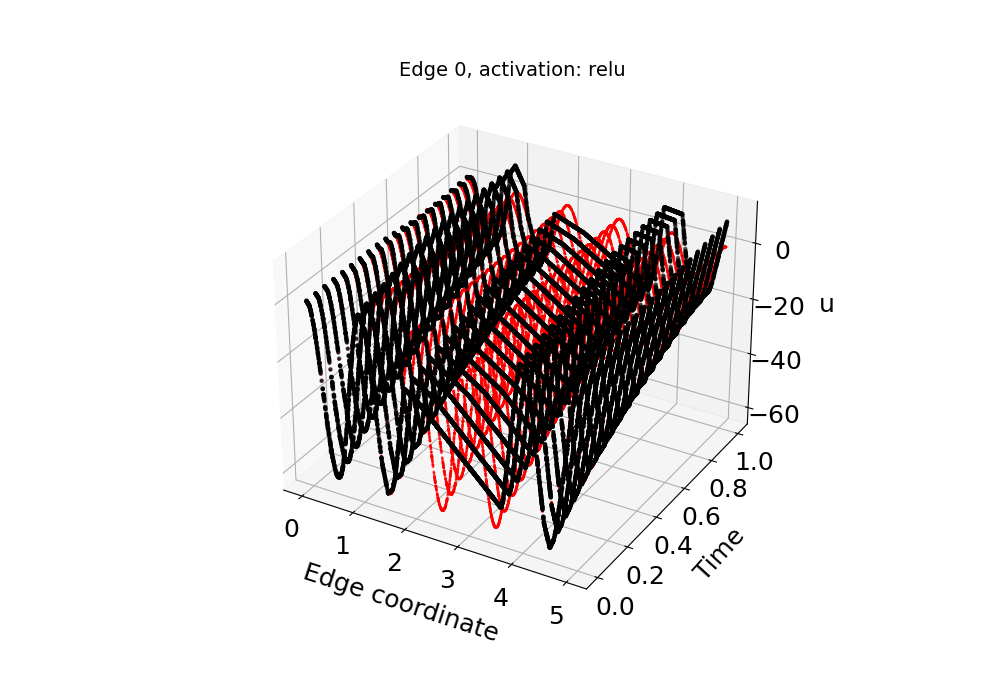}
\raisebox{0.2\height}{\includegraphics[width=0.3\textwidth]{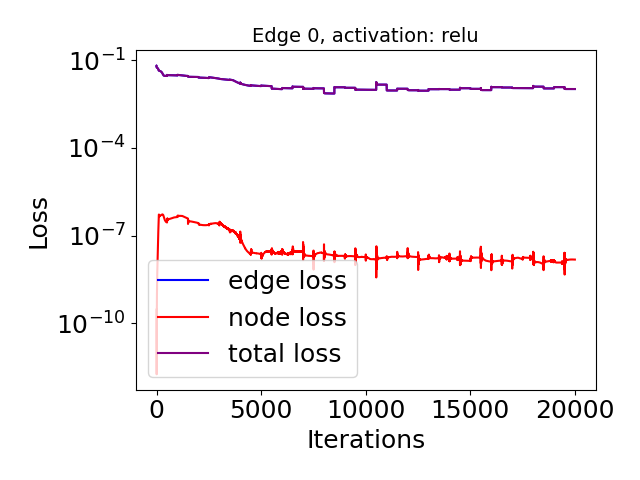}}\\
\vspace{-0.5cm}
\hspace{1cm}\includegraphics[width=0.5\textwidth]{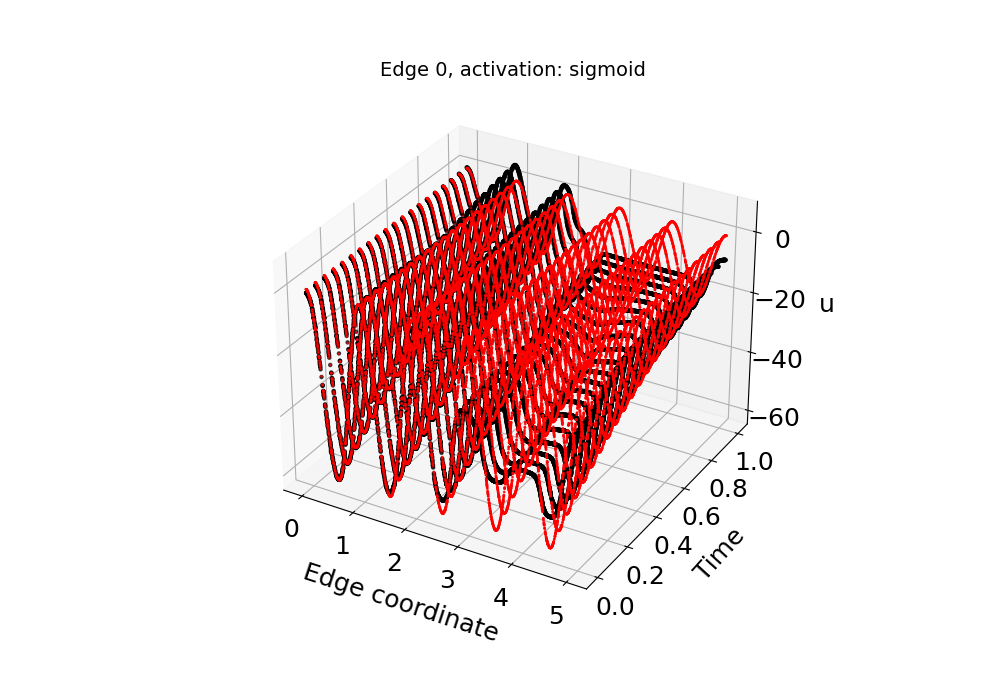}
\raisebox{0.2\height}{\includegraphics[width=0.3\textwidth]{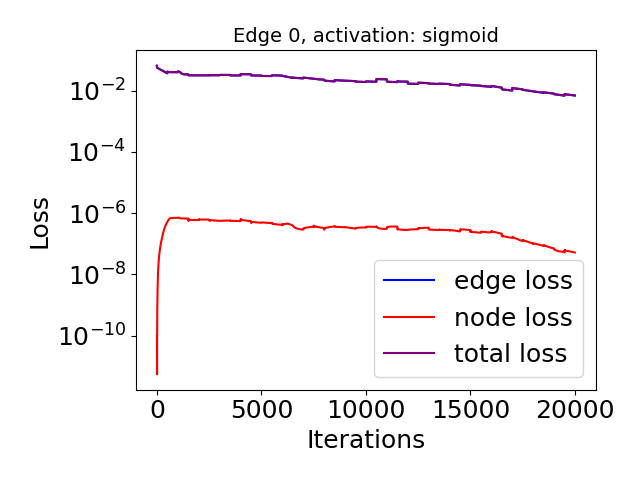}}\\
\vspace{-0.5cm}
\hspace{1cm}
\includegraphics[width=0.5\textwidth]{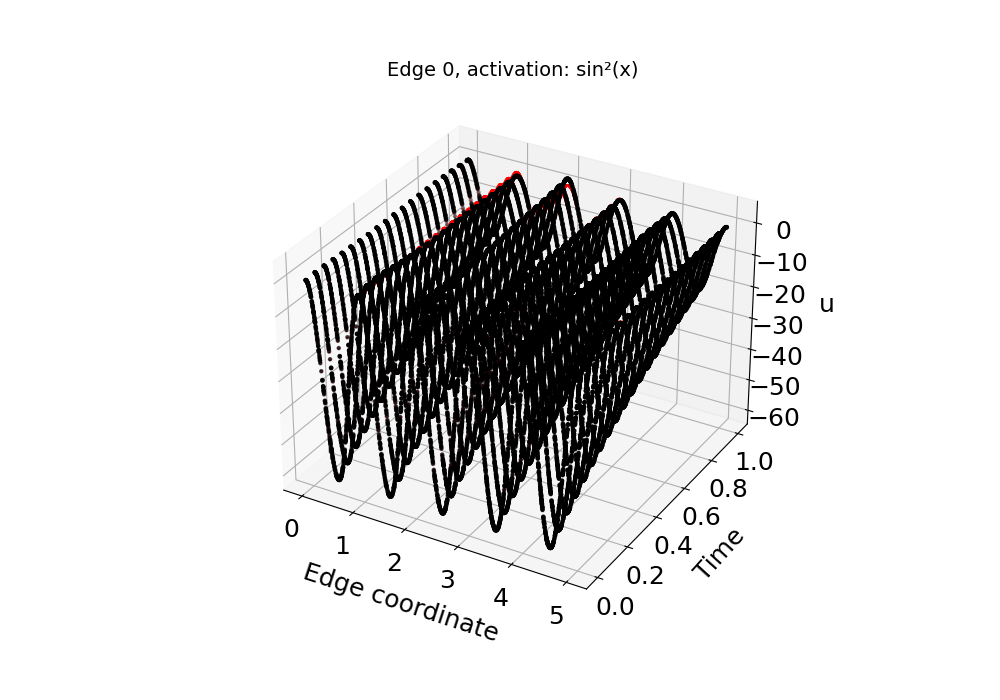}
\raisebox{0.2\height}{\includegraphics[width=0.3\textwidth]{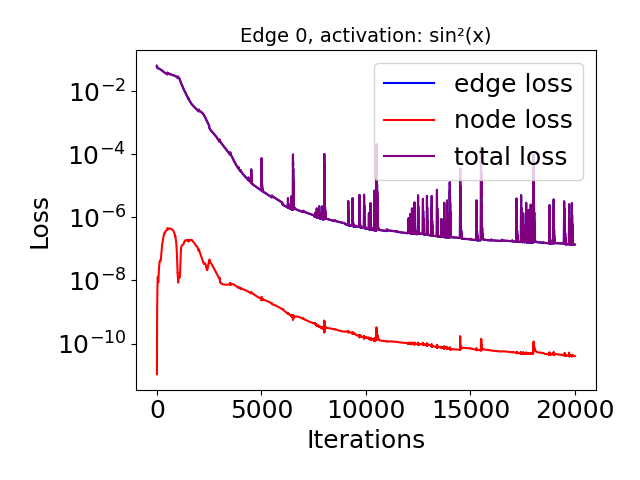}}\\
\vspace{-0.5cm}
\hspace{1cm}\includegraphics[width=0.5\textwidth]{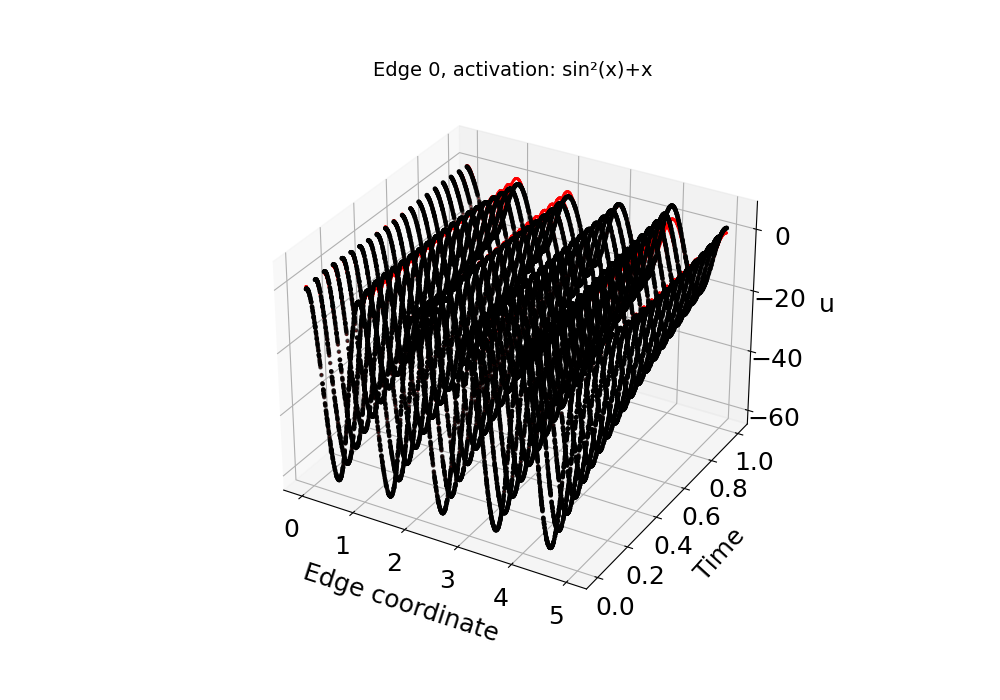}
\raisebox{0.2\height}{\includegraphics[width=0.3\textwidth]{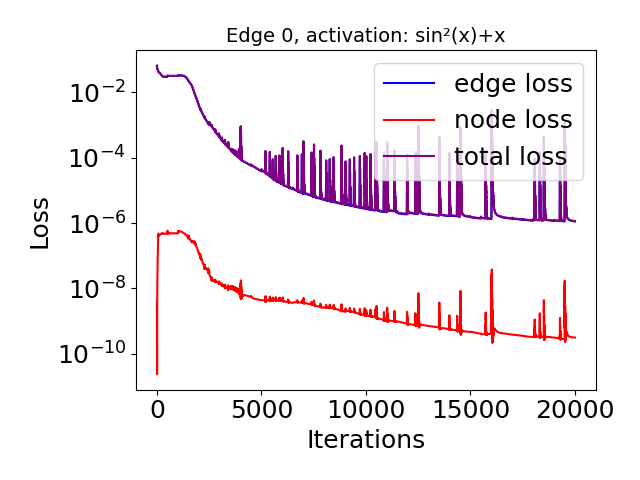}}\\
\caption{Model output for the non-linear parabolic problem on one edge of length $5$ using different activation functions} \label{para5}

\end{figure}

%%%%%%%%%%%%%%%%%%%%%%%%%%%%%%%%%%%%%%%%%%%%%%%%%%%%

\begin{figure}[!htbp]
\vspace{-0.5cm}
\hspace{1cm}
\includegraphics[width=0.5\textwidth]{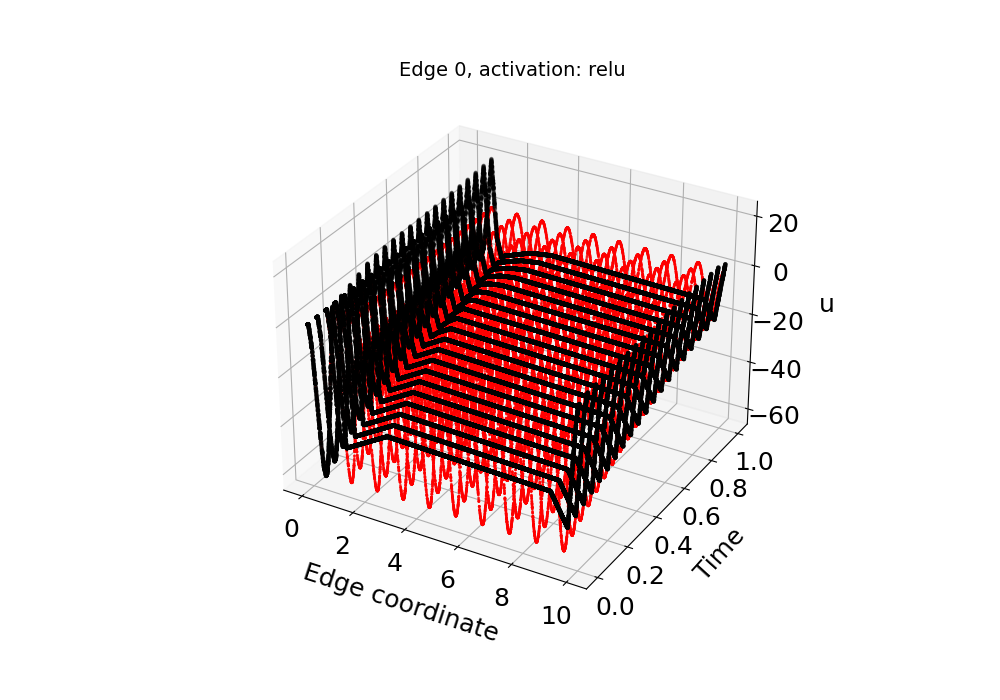}
\raisebox{0.2\height}{\includegraphics[width=0.3\textwidth]{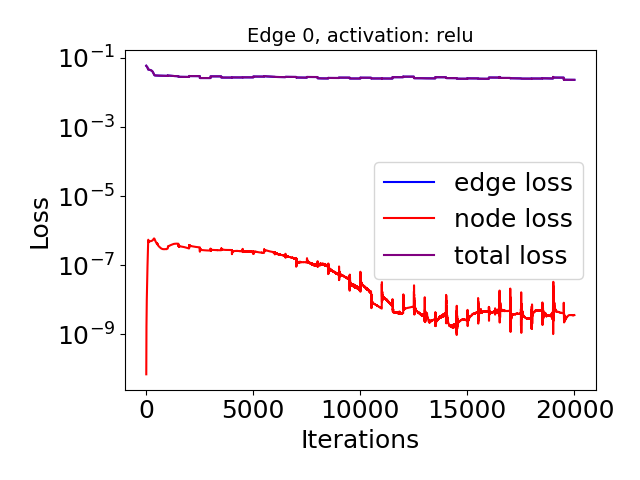}}\\
\vspace{-0.5cm}
\hspace{1cm}\includegraphics[width=0.5\textwidth]{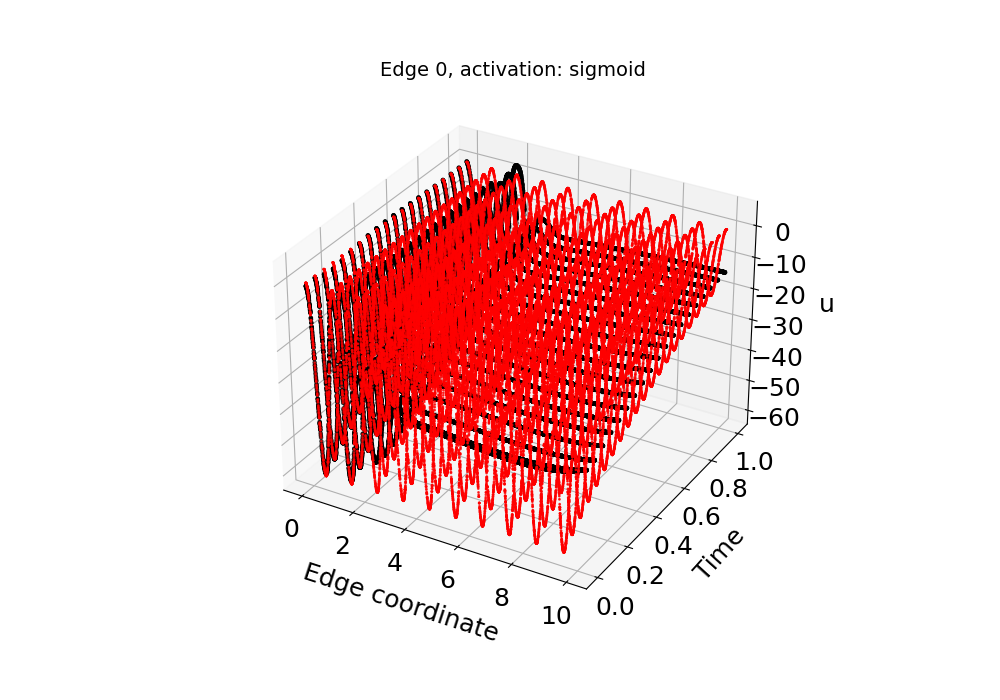}
\raisebox{0.2\height}{\includegraphics[width=0.3\textwidth]{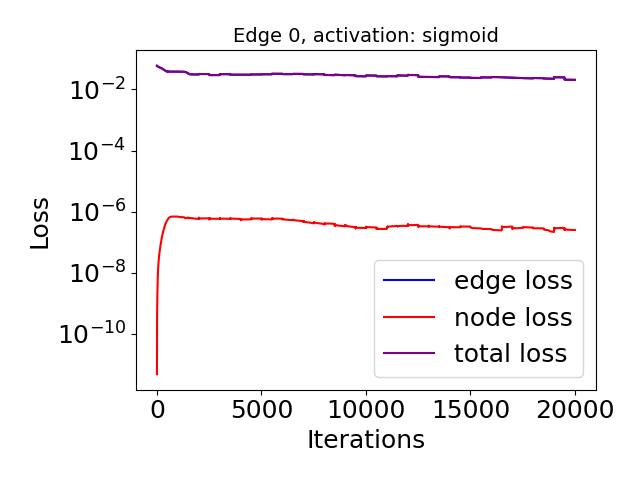}}\\
\vspace{-0.5cm}
\hspace{1cm}\includegraphics[width=0.5\textwidth]{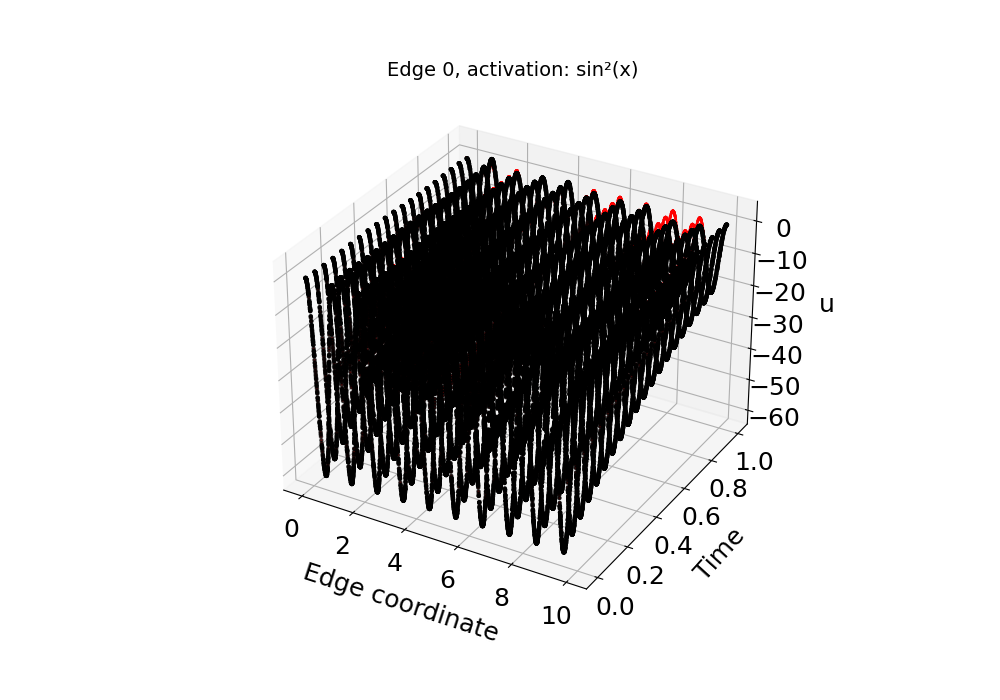}
\raisebox{0.2\height}{\includegraphics[width=0.3\textwidth]{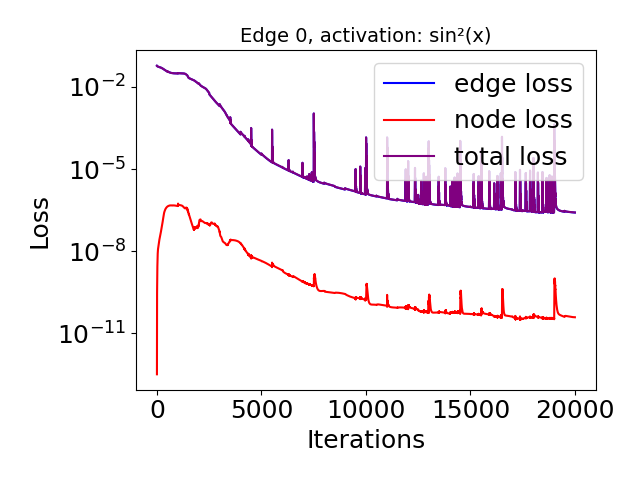}}\\
\vspace{-0.5cm}
\hspace{1cm}\includegraphics[width=0.5\textwidth]{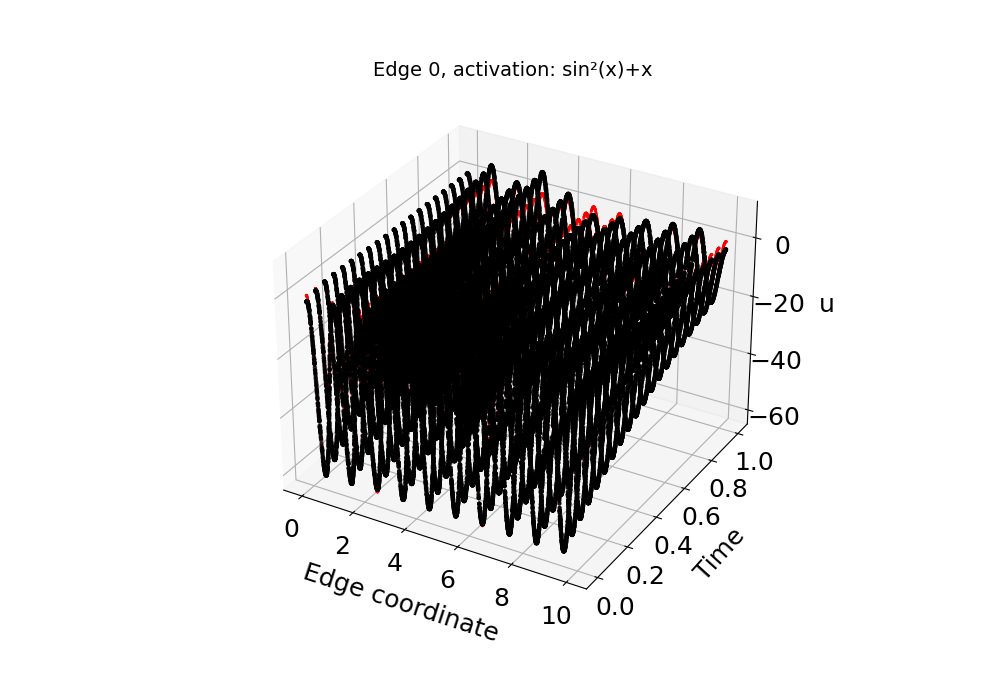}
\raisebox{0.2\height}{\includegraphics[width=0.3\textwidth]{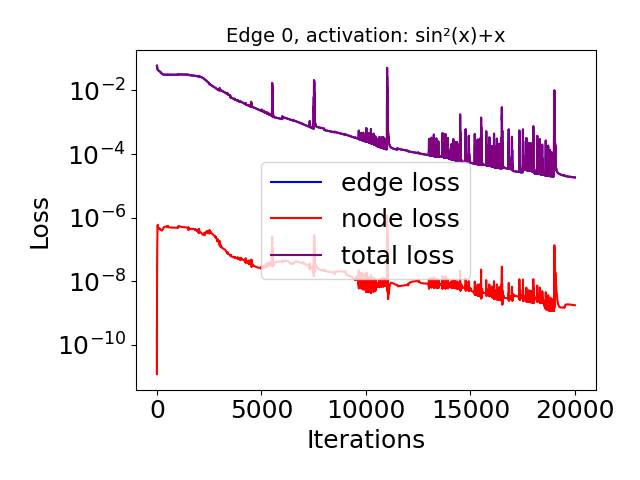}}\\
\caption{Model output for the non-linear parabolic problem on one edge of length $10$ using different activation functions} \label{para10}
\end{figure}

\begin{figure}[htbp!]
\vspace{-0.5cm}
\hspace{1cm}\includegraphics[width=0.5\textwidth]{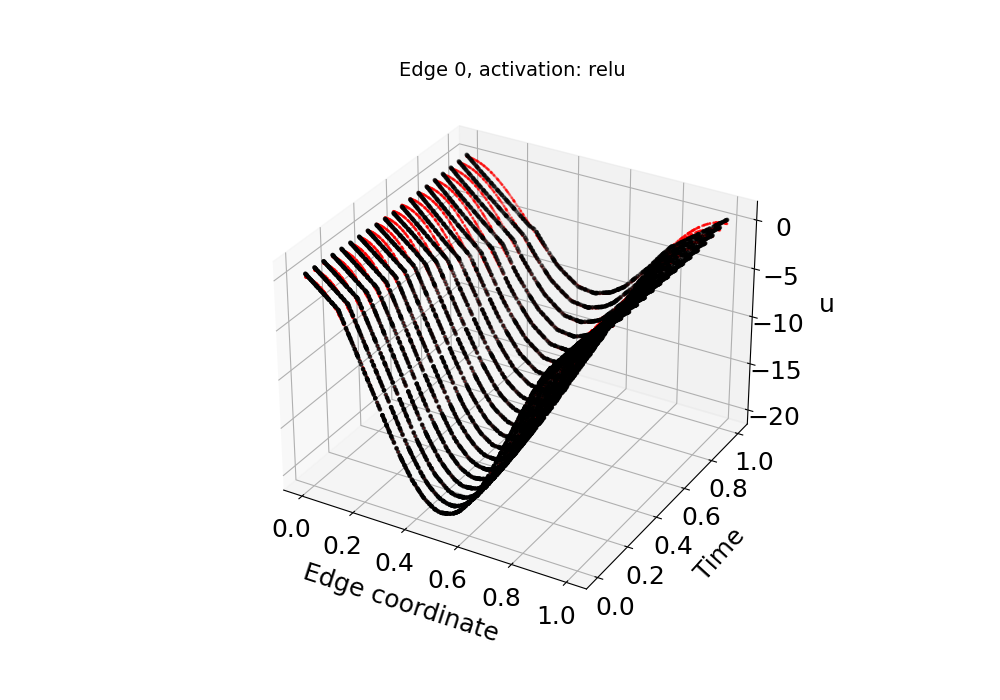}
\raisebox{0.2\height}{\includegraphics[width=0.3\textwidth]{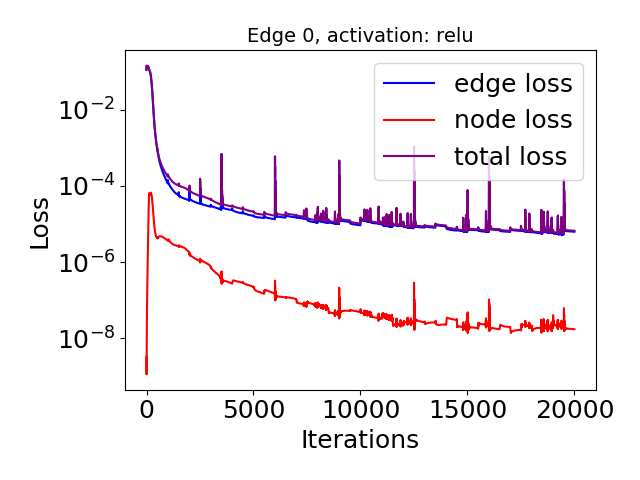}}\\
\vspace{-0.5cm}
\hspace{1cm}\includegraphics[width=0.5\textwidth]{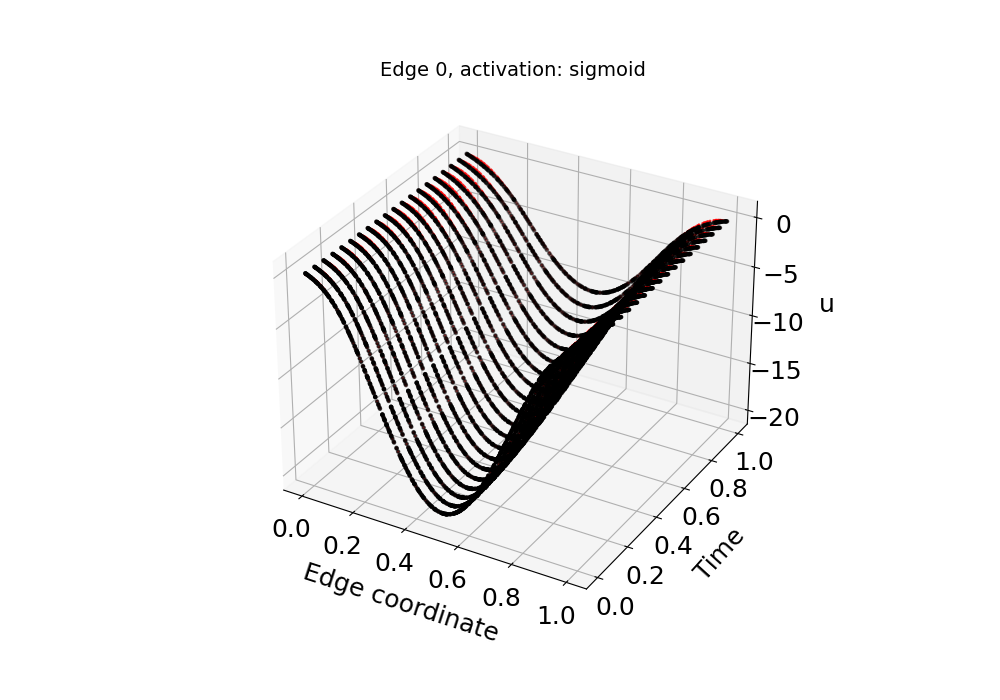}
\raisebox{0.2\height}{\includegraphics[width=0.3\textwidth]{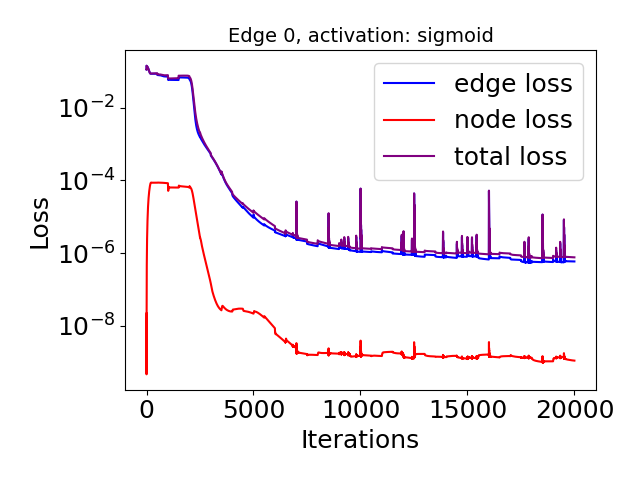}}\\
\vspace{-0.5cm}
\hspace{1cm}\includegraphics[width=0.5\textwidth]{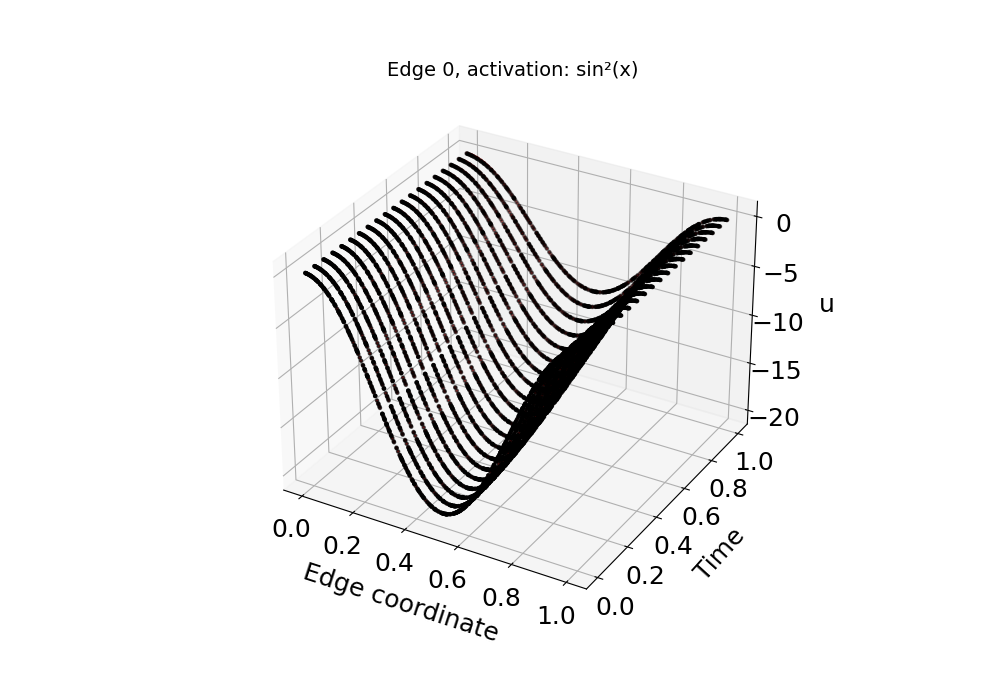}
\raisebox{0.2\height}{\includegraphics[width=0.3\textwidth]{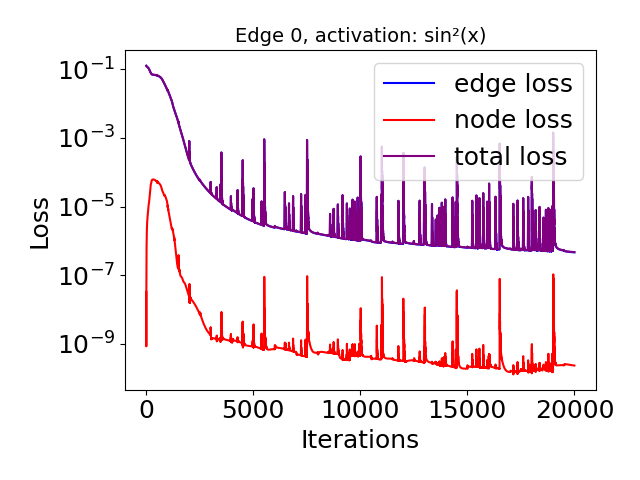}}\\
\vspace{-0.5cm}
\hspace{1cm}\includegraphics[width=0.5\textwidth]{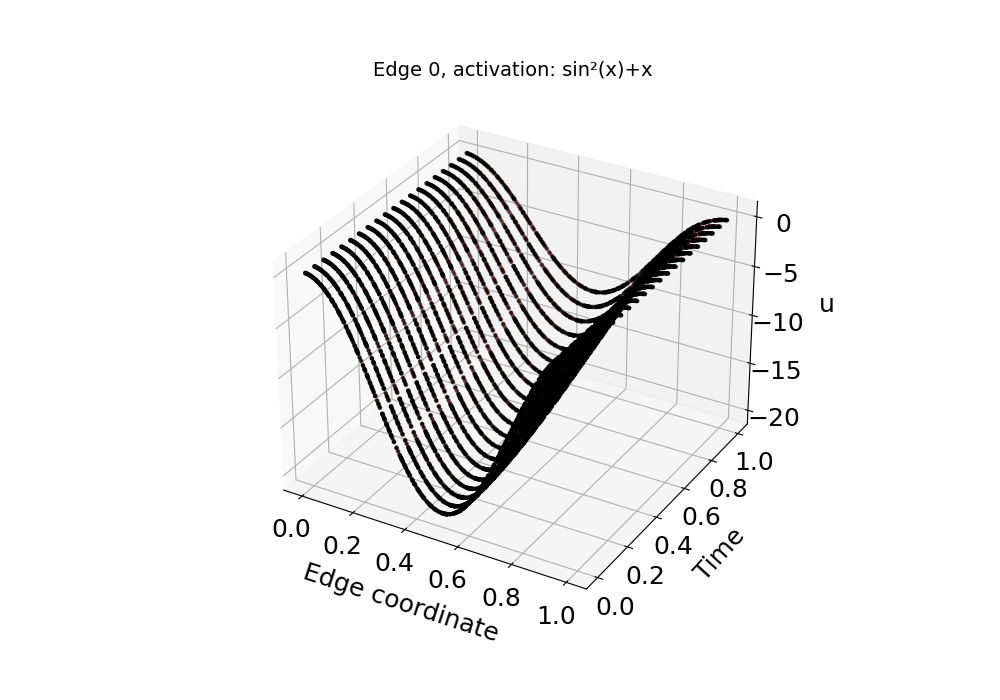}
\raisebox{0.2\height}{\includegraphics[width=0.3\textwidth]{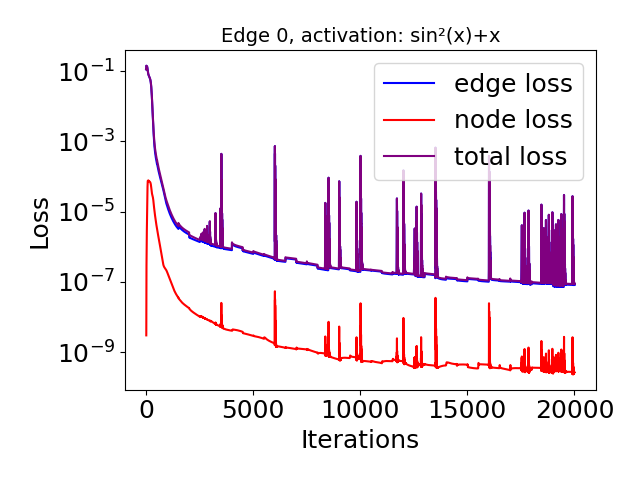}}\\
\caption{Model output for the non-linear hyperbolic problem on one edge of length $1$ using different activation functions} \label{ad1}
\end{figure}
%%%%%%%%%%%%%%%%%%%%%%%%%%%%%%%%%%%%%%%%

\begin{figure}[htbp!]
\vspace{-0.5cm}
\hspace{1cm}\includegraphics[width=0.5\textwidth]{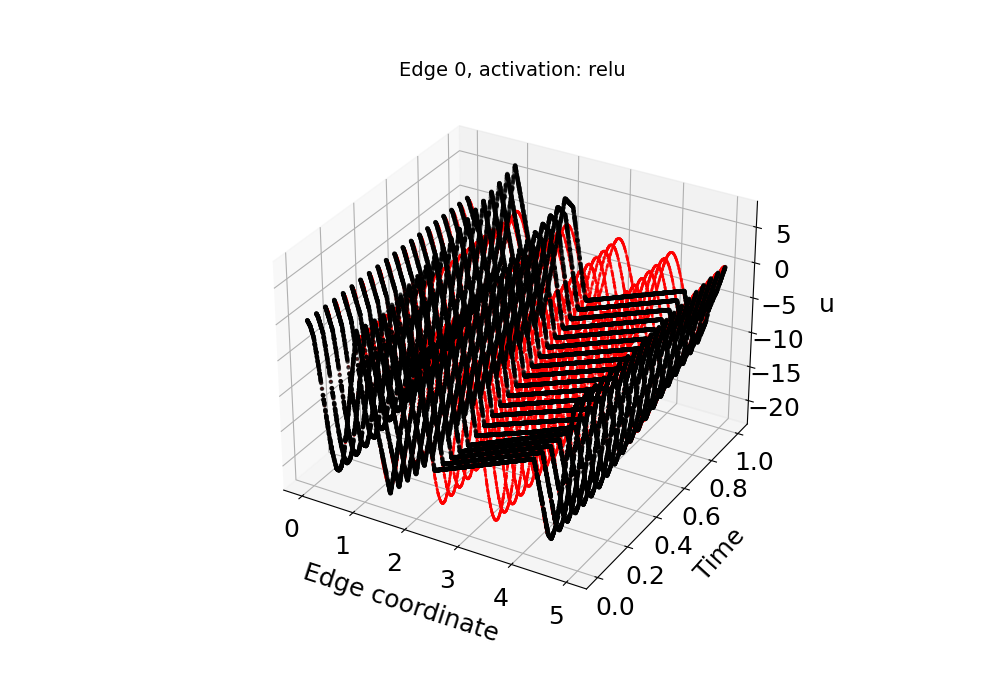}
\raisebox{0.2\height}{\includegraphics[width=0.3\textwidth]{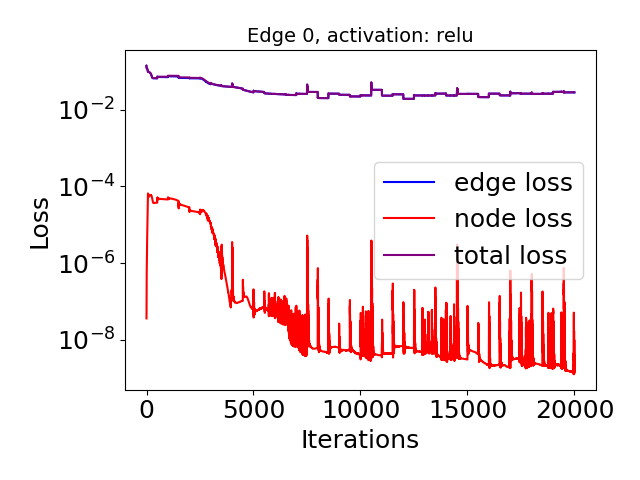}}\\
\vspace{-0.5cm}
\hspace{1cm}\includegraphics[width=0.5\textwidth]{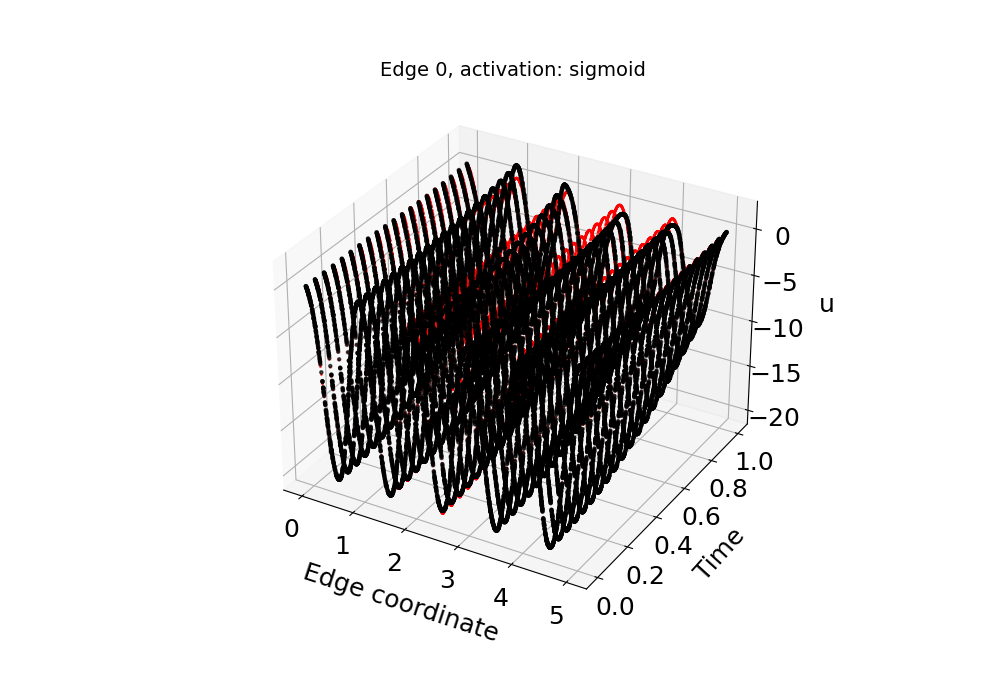}
\raisebox{0.2\height}{\includegraphics[width=0.3\textwidth]{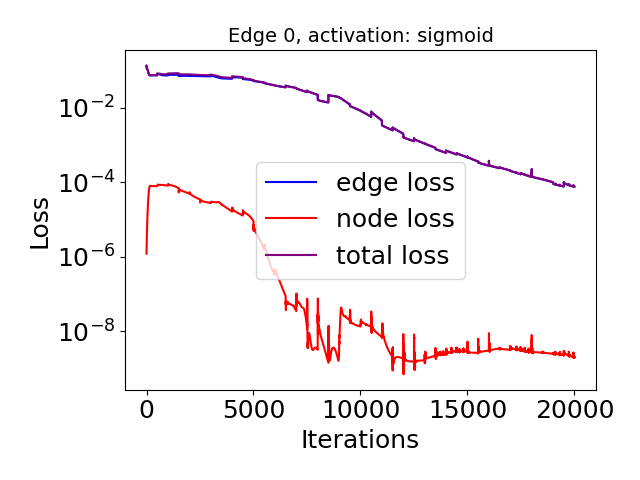}}\\
\vspace{-0.5cm}
\hspace{1cm}\includegraphics[width=0.5\textwidth]{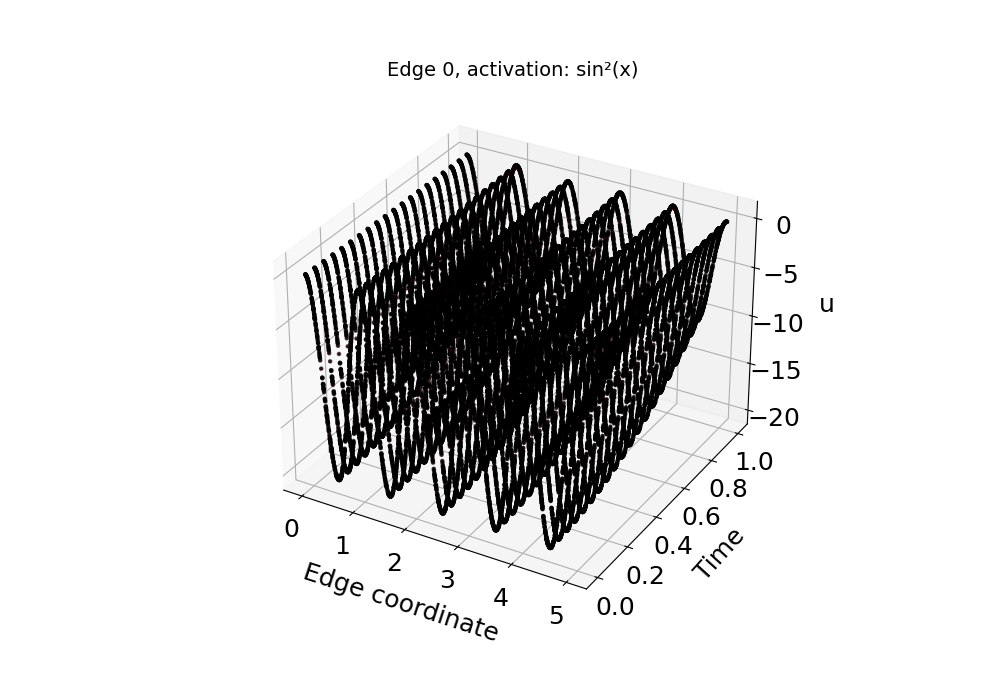}
\raisebox{0.2\height}{\includegraphics[width=0.3\textwidth]{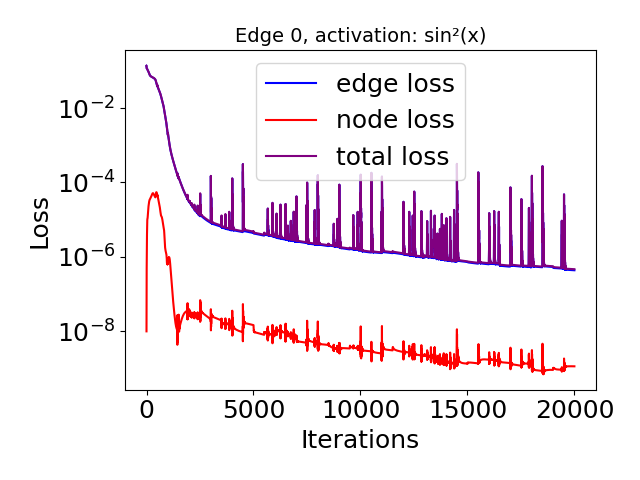}}\\
\vspace{-0.5cm}
\hspace{1cm}\includegraphics[width=0.5\textwidth]{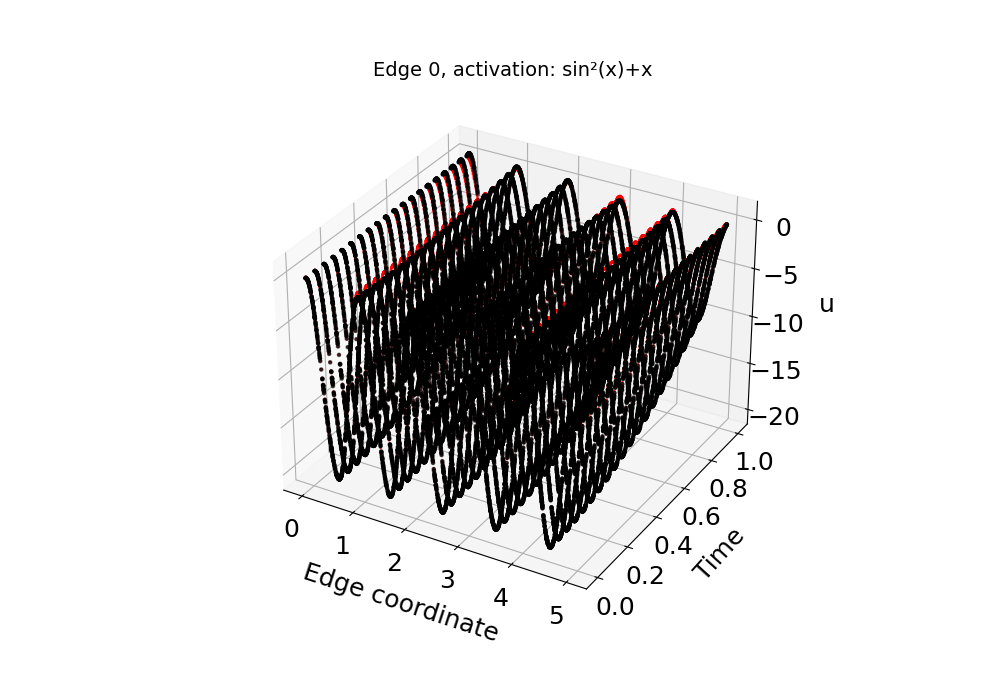}
\raisebox{0.2\height}{\includegraphics[width=0.3\textwidth]{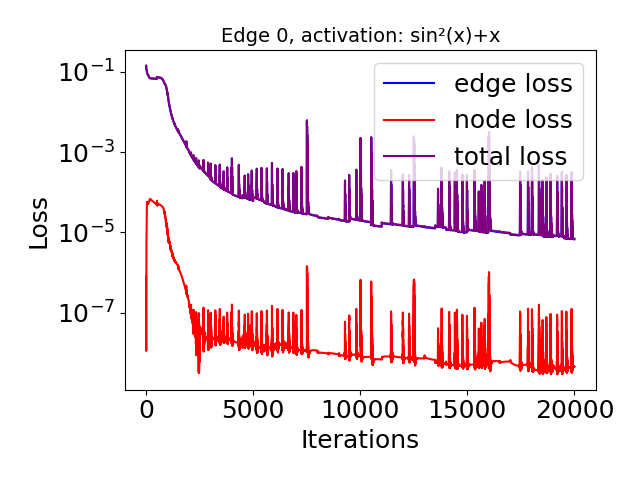}}\\
\caption{Model output for the non-linear hyperbolic problem on one edge of length $5$ using different activation functions} \label{ad5}

\end{figure}

%%%%%%%%%%%%%%%%%%%%%%%%%%%%%%%%%%%%%%%%

\begin{figure}[htbp!]
\vspace{-0.5cm}
\hspace{1cm}\includegraphics[width=0.5\textwidth]{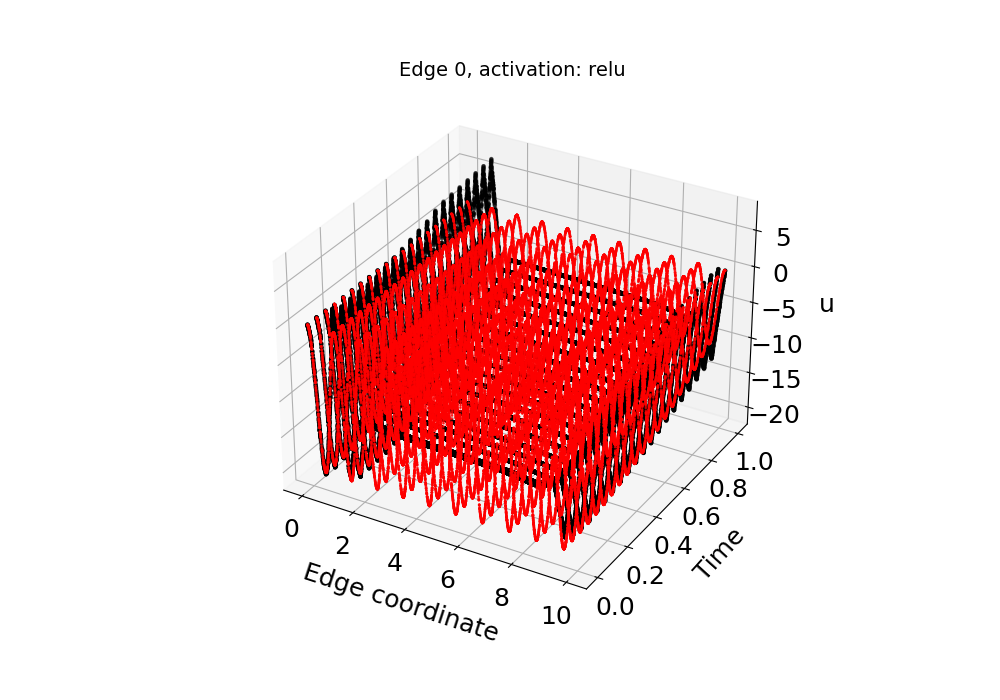}
\raisebox{0.2\height}{\includegraphics[width=0.3\textwidth]{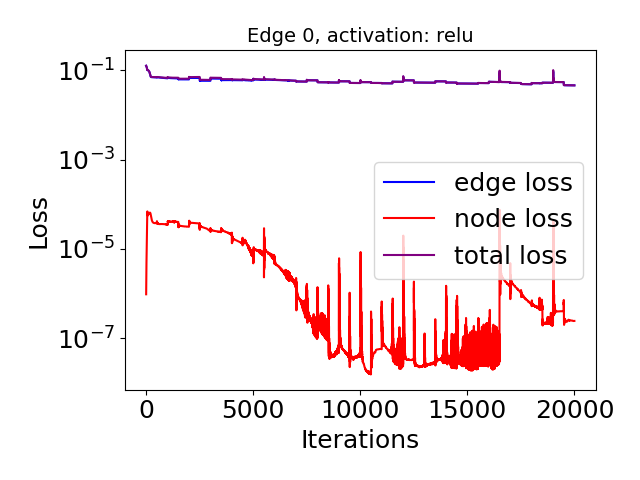}}\\
\vspace{-0.5cm}
\hspace{1cm}\includegraphics[width=0.5\textwidth]{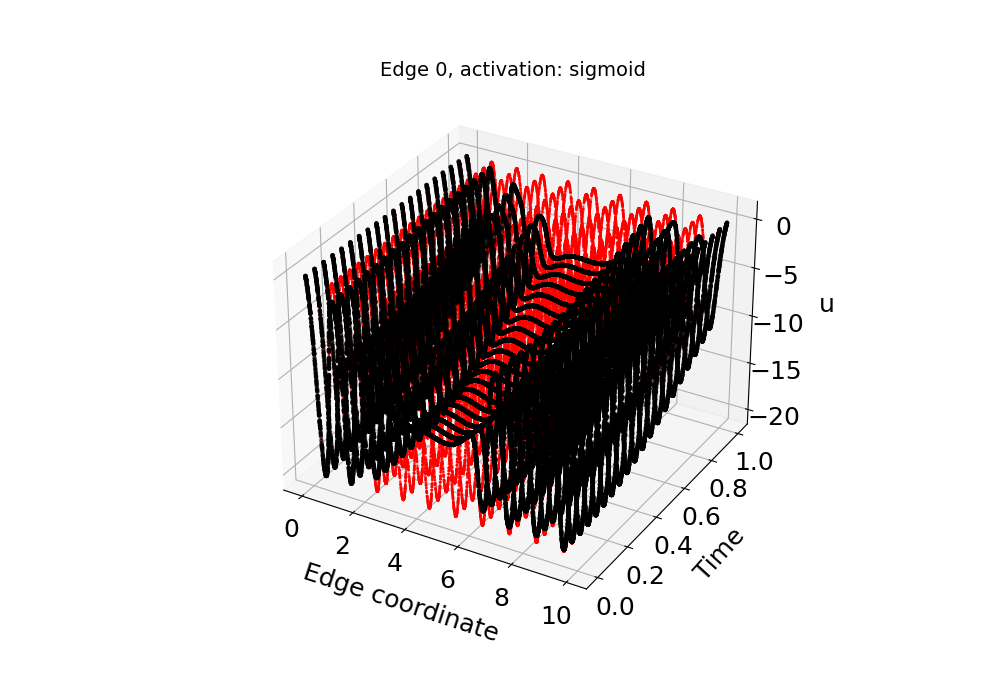}
\raisebox{0.2\height}{\includegraphics[width=0.3\textwidth]{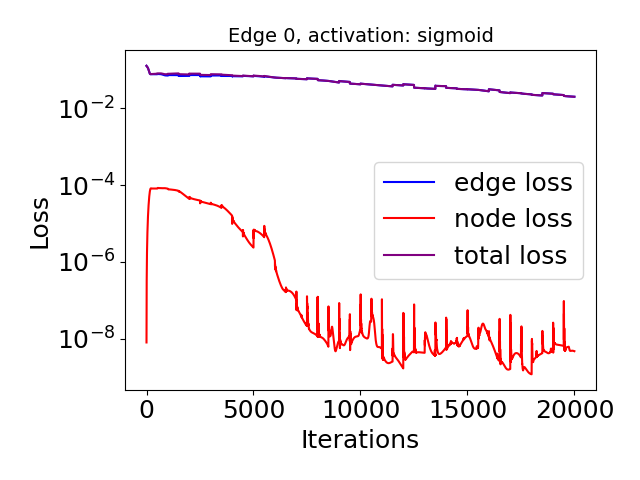}}\\
\vspace{-0.5cm}
\hspace{1cm}\includegraphics[width=0.5\textwidth]{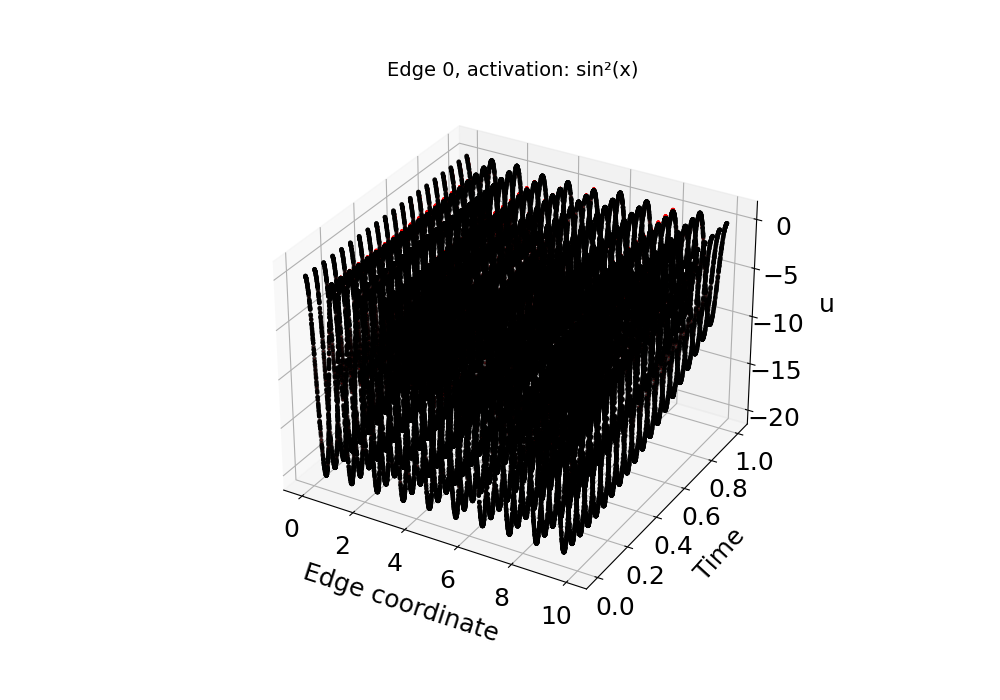}
\raisebox{0.2\height}{\includegraphics[width=0.3\textwidth]{advection_edge10_s2_Amp=10_lalr=1000.000000_edge_0_loss_50.png}}\\
\vspace{-0.5cm}
\hspace{1cm}\includegraphics[width=0.5\textwidth]{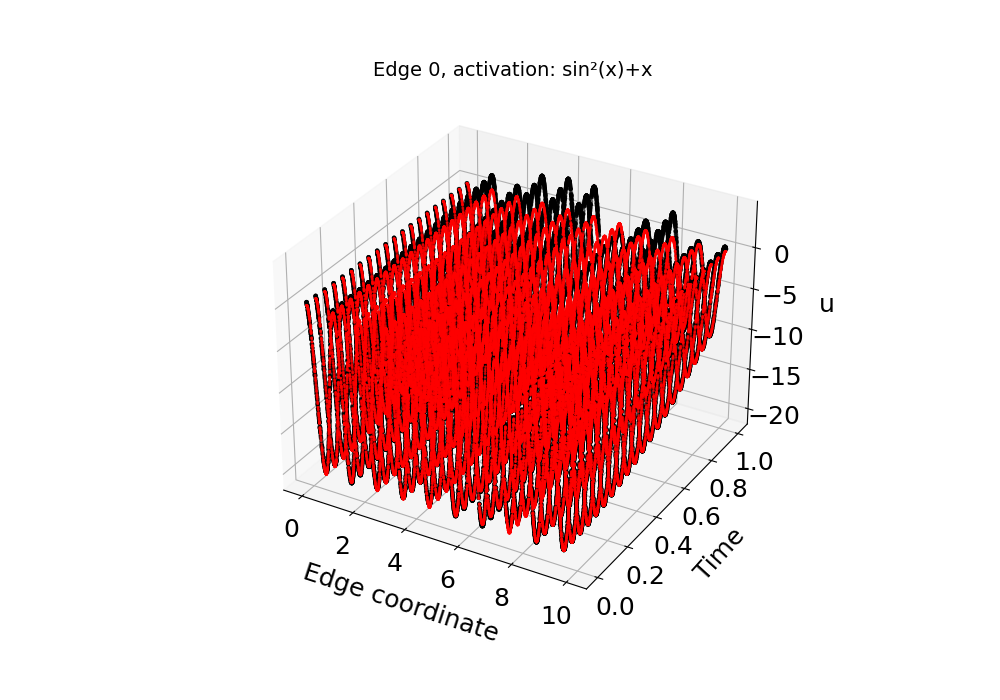}
\raisebox{0.2\height}{\includegraphics[width=0.3\textwidth]{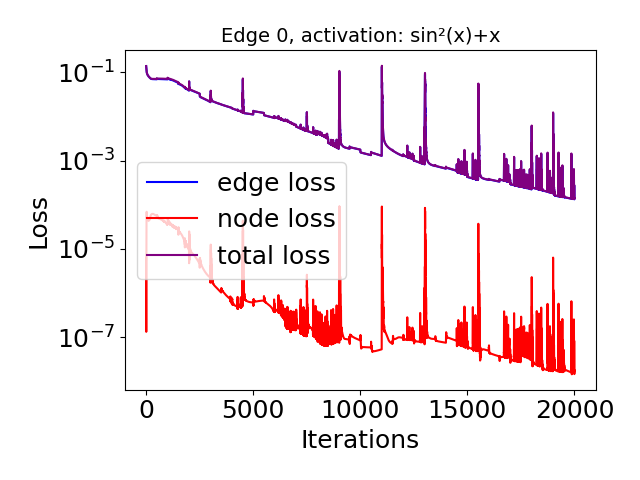}}\\
\caption{Model output for the non-linear hyperbolic problem on one edge of length $10$ using different activation functions} \label{ad10}
\end{figure}
%%%%%%%%%%%%%%%%%%%%%%%%%%%%%%%%%%%%%%%%

\begin{figure}[htbp!]

{\hspace{5cm}\includegraphics[width=0.35\textwidth]{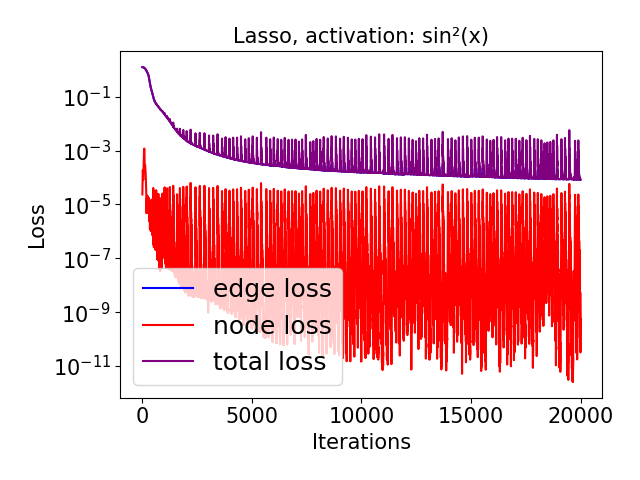}}\\
\includegraphics[width=0.33\textwidth]{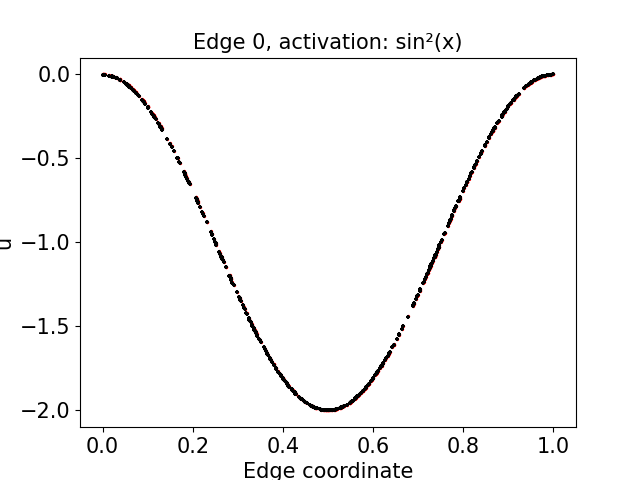}
\includegraphics[width=0.33\textwidth]{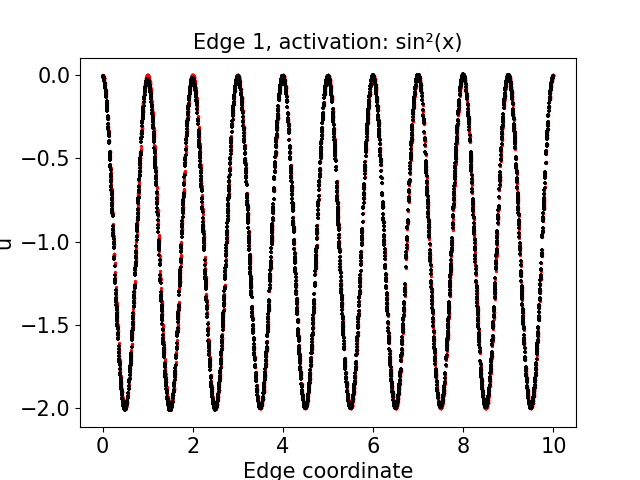}
\includegraphics[width=0.33\textwidth]{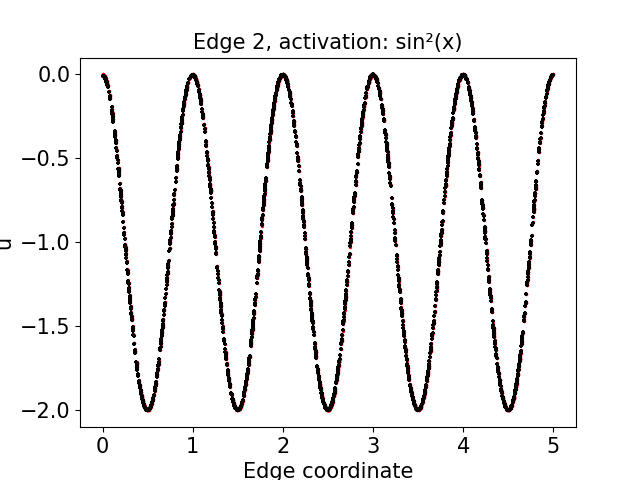}
\caption{Validation error (top figure) and comparison on each edge (figures on the second and third line) of the exact solution (red) and PINN approximated solution (black) of the non-linear elliptic problem defined on the Lasso graph.}
\label{lasso_graph_elliptic}
\end{figure}

\begin{figure}[htbp!]

{\hspace{5cm}\includegraphics[width=0.35\textwidth]{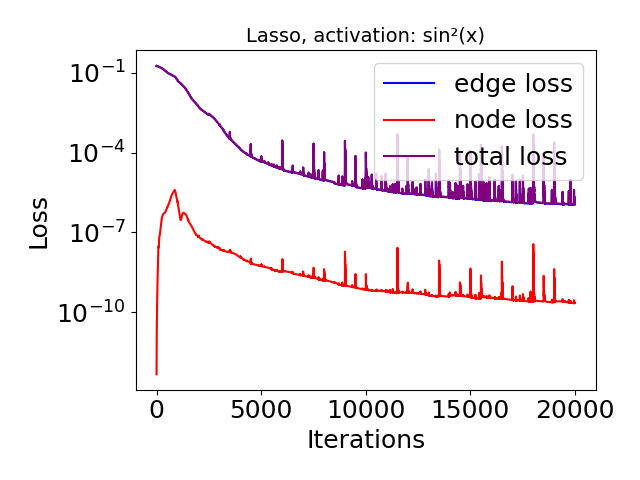}}\\
\vspace{-0.5cm}
\adjustbox{trim={.25\width} {0\height} {.15\width} {0\height},clip}{\includegraphics[width=0.55\textwidth]{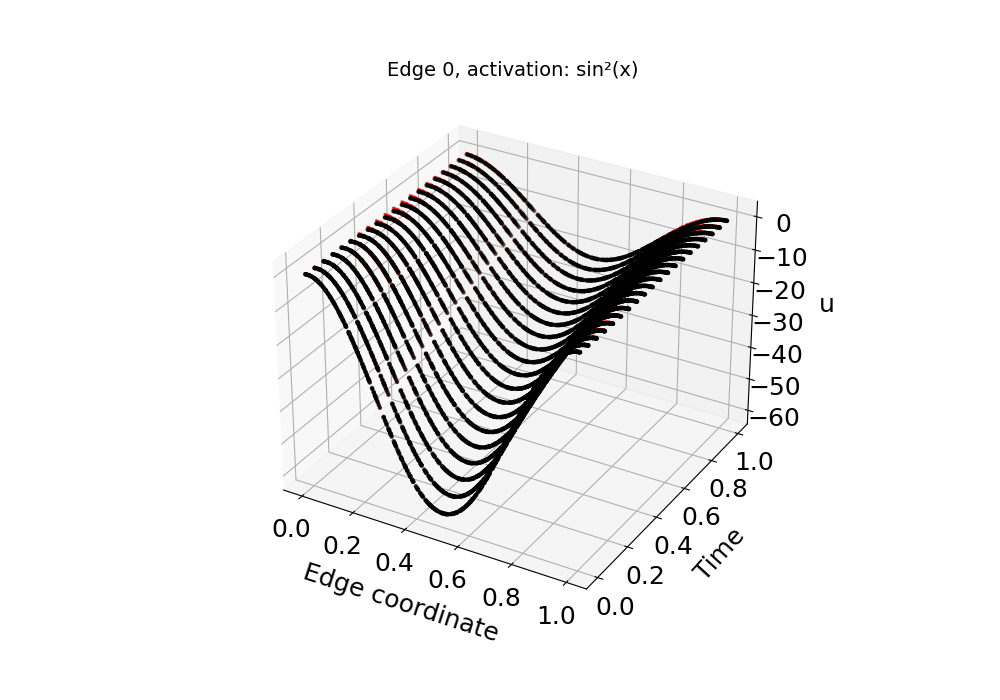}}
\adjustbox{trim={.25\width} {0\height} {.15\width} {0\height},clip}{\includegraphics[width=0.55\textwidth]{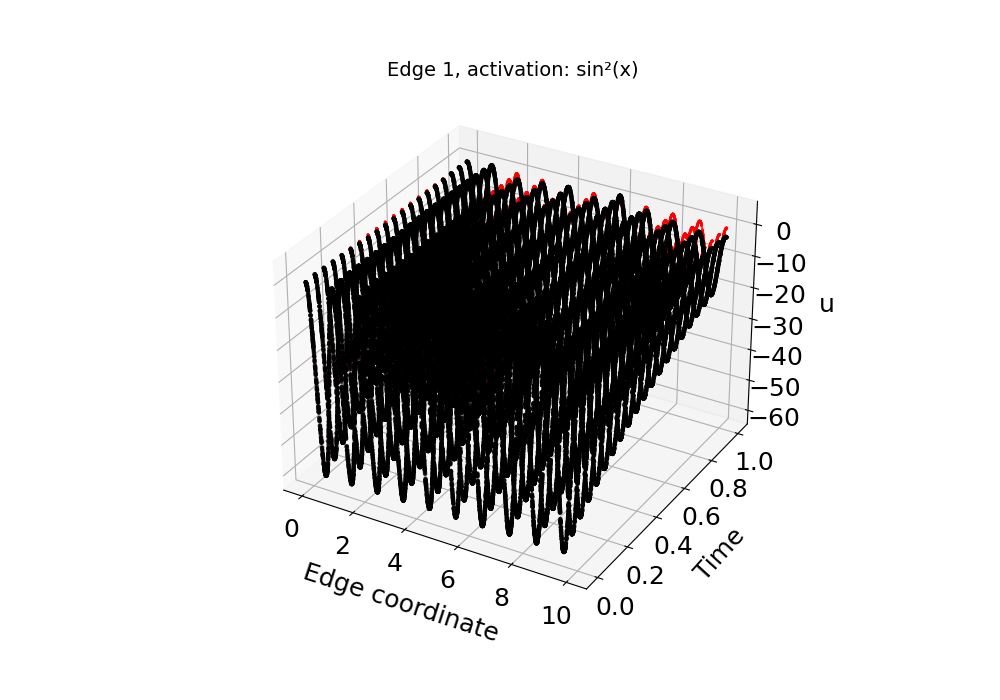}}
\adjustbox{trim={.25\width} {0\height} {.15\width} {0\height},clip}{\includegraphics[width=0.55\textwidth]{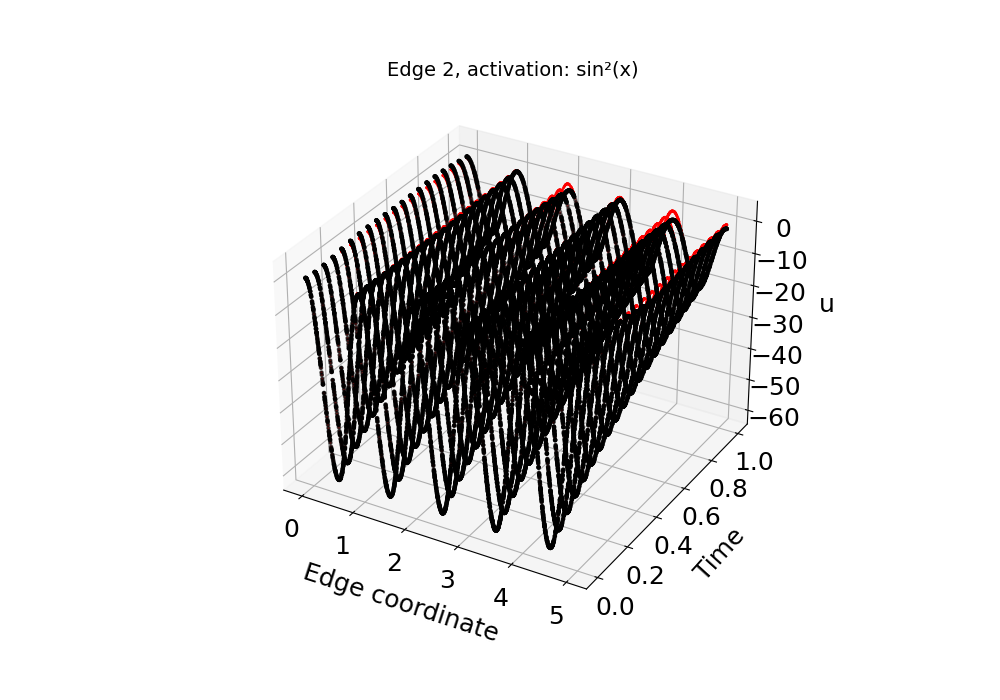}}
\caption{Validation error (top figure) and comparison on each edge (figures on the second and third line) of the exact solution (red) and PINN approximated solution (black) of the non-linear parabolic problem defined on the Lasso graph.}
\label{lasso_graph_parabolic}
\end{figure}

\begin{figure}[htbp!]

{\hspace{5cm}\includegraphics[width=0.35\textwidth]{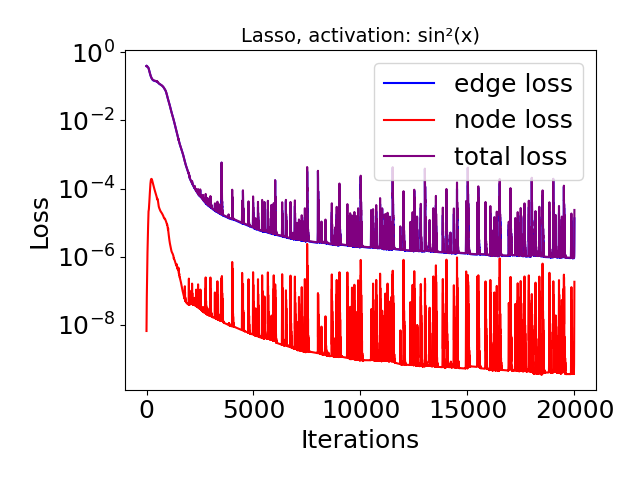}}\\
\vspace{-0.5cm}
\adjustbox{trim={.25\width} {0\height} {.15\width} {0\height},clip}{\includegraphics[width=0.55\textwidth]{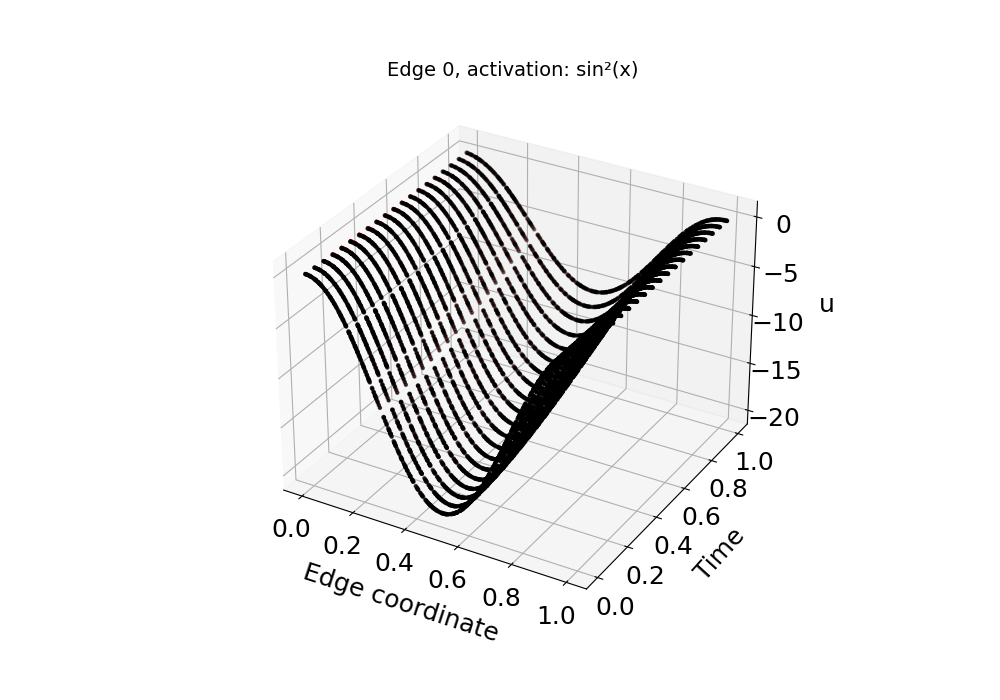}}
\adjustbox{trim={.25\width} {0\height} {.15\width} {0\height},clip}{\includegraphics[width=0.55\textwidth]{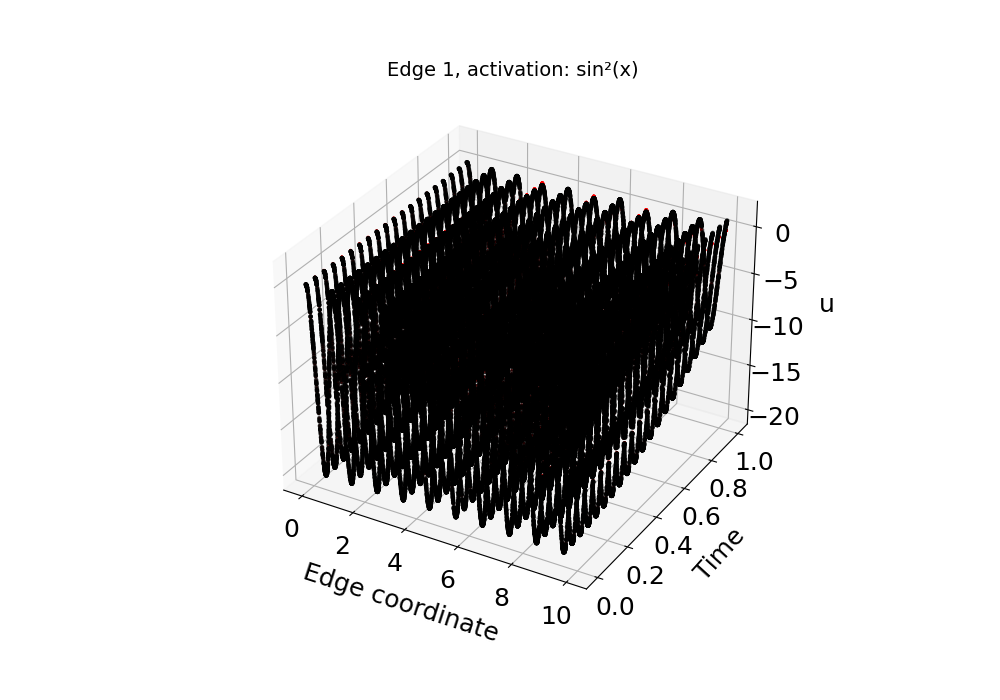}}
\adjustbox{trim={.25\width} {0\height} {.15\width} {0\height},clip}{\includegraphics[width=0.55\textwidth]{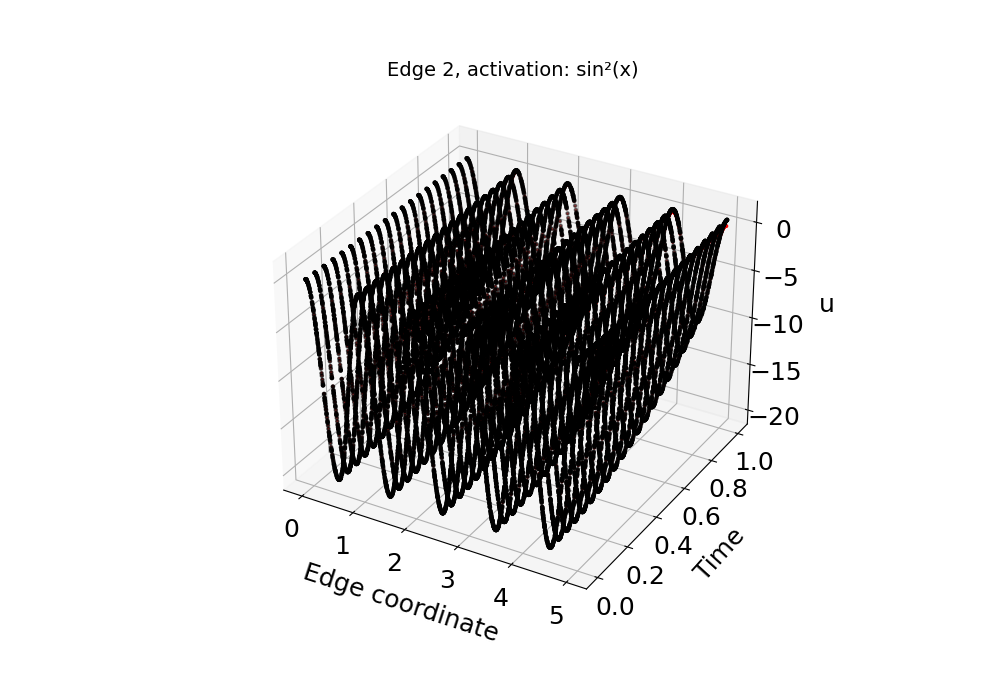}}
\caption{Validation error (top figure) and comparison on each edge (figures on the second and third line) of the exact solution (red) and PINN approximated solution (black) of the non-linear hyperbolic problem defined on the Lasso graph.}\label{lasso_graph_hyperbolic}
\end{figure}

\end{document}